\documentclass{article}

\usepackage{arxiv}

\usepackage{lineno,hyperref}
\modulolinenumbers[5]
\usepackage[T1]{fontenc}
\usepackage{graphicx}
\usepackage{amssymb,amsmath,amsthm}
\usepackage{mathtools}
\usepackage{booktabs}	
\usepackage{float}	
\usepackage{subcaption} 
\usepackage{enumitem}
\usepackage{bm}
\usepackage{color,soul} 
\usepackage{array}
\usepackage[ruled,linesnumbered]{algorithm2e}
\usepackage{siunitx} 
\usepackage{cancel} 
\usepackage{comment} 

\usepackage[draft,inline,nomargin,index]{fixme}
\fxsetup{theme=color,mode=multiuser}
\FXRegisterAuthor{rn}{arn}{\color{blue} RN}
\FXRegisterAuthor{sn}{asn}{\color{red} SN}

\newcommand{\mysize}{0.6}  
\newcommand{\norm}[1]{\left\lVert#1\right\rVert}

\newcolumntype{d}[1]{D{.}{.}{#1}}
\newcommand{\matr}[1]{\mathbf{{#1}}}
\newcommand{\ve}[1]{\bm{{#1}}}
\newcommand{\E}[1]{\mathbb{E} \left[ #1 \right]}

\DeclareMathOperator{\R}{\mathbb{R}}
\DeclareMathOperator{\n}{\mathit{n}_{\text{dof}}}
\DeclareMathOperator{\np}{\mathit{n}_{\mathit{p}}}
\DeclareMathOperator{\nf}{\mathit{n}_{\mathit{f}}}
\DeclareMathOperator{\ns}{\mathit{n}_{\mathit{x}}}
\DeclareMathOperator{\no}{\mathit{n}_{\mathit{y}}}

\DeclareMathOperator{\x}{\ve{x}}
\DeclareMathOperator{\z}{\ve{z}}
\DeclareMathOperator{\y}{\ve{y}}
\DeclareMathOperator{\p}{\ve{p}}
\DeclareMathOperator{\f}{\ve{f}}
\DeclareMathOperator{\m}{\ve{m}}
\DeclareMathOperator{\etav}{\ve{\eta}}
\DeclareMathOperator{\udd}{\ddot{\ve{u}}}
\DeclareMathOperator{\wn}{\ve{w}}
\DeclareMathOperator{\vn}{\ve{v}}
\DeclareMathOperator{\g}{\ve{g}}
\DeclareMathOperator{\tcf}{\ve{\theta}}
\DeclareMathOperator{\zeros}{\matr{0}}
\DeclareMathOperator{\eye}{\matr{I}}
\DeclareMathOperator{\Ac}{\matr{A}_c}
\DeclareMathOperator{\Bc}{\matr{B}_c}
\DeclareMathOperator{\Bcstr}{\matr{B}_c^*}

\DeclareMathOperator{\Gc}{\matr{G}_c}
\DeclareMathOperator{\Jc}{\matr{J}_c}
\DeclareMathOperator{\Jcstr}{\matr{J}_c^*}
\DeclareMathOperator{\F}{\matr{F}}
\DeclareMathOperator{\Qn}{\matr{Q}}
\DeclareMathOperator{\Rn}{\matr{R}}
\DeclareMathOperator{\Pc}{\matr{P}}
\DeclareMathOperator{\GP}{\mathcal{GP}}

\newcommand{\brc}[1]{\left(#1\right)}

\title{A Gaussian process latent force model for joint input-state estimation in linear structural systems}

\author{
  Rajdip Nayek \\
  Department of Civil and Environmental Engineering\\
  University of Waterloo \\
  Ontario, N2L 3G1, Canada\\
  \texttt{rnayek@uwaterloo.ca} \\
   \And
  Souvik Chakraborty \\
  Center for Informatics and Computational Science (CICS)\\
  University of Notre Dame\\
  IN-46556, U.S.A. \\
  \texttt{csouvik41@gmail.com} \\
  \And
  Sriram Narasimhan \\
  Department of Civil and Environmental Engineering\\
  University of Waterloo \\
  Ontario, N2L 3G1, Canada\\
  \texttt{snarasim@uwaterloo.ca} \\
}

\begin{document}
\maketitle

\begin{abstract}
The problem of combined state and input estimation of linear structural systems based on measured responses and a priori knowledge of structural model is considered. A novel methodology using Gaussian process latent force models is proposed to tackle the problem in a stochastic setting. Gaussian process latent force models (GPLFMs) are hybrid models that combine differential equations representing a physical system with data-driven non-parametric Gaussian process models. In this work, the unknown input forces acting on a structure are modelled as Gaussian processes with some chosen covariance functions which are combined with the mechanistic differential equation representing the structure to construct a GPLFM. The GPLFM is then conveniently formulated as an augmented stochastic state-space model with additional states representing the latent force components, and the joint input and state inference of the resulting model is implemented using Kalman filter. The augmented state-space model of GPLFM is shown as a generalization of the class of input-augmented state-space models, is proven observable, and is robust compared to conventional augmented formulations in terms of numerical stability. The hyperparameters governing the covariance functions are estimated using maximum likelihood optimization based on the observed data, thus overcoming the need for manual tuning of the hyperparameters by trial-and-error. To assess the performance of the proposed GPLFM method, several cases of state and input estimation are demonstrated using numerical simulations on a 10-dof shear building and a 76-storey ASCE benchmark office tower. Results obtained indicate the superior performance of the proposed approach over conventional Kalman filter based approaches.
\end{abstract}


\section{Introduction}
	Ordinary differential equations relating the applied input forces to the system state variables are commonly employed in modeling structural dynamic systems. Direct measurements of the system state variables are not always possible; instead, noisy observations as a function of the system state variables are obtained at discrete time instants. State estimation addresses the problem of obtaining optimal estimates of the unobserved system states from noisy measurements given the knowledge of the system model and the dynamic input forces. This problem is at the core of many engineering applications ranging from condition monitoring and prediction of stresses, to feedback control and performance evaluation. The Kalman filter and its variants have been widely used for state estimation of a broad class of both linear and nonlinear dynamical systems \cite{brown1992introduction, grewal2014kalman, sarkka2013bayesian}. A significant challenge in dynamical state estimation arises due to inadequate knowledge of the inputs acting on the system. In many civil engineering systems such as buildings or bridges, measuring the operational external forces (wind loads, vehicular loads, etc.) is not practical. A commonly adopted approach in such cases of unmeasured input forces is to assume that the input force is a zero mean white noise process and then apply filtering techniques for state estimation; however, in many cases the actual inputs may violate this white noise assumption and could potentially yield poor estimation results. Thus, there is a need to inversely determine the unmeasured input forces. Input estimation, also known as force identification, involves reconstructing the latent or unmeasured dynamic input forces given a limited number of sensor measurements and knowledge of the system model \textemdash obtained either from a finite element model or inferred a priori using system identification methods. Cases where input estimation and state estimation are performed together are referred to as joint input-state estimation.
	
	Originally, force identification problems were treated separately from state estimation problems, and the earliest algorithms used frequency response functions to inversely compute the unknown forces \cite{liu2005dynamic, parloo2003force}. Subsequently, a wide variety of force identification methods were proposed, both in time-domain and frequency-domain (see \cite{sanchez2014review} for a recent survey on force reconstruction techniques). Among time-domain approaches, two distinct modelling frameworks exist: deterministic \cite{nordstrom2007strategy, bernal2015sequential} and stochastic \cite{loh2008input, hwang2009estimation, klinkov2012identification}. The stochastic time-domain approaches differ from their deterministic counterparts in the manner that uncertainty in the form of noise is incorporated in the system and measurement model, and the state and/or	input estimates so obtained under stochastic framework are characterized by probability distributions.
	
	Recently, there has been a considerable amount of interest in developing and applying stochastic joint input-state estimation algorithms. Gillijns and De Moor \cite{gillijns2007unbiased}  proposed a minimum-variance unbiased filter for joint input and state estimation of linear systems without a direct transmission term, where the input estimation and state estimation are performed in two independent, sequential steps. This algorithm was extended to include the direct transmission term  \cite{gillijns2007unbiaseddf} which was later applied to structural dynamics by Lourens et al.\ \cite{lourens2012joint}	for use with reduced-order models (i.e., models obtained using a relatively small number of modes). Lourens et al.\ \cite{lourens2012augmented} also proposed an augmented Kalman filter (AKF) that appends the unknown forces to the state vector to jointly estimate the dynamic forces and system states using a classical Kalman filter. This method results in spurious low frequency components in the force and displacement estimates when only acceleration measurements are used. An analytical investigation performed in \cite{naets2015stable} showed that the augmented state-space formulation of AKF is susceptible to numerical instability and un-observability issues when only acceleration measurements are used. In an effort to overcome these shortcomings, Naets et al.\ \cite{naets2015stable}  proposed the use of artificial displacement measurements to stabilize systems with only acceleration measurements. This algorithm, known as AKF with dummy measurements (AKFdm), is similar to the one proposed by Chatzi and Fuggini \cite{chatzi2012structural} for alleviating the low frequency drifts in displacement estimates needed for structural monitoring purposes. Azam et al. \cite{azam2015dual} proposed a dual Kalman filter (DKF) for state and input estimation using only acceleration measurements. Unlike the AKF formulation where the input and states are jointly predicted and updated, the DKF follows a successive structure of implementation, in that, the input prediction and update are followed by the state prediction and update. Due to the separation of the estimation process in two successive steps, the DKF typically overcomes the issue of un-observability.
	Maes et al.\ \cite{maes2016joint} proposed a modified AKF which directly accounts for the correlation between the process noise and the measurement noise.

	An attractive feature of the above Kalman filtering-based joint input-state estimation methods lies in their amenability for online implementation. That said, their accuracy in force reconstruction critically depends on prior tuning of the covariance matrices associated with the unknown inputs. For example, in the AKF and DKF algorithms, the covariance matrix for the unknown input is tuned while in the AKFdm, an additional covariance matrix for the dummy displacements must be tuned in addition to the input covariance. In the absence of expert knowledge of the covariance matrices, a trial-and-error approach or a heuristic rule of thumb is often adopted to tune the input covariance matrices prior to implementation. This may not be possible when there is very little prior knowledge about the input and/or the states of the system. The tacit approach followed in most cases is to have an offline \emph{tuning} phase \textemdash where some response data collected over a certain time duration is used to obtain optimal values of the covariance matrices \textemdash prior to the online estimation phase with Kalman filtering.
	
	In 2009, Alvarez et al.\ \cite{alvarez2009latent} proposed a Bayesian learning technique termed latent force model (LFM), that allows flexible modelling of unknown inputs (or latent forces) in a system. The unknown or latent forces of a system are modelled using non-parametric Gaussian process (GP) model \cite{rasmussen2006gaussian} with some chosen covariance functions. The choice of covariance function provides flexibility to incorporate generic prior knowledge about the behavior of the inputs (e.g. random, periodic, smoothly varying). A Gaussian process based LFM, GPLFM in short, is then constructed incorporating the physics-based mathematical model of the system (e.g.\ differential equations of the system) within the GP covariance functions to infer inputs to the system. The behavior of the latent forces depend on the parameters (also called hyperparameters) of the chosen GP covariance	functions which constitute the tunable parameters in this model. These hyperparameters can be set based on expert knowledge, or they can be tuned by optimization using observed data.   
	Hartikainen and S\"{a}rkk\"{a} \cite{hartikainen2011sequential} and more recently S\"{a}rkk\"{a} et al.\ \cite{sarkka2017gaussian} advocated a state-space approach to GPLFM inference wherein all information about the system and the GP force model is fully captured in an augmented state-space representation, thus making the inference amenable to use with 	Bayesian filtering and smoothing methods. Once the hyperparameters of GPLFM are computed, a GPLFM can be adapted for online implementation with Kalman filter.
	
    This paper introduces GPLFM for joint input and state estimation of linear time-invariant structural systems. To the authors' knowledge, this is the first application of GPLFM in the domain of structural dynamics. Additionally, the following are the key contributions of this paper:
    \begin{enumerate}
        \item The GPLFM is shown to be a generalization of the class of input-augmented state-space models underlying some popular Kalman filter-based joint input-state estimation approaches \cite{lourens2012augmented, naets2015stable, maes2016joint}. The augmented state-space model used in AKF \cite{lourens2012augmented} is obtained as a special case of GPLFM.
        
        \item Unlike the AKF, the GPLFM formulation is analytically proved to be observable and is shown to be robust against drift in force estimation even when used with only acceleration measurements. 
        \item A maximum likelihood optimization based on the observed data is used to compute the optimal hyperparameters of the GPLFM, which eliminates the need for tuning the parameters heuristically or through trial-and-error.
    \end{enumerate}
    An in-depth numerical study is also presented comparing the performance of the proposed GPLFM algorithm with the existing AKF, AKFdm, and DKF algorithms. 
    
    The paper is organized as follows. Section \ref{sec:probform} presents the mathematical model of the structure and outlines the objective of this study. Next, a concise background on GPLFM and its use in regression is provided in Section \ref{sec:GPLFM_back}. In Section \ref{sec:GPLFM_ssm_inference}, the GPLFM is formulated into a linear state-space model, and a procedure for joint posterior inference on the inputs and states is discussed. Section \ref{sec:GPLFMdiscussion} presents GPLFM as a generalization of the augmented state-space models used in joint input-state estimation and provides proofs of observability and stability of the GPLFM. Section \ref{sec:numerical} encompasses numerical studies using a 10-dof shear building, illustrating the performance of GPLFM under different excitation and measurement scenarios. The paper is finally concluded with an application of the proposed method on a 76-storey ASCE benchmark office tower for joint input and state estimation.
	
	\section{Problem Formulation} \label{sec:probform}
	\subsection{Mathematical model of structural system}
    The equation of motion of a $\n$ degrees-of-freedom (dofs) linear structural system under forced excitation including ground motions can be represented by the following second order ordinary differential equation (obtained after discretization in space):
    	\begin{equation} \label{eq:eom}
    	\matr{M}\ddot{\ve{u}}(t) + \matr{C}\dot{\ve{u}}(t) + 
    	\matr{K}\ve{u}(t) =  \matr{S}_p \p(t) - \matr{M} \matr{S}_g 
    	\udd^g(t), 
    	\end{equation} 
    where, $\ve{u}(t) \in \R^{\n}$ is the vector of displacements at the degrees of freedom and $\matr{M}$, $\matr{C}$ and $\matr{K} \in \mathbb{R}^{\n \times \n}$ represent the mass, damping, and stiffness matrices of the structural	system, 	respectively. The external forces acting on the structure are represented by a combination of the external loads acting on the structure $\matr{S}_p \p(t)$ and the forces generated due to earthquake ground motion $\matr{M} \matr{S}_g \udd^g(t)$. The first term denotes the product of a load influence matrix $\matr{S}_p \in \mathbb{R}^{\n \times \np}$ and the vector $\p(t) \in \R^{\np}$ representing the $\np$ external load time histories while the second term represents the forces generated in the structure due to earthquake ground motion $\ve{u}^g(t) \in \R^{n_g}$ where $\matr{S}_g \in \R^{\n \times n_g}$ is the matrix of influence coefficients which specifies the dofs affected by the ground motion. In cases where the size of the structural model is very large, it is common to use modal reduced-order models.
    
    Defining a state vector $\x(t)$ and a force vector $\f(t)$ as 
    \begin{equation} \label{eq:defn}
    	\x(t) = \begin{bmatrix}
    	\ve{u}(t) \\ \dot{\ve{u}}(t)
    	\end{bmatrix}, \quad
    	\f(t) = \begin{bmatrix}
    	\p(t) \\ \udd^g(t)
    	\end{bmatrix},
    \end{equation}
    where $\x(t) \in \R^{\ns}$ and $\f(t) \in \R^{n_f}$; $\ns = 2\n$ and $n_f = n_p + n_g$, the set of second-order differential equations in Equation \ref{eq:eom} can be converted to a set of first-order differential equations called as the continuous-time state-space equation
    \begin{equation} \label{eq:processeqn}
    	\dot{\x}(t) = \Ac \x(t) + \Bc \f(t) 
    \end{equation} 
    where the continuous-time system matrices $\Ac \in \R^{\ns \times \ns}$ and $\Bc \in \R^{\ns \times \nf}$ are defined as 
    \begin{equation}
    	\Ac = \begin{bmatrix}
    	\zeros & \eye \\ -\matr{M}^{-1} \matr{K} & -\matr{M}^{-1} \matr{C}
    	\end{bmatrix}, \quad 
    	\Bc = \begin{bmatrix}
    	\zeros & \zeros \\ \matr{M}^{-1} \matr{S}_p & -\matr{S}_g
    	\end{bmatrix} ,
    \end{equation}
    where $\eye$ is the identity matrix of appropriate dimension. Consider next the measurement equation, where it is assumed that a combination of displacements, velocities and accelerations can be measured. Hence the output vector $\y(t)$, containing $\no$ measured quantities, assumes the following form
    \begin{equation} \label{eq:out}
    	\y(t) = \begin{bmatrix}
    	\matr{S}_{dis} & \zeros & \zeros \\
    	\zeros & \matr{S}_{vel} & \zeros \\
    	\zeros & \zeros & \matr{S}_{acc}
    	\end{bmatrix} 
    	\begin{bmatrix}
    	\ve{u}(t) \\ \dot{\ve{u}}(t) \\ \ddot{\ve{u}}(t)
    	\end{bmatrix},
    \end{equation}
    where, $\matr{S}_{dis}$, $\matr{S}_{vel}$ and $\matr{S}_{acc}$ are the 	selection matrices for displacements, velocities and accelerations, respectively. Using Equation \ref{eq:eom} and the defined state and force vectors, Equation \ref{eq:out} can be written in the state-space form as
    \begin{equation} \label{eq:obseqn}
    	\y(t) = \Gc \x(t) + \Jc \f(t),
    \end{equation}
    where, the output influence matrix $\Gc \in \R^{\no \times \ns}$ and direct transmission matrix $\Jc \in \R^{\no \times \nf}$ are 
    \begin{equation}
    	\Gc = \begin{bmatrix}
    	\matr{S}_{dis} & \zeros \\
    	\zeros & \matr{S}_{vel} \\
    	-\matr{S}_{acc} \matr{M}^{-1} \matr{K} & -\matr{S}_{acc} \matr{M}^{-1}
    	\matr{C}
    	\end{bmatrix}, \quad 
    	\Jc = \begin{bmatrix}
    	\zeros & \zeros \\ \zeros & \zeros \\ \matr{S}_{acc} \matr{M}^{-1} 
    	\matr{S}_p & \zeros
    	\end{bmatrix}.
    \end{equation}
    Note that the second column of the direct transmission term $\Jc$ corresponds to the earthquake ground motion input and takes zero values as the direct contribution of earthquake ground motion vanishes when measuring absolute accelerations (see derivation in \ref{sec:feedthru_derive})
	
	Equations \ref{eq:processeqn} and \ref{eq:obseqn} form the continuous-time state-space model (SSM) for the system described by Equation \ref{eq:eom}.
	In practical applications, continuous time outputs $\y(t)$ are of course not observed, instead noisy measurements are obtained via sampling of system responses at discrete time instants. Assuming a sampling interval of $\Delta t$, discrete time instances are defined at $t_k = k \Delta t$ for $k=1,\ldots,N$, and a discrete measurement model can be expressed as 
	\begin{equation} \label{eq:contdismeas}
	\y(t_k) = \matr{G}_c \x(t_k) + \matr{J}_c \f(t_k) + \vn_k,
	\end{equation}
	where $\vn_k$ is the measurement noise vector.
	Combined together, Equations \ref{eq:processeqn} and \ref{eq:contdismeas} represent a SSM with continuous-time dynamics and discrete-time measurements known as the continuous-discrete state-space model (see \cite{jazwinski1970stochastic}, pg170 or \cite{sarkkabook2019}, pg93)
	\begin{align} \label{eq:contdiscssm}
	\begin{split}
	\dot{\x}(t) &= \Ac \x(t) + \Bc \f(t), \\
	\y(t_k) &= \matr{G}_c \x(t_k) + \matr{J}_c \f(t_k) + \vn_k,
	\end{split}
	\end{align} 
	For numerical implementation, the continuous-time state-space form in Equations \ref{eq:processeqn} and \ref{eq:obseqn} is converted to the discrete-time state-space following the zero-order-hold assumption 
	\begin{align} \label{eq:dssm}
	\begin{split}
	\x_k &= \matr{A} \x_{k-1} + \matr{B} \f_{k-1} + \wn_{k-1},\\
	\y_{k} &= \matr{G} \x_{k} + \matr{J} \f_{k} + \vn_k,
	\end{split}
	\end{align}
	where
	$\matr{A} = \exp(\Ac \Delta t)$, $\matr{B} = [\matr{A} - \eye] 
	\Ac^{-1} \Bc$, $\matr{G} = \Gc$, $\matr{J} = \Jc$.
	The terms $ \wn_{k-1}$ and $\vn_k$ are the process and the measurement noise vectors, typically added to to account for modelling errors and measurement errors, respectively. They are assumed zero-mean Gaussian white noise with covariances
	\begin{equation} \label{eq:noisecov}
	\E{\begin{Bmatrix}
	\wn_{k}\\ \vn_k
	\end{Bmatrix} \begin{Bmatrix}
	\wn^T_l & \vn^T_l
	\end{Bmatrix}} = \begin{bmatrix}
	\Qn^x & \zeros \\ \zeros^T & \Rn
	\end{bmatrix} \delta_{kl},
	\end{equation} 
	with $\Rn \succ \zeros $ and $\Qn^x \succcurlyeq \zeros$.
	
	\subsection{Objective} \label{sec:objective}
	Having elaborated on the mathematical model of structure of interest, attention is now focused on the problem of stochastic joint input and state estimation. Here, both the inputs $\f_k$ and the system states $\x_k $ are taken to be unknown sequence of Gaussian random variables, and the goal lies in inferring the joint posterior distribution of the inputs and states from a set of noisy measurements $\y_k$ given the a priori knowledge of the structural model parameters. In pursuit of this goal, the objective of this paper is to implement a Gaussian process latent force model which improves upon existing methods by reducing the dependency on manual tuning of covariance matrices associated with the unknown inputs and also provides numerical stability with respect to observability of the augmented state-space formulation when used with only acceleration measurements.
	The location of the inputs is assumed to be known, however studies on input localization can be found in \cite{rezayat2016identification, aucejo2017multiplicative, kirchner2018exploiting}. 
	
	\section{Background on Gaussian process latent force models} \label{sec:GPLFM_back}
	\subsection{Brief overview of Gaussian processes}
	Gaussian processes (GPs) \cite{hagan1978curve, rasmussen2006gaussian, kocijan2016modelling} are a class of stochastic processes that 	provide a paradigm for specifying probability distributions over functions. Gaussian processes have been widely studied and used in many different 	fields such as signal processing \cite{PrezCruz2013GaussianPF}, geostatistics \cite{chiles2012, cressie1993spatial} and inverse problems \cite{kocijan2016modelling}. For instance, in geostatistics literature Gaussian process regression is known as \emph{Kriging}. A Gaussian process is a generalization of the Gaussian probability	distribution. While a probability distribution is defined over random 
	variables which are scalars or vectors (for multivariate distributions), a 	stochastic process governs the properties of functions. 
	Consider an independent variable $\ve{\ell} \in \R^d$ and a function $g(\ve{\ell})$  such that $g: \R^d \rightarrow \R$. Then a GP defined over $g(\ve{\ell})$ with the mean $\mu(\ve{\ell})$ and covariance function $\kappa(\ve{\ell},\ve{\ell}';\tcf)$ is as follows
	\begin{equation} \label{eq:gpdefn}
	g(\ve{\ell})  \sim \GP \left(\mu(\ve{\ell}), \kappa(\ve{\ell},\ve{\ell}';\tcf) 
	\right),
	\end{equation}
	\begin{align}
	\begin{split}
	\mu(\ve{\ell}) &= \E{ g(\ve{\ell}) },\\
	\kappa(\ve{\ell},\ve{\ell}';\tcf) &= \E{ \left(g(\ve{\ell}) - 
	\mu(\ve{\ell})\right) 
	\left(g(\ve{\ell}') - \mu(\ve{\ell}')\right) },
	\end{split}
	\end{align}
	where $\tcf$ denotes the hyperparameters of the covariance function $\kappa$.
	The choice of the covariance function $\kappa$ allows encoding any prior knowledge about $g(\ve{\ell})$ (e.g., periodicity, linearity, smoothness), and can accommodate approximation of arbitrarily complex functions \cite{rasmussen2006gaussian}. The notation in Equation \ref{eq:gpdefn} implies that any finite collection of function values has a joint multivariate Gaussian distribution, that is $\left(g(\ve{\ell}_1), g(\ve{\ell}_2), \ldots , g(\ve{\ell}_N)\right) \sim \mathcal{N}(\ve{\mu}, \matr{K})$, where $\ve{\mu} = 	\left[\mu({\ve{\ell}_1}),\ldots,\mu({\ve{\ell}_N})  \right]^T$ is the mean vector and $\matr{K}$ is the covariance matrix with $\matr{K}(i,j) = \kappa(\ve{\ell}_i, \ve{\ell}_j)$ for $i, j = 1, 2, \ldots, N$. If no prior information is available about the mean function, it is generally set to zero, i.e.\ $\mu(\ve{\ell}) = 0$. However, for the covariance function, any function $\kappa(\ve{\ell}, \ve{\ell}')$ that generates a positive, semi-definite, covariance matrix $\matr{K}$ is a valid covariance function. For example, the following squared exponential covariance function has the form
	\begin{equation}
	\kappa(\ve{\ell},\ve{\ell}') = \sigma_g^2 \exp \left[- \sum_{k=1}^{d} 
	\frac{\left(\ve{\ell}(k) - \ve{\ell}'(k)\right)^2}{2 l_k^2} \right],
	\end{equation}
	where $\tcf = \left\{\sigma_g, l_1,\ldots,l_d \right\}$ are the hyperparameters of the covariance function.	The	mean and covariance functions dictate how the random functions $g(\ve{\ell})$	behave on average and how the different points in the input	space vary with respect to each other. The covariance function thus encodes a correlation structure which introduces inter-dependencies between function values at different inputs.	

	\subsection{Gaussian process regression}
	GPs are often used to solve regression problems \cite{williams1996gaussian, diazdelao2010structural, moustapha2018comparative}. In general, regression problems are concerned with the estimation of values of a dependent variable $g(\ve{\ell})$ observed at certain values of an independent variable $\ve{\ell}$, given a set of noisy measurements $y$. The relation can be written as 
	\begin{equation} \label{eq:regrsn}	
	y = g(\ve{\ell}) + \nu,
	\end{equation}
	where $\nu$ is noise. In other words, regression involves estimating the latent (unobserved) function $g(\ve{\ell})$ that will enable prediction of $\hat{y}$ at new values of $\ve{\ell}$.
	
	One	perspective of looking at GPs representing the input-output relation is called the function-space view, given in Rasmussen and Williams \cite{rasmussen2006gaussian}. Unlike traditional modelling approaches which rely on fitting a parameterized mathematical form to approximate the input-output relation $g(\ve{\ell})$, a GP does not assume any explicit form of $g(\ve{\ell})$, rather a prior belief (in the form of the mean function and the covariance function) is placed on the space of all possible functions $g(\ve{\ell})$. GPs belongs to the class of `non-parametric' models because the number of parameters in the model is not fixed, but rather determined by the number of data points. Upon data collection, the posterior distribution over $g(\ve{\ell})$ is updated according to Bayes' rule. All the candidate functions represented by GPs can be used to express linear or nonlinear relationships between the input $\ve{\ell}$ and the output $y$. 
	
	It is to be remarked at this point that the input dimension $d$ (dimension	of the independent variable $\ve{\ell}$) is usually greater than 1, however in 	this work, the primary focus is on doing inference using time-series data 	where time is the only independent input variable. Thus $d=1$, and 	$\ve{\ell}$ is hereafter replaced by $t$ --- referred to as the time domain. 	Examples where input dimension is greater that one ($d>1$) are situations	where the input variables may represent time and one-dimensional space ($d 	= 2$), or space in two dimensions ($d = 2$), or even time with 	two-dimensional space ($d = 3$).
	
	Now consider the prediction of the value of $g(t)$ at a test point $t_*$, 	based on a previously measured set of outputs $\mathcal{D} = \left\{ (t_1,y_1), \ldots,(t_N, y_N) \right\}$ which are corrupted with white Gaussian noise:
	\begin{equation}
	y_k = g(t_k) + \nu_k, \;\; \nu_k \sim \mathcal{N}\left(0, \sigma_{n}^2 
	\right).
	\end{equation} 
	Assume 	that $g(t)$ is modelled as a Gaussian process with zero mean and a 	covariance function, that is
	\begin{equation}
	g(t) \sim \GP(0,k(t,t';\tcf)),
	\end{equation} 
	where $k(t,t';\tcf)$ is the covariance function with hyperparameters $\tcf$.
	Under the Gaussian process assumption, the 	joint distribution between $y_1,\ldots,y_N$ and $g(t_*)$ is Gaussian, and can be expressed as
	\begin{equation}
	\begin{bmatrix}
	\y \\ g(t_*)
	\end{bmatrix} \sim \mathcal{N}
	\left(
	\begin{bmatrix}
	\zeros \\ 0
	\end{bmatrix}, \begin{bmatrix}
	\matr{K}(\ve{t},\ve{t}; \tcf) + \sigma^2_n \eye & \ve{k}(\ve{t},t_*; 
	\tcf) \\ 
	\ve{k}^T(\ve{t},t_*; \tcf) & k(t,t_*; \tcf)
	\end{bmatrix}
	\right),
	\end{equation}
	where $\ve{t} = \left[t_1, \ldots, t_N\right]^T$ is the vector consisting  	of time	points, $\y = \left[y_1, \ldots, y_N\right]^T$ is a vector of measured outputs, $\matr{K}(\ve{t},\ve{t}; \tcf)\in \R^{N \times N}$ is the covariance matrix comprising elements $\left\{k (t_i,t_j; \tcf) \right\}_{i,j=1}^N$, and $\ve{k}(\ve{t},t_*; \tcf) \in \R^n$ is a vector with elements $\left\{k (t_i,t_*;\tcf) \right\}_{i=1}^N$.
	
	Using the properties of multivariate Gaussian distributions, it can be 	shown that the posterior distribution of $g(t_*)$ given the dataset $\y$ 	and hyperparameters $\tcf$ is also Gaussian 
	\begin{equation}
	p(g(t_*)|\y, \tcf) = \mathcal{N} \left(g(t_*) \; | \; 
	\mu_{\GP}(t_*),  
	\sigma^2_{\GP}(t_*)  \right),
	\end{equation}
	with mean and variance
	\begin{align}
	\begin{split}
	\mu_{\GP}(t_*) &= \ve{k}^T(\ve{t}, t_*; \tcf) 
	\left[\matr{K}(\ve{t},\ve{t}; \tcf) + \sigma^2_n \eye\right]^{-1} 
	\y,\\
	\sigma^2_{\GP}(t_*) &= k(t_*,t_*; \tcf) -  \ve{k}^T(\ve{t}, 
	t_*; \tcf) 
	\left[\matr{K}(\ve{t},\ve{t}; \tcf) + \sigma^2_n \eye\right]^{-1} 
	\ve{k}(\ve{t}, 
	t_*; \tcf).
	\end{split}
	\end{align}
	\begin{figure}[htbp!]
		\centering
		\includegraphics[scale=0.5]{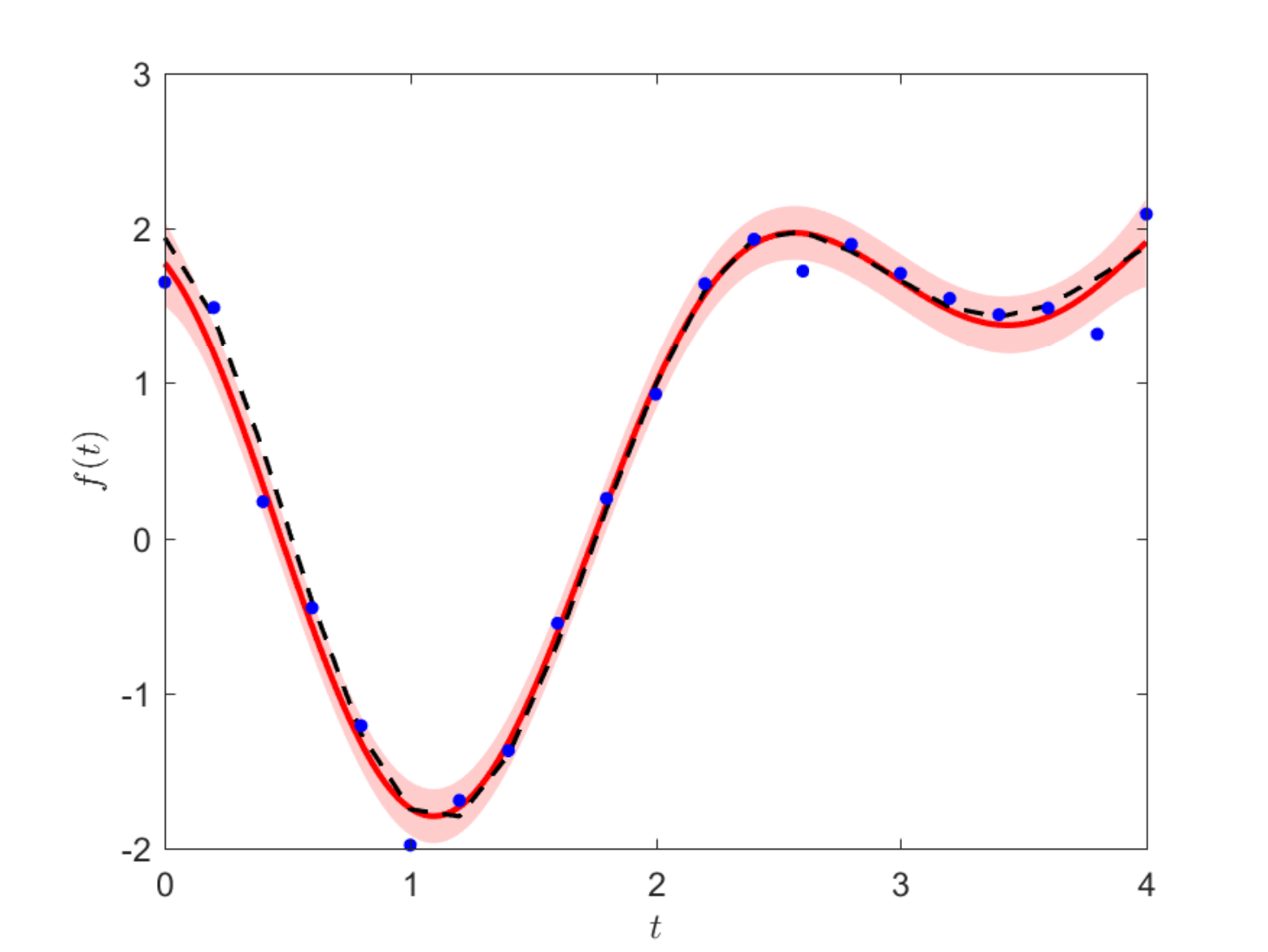}
		\caption{Example of Gaussian process regression. The dashed line represents the unknown function $g(t)$. The dots are the measured noisy data $\{t_k,y_k\}_{k=1}^N$. The solid line is the posterior mean $\mu_{\GP}(t)$, and the shaded region represents two standard deviations ($\sigma^2_{\GP}$) away from the mean}
		\label{fig:GPRdemo}
	\end{figure}
	
	\subsection{Latent force models using Gaussian processes}
	Latent force models \cite{alvarez2009latent, alvarez2013linear} are a hybrid approach of modelling which combines \textit{mechanistic} models (i.e.\ models based on physical laws) with \textit{non-parametric} components such as GPs.	Consider the following linear continuous-time state-space equations obtained from Equations \ref{eq:processeqn} and \ref{eq:contdismeas} 
	\begin{align} \label{eq:lfmprocess}
	\begin{split}
	\dot{\x}(t) &= \Ac \x(t) + \Bc \f(t), \\
	\y(t_k) &= \matr{G}_c \x(t_{k}) + \matr{J}_{c} \f(t_{k}) + \vn_k,
	\end{split}
	\end{align}
	with all components of the unknown input vector $\f(t)$ being modelled as zero-mean independent Gaussian processes  
	\begin{equation} \label{eq:gpinput}
	f^{(j)}(t) \sim \GP (0, \kappa_{j}(t,t')) ,
	\end{equation}
	Here $f^{(j)}(t)$ is the $j$th component of $\f(t)$, $j = 1,\ldots,\nf$. The assumption of zero mean leads to no loss of generality. Note that the hyperparameters $\tcf$ of the covariance 
	function have been omitted for notational compactness. The model, given by Equations \ref{eq:lfmprocess} and \ref{eq:gpinput}, leads to a (linear) latent force model, in which the basic idea is to combine the mechanistic state-space model with unknown (latent) forcing functions $\f(t)$ modelled by non-parametric Gaussian processes. 
	
	The solution of the state in the differential equation in Equation \ref{eq:lfmprocess} can be obtained as
	\begin{equation} \label{eq:GPx}
	\x(t) = \bm{\Psi}(t) \x_0 + \int_0^t \bm{\Psi}(t-s) \Bc \f(s) ds  ,
	\end{equation}
	where $\bm{\Psi}(\tau) = e^{\Ac \tau}$ denotes the matrix exponential and $\x_0$ is the initial condition of state at time $t = 0$. The state covariance matrix function $\matr{K}_{xx}$ can then be calculated as
	\begin{equation} \label{eq:Kxx}
	\matr{K}_{xx}(t,t') = \E{\x(t) \x(t')^T} = \bm{\Psi}(t) 
	\Pc_0^{x} \bm{\Psi}(t)^T + 
	\int_{0}^{t'} \int_{0}^{t} \bm{\Psi}(t-s) \Bc \matr{K}_{ff}(s,s') \Bc^T 
	\bm{\Psi}(t'-s')^T ds ds',
	\end{equation}
	where $\Pc_0^{x}$ is the prior covariance matrix of $\x(t)$ and $\matr{K}_{ff}(t,t')$ is the joint covariance matrix of all latent forcing functions between time instants $t$ and $t'$ 
	\begin{equation} \label{eq:Kff}
	\matr{K}_{ff}(t,t') =  \E{\f(t) \f(t')^T} = \text{diag} 
	\left[\kappa_{1}(t,t'), \ldots, \kappa_{n_f}(t,t') \right] .
	\end{equation} 
	$\matr{K}_{ff}(t,t')$ is a diagonal matrix due the assumption of independence across all force components. Observe that the states $\x(t)$ form a multi-dimensional GP
	\begin{equation}
	\x(t) \sim \GP(\zeros, \matr{K}_{xx}(t,t')).
	\end{equation} 
	and thus both $\f(t)$ and $\x(t)$ are GPs. Next the measurement equation is rewritten as 
	\begin{align}
	\y_k = \g(t_k) + \vn_k
	\end{align}
	with
	\begin{equation} \label{eq:gfun}
	\g(t_k) = \matr{G}_c \x(t_k) + \matr{J}_c \f(t_k) = 
	\underbrace{\begin{bmatrix}	\matr{G}_c & \matr{J}_c 
		\end{bmatrix}}_{\matr{G}_a}
	\underbrace{\begin{bmatrix}	\x(t) \\ \f(t) \end{bmatrix}}_{\etav(t)} = 
	\matr{G}_a \etav(t).
	\end{equation} 
	Since the linear combination of Gaussian processes is a Gaussian process, $\g(t)$ is a Gaussian process, and can be expressed as
	\begin{equation}
	\g(t) \sim \GP (\zeros, \matr{K}_{gg}(t,t')),
	\end{equation}
	where
	\begin{subequations}
		\begin{equation}
		\matr{K}_{gg}(t,t') = \matr{G}_a \matr{K}_{\eta \eta}(t,t') 
		\matr{G}_a^T ,
		\end{equation}
		\begin{equation}
		\matr{K}_{\eta \eta}(t,t') = \begin{bmatrix}
		\matr{K}_{xx}(t,t') & \matr{K}_{xf}(t,t') \\
		\matr{K}^T_{xf}(t,t') & \matr{K}_{ff}(t,t')
		\end{bmatrix} ,
		\end{equation}
		\begin{equation} \label{eq:Kxf}
		\matr{K}_{xf}(t,t') = \E{ \x(t) \f(t')^T } = 
		\int_{0}^{t} \bm{\Psi}(t-s) \Bc \matr{K}_{ff}(s,t') ds ,
		\end{equation}
	\end{subequations}
	and $\matr{K}_{ff}(t,t')$ and $\matr{K}_{xx}(t,t')$ are defined by Equations \ref{eq:Kff} and \ref{eq:Kxx} respectively. A derivation for $\matr{K}_{xf}(t,t')$ can be found in \ref{sec:Kfx_derivation}. Thus, one can replace the problem defined by Equations \ref{eq:lfmprocess}, and \ref{eq:gpinput} with an equivalent Gaussian process regression problem
	\begin{align}\label{eq:gpr}
	\begin{split}
	\y_k &= \g(t_k) + \vn_k, \\
	\g(t) &\sim \GP (\zeros, \matr{K}_{gg}(t,t')).
	\end{split}
	\end{align}
	This approach, introduced by Alvarez et al.\ \cite{alvarez2009latent, alvarez2013linear} reduces the latent force model in Equation \ref{eq:lfmprocess} and \ref{eq:gpinput} to a GP regression problem as expressed by Equation \ref{eq:gpr}. However, the approach requires analytical computation of the covariance matrix functions $\matr{K}_{xx}$, which may not always be obtained in a closed form and may need computation via numerical integration. In the next section, a formulation to convert covariance functions to state-space form for a certain class of GP covariance functions is considered. In that formulation, numerical integration is avoided altogether and replaced with matrix exponential computations and solutions to Lyapunov equations.
	
	\section{GPLFM as stochastic SSMs for sequential inference} \label{sec:GPLFM_ssm_inference}
	This section discusses how temporal GPs with a certain class of covariance functions can be reformulated as SSMs and demonstrates how they can be combined with linear LFMs for sequential posterior inference of latent inputs and states. It must be emphasized at this point that only stationary covariance functions which have the form $\kappa(t,t') = \kappa(\tau)$, where $\tau = t-t'$, are considered in this work. 
	
	\subsection{State space representation of temporal GPs} \label{sec:GPss}
	Consider a scalar-valued GP with stationary covariance function as follows
	\begin{equation} \label{eq:gpform}
	h(t) \sim \GP(0, \kappa(\tau)).
	\end{equation} 
	It has been shown \cite{hartikainen2010kalman, hartikainen2011sequential, sarkka2013spatiotemporal, sarkka2014convergence} that $h(t)$ can be represented as the output of a scalar linear time-invariant system driven by white noise  
	\begin{equation} \label{eq:scalarlti}
	\frac{d^m h(t)}{dt^m} + a_{m-1} \frac{d^{m-1} h(t)}{dt^{m-1}} + \ldots + 
	a_1 \frac{dh(t)}{dt} + a_0 h(t) = w(t),
	\end{equation}
	where $a_0,\ldots,a_{m-1}$ are known constants and $w(t)$ is a white noise process with spectral density $S_w (\omega) = \sigma_w$. Equation \ref{eq:scalarlti} can be written in state-space form
	\begin{align} \label{eq:gpssm}
	\begin{split}
	\dot{\z}(t) &= \F_{cf} \z(t) + \matr{L}_{cf} w(t),\\
	h(t) &= \matr{H}_{cf} \z(t),
	\end{split}
	\end{align}
	where the state $\z(t) = \begin{pmatrix} h(t) & \frac{dh(t)}{dt} & \ldots & \frac{d^{m-1}h(t)}{dt^{m-1}} \end{pmatrix}^T$ and the matrices $\F_{cf} \in \R^{m \times m}$, $\matr{L}_{cf}	\in \R^{m \times 1}$ and $\matr{H}_{cf} \in \R^{1\times m}$ are given by:
	\begin{equation}
	\F_{cf} = \begin{bmatrix}
	0 & 1 & 0 & \ldots & 0 \\
	0 & 0 & 1 & \ldots & 0 \\
	\vdots & \vdots & \vdots & \ddots & \vdots \\
	0 & 0 & 0 & \ldots & 1 \\
	-a_0 & -a_1 & \ldots & \ldots & -a_{m-1}
	\end{bmatrix},\quad 
	\matr{L}_{cf} = \begin{bmatrix}
	0 \\ 0 \\ \vdots \\ 0 \\ 1
	\end{bmatrix}, \quad
	\matr{H}_{cf} = \begin{bmatrix}
	1 & 0 & \ldots & 0 & 0
	\end{bmatrix}.
	\end{equation}
	The discrete-time form of the LTI model in Equation \ref{eq:gpssm} is given by 
	\begin{align} \label{eq:gpcfssm}
	\begin{split}
	\z_{k} &= \F x_{k-1} + \matr{L} w_{k-1},\\
	h_{k} &= \matr{H} \z_k,
	\end{split}
	\end{align} 
	where $\F = e^{\F_{cf} \Delta t}$, $\matr{L} = 	(\F-\eye) \F_{cf}^{-1} \matr{L}_{cf}$ and $\matr{H} = \matr{H}_{cf}$. This formulation allows GP regression problems to be solved sequentially with Kalman filtering and smoothing techniques. The iteration can be started with initial state $\z_0 \sim \mathcal{N}(\zeros, \Pc^{cf}_0)$. Note that the model (and the corresponding covariance function) is stationary, and therefore $\z_k$ has a steady state solution distributed as $\z_k \sim \mathcal{N}(\zeros, \Pc^{cf}_{\infty})$, where the steady state covariance matrix $\Pc^{cf}_{\infty}$ can be obtained as the solution of continuous-time Riccati solution
	\begin{equation} \label{eq:ctre}
	\frac{d \Pc^{cf}_{\infty}}{dt} = \F_{cf} \Pc^{cf}_{\infty} 
	+ 
	\Pc^{cf}_{\infty} \F_{cf}^T +  \matr{L}_{cf} \sigma_{w} 
	\matr{L}_{cf}^T.
	\end{equation} 
	Since steady state is invariant with respect to time, the initial covariance matrix is set equal to the steady state covariance, $\Pc^{cf}_0 \coloneqq \Pc^{cf}_{\infty}$.
	
	Using the SSM given by Equation \ref{eq:gpssm}, the power spectral density of $h(t)$ can be computed as 
	\begin{equation} \label{eq:spfactex}
	S_h(\omega) = \underbrace{\matr{H} \; (\F_{cf} +	i\omega \eye)^{-1} 
		\matr{L}_{cf}}_{H(i \omega)}\; 
	\sigma_w \; \underbrace{\matr{L}_{cf}^T (\F_{cf} +	i\omega 
		\eye)^{-T} 
		\matr{H}^T}_{H(-i \omega)}.
	\end{equation}
	The stationary covariance function for $h(t)$ is related to its spectral density through the inverse Fourier transform
	\begin{equation} \label{eq:cfFT}
	\kappa(\tau) = \mathcal{F}^{-1}[S_h(\omega)] = \frac{1}{2 \pi} 
	\int_{-\infty}^{\infty} S_h(\omega) e^{i\omega 
		\tau} d \omega.
	\end{equation}
	In terms of the state-space matrices, $\kappa(\tau)$ 
	can be calculated using \cite{solin2014explicit}
	\begin{equation}
	\kappa(\tau) = 
	\begin{cases}
	\matr{H} \Pc^{cf}_{\infty} \bm{\Phi}(\tau)^T \matr{H}^T, & \text{if } 
	\tau 
	\ge 0 \\
	\matr{H} \bm{\Phi}(-\tau) \Pc^{cf}_{\infty} \matr{H}^T, & \text{if } 
	\tau<0
	\end{cases},
	\end{equation}
	where $\bm{\Phi}(\tau) = e^{\F_{cf} \tau}$.
	
	Hartikainen and S\"{a}rkk\"{a} \cite{hartikainen2010kalman} showed that it is possible to form $\F_{cf}$, $\matr{L}_{cf}$ and $S_h(\omega)$ such that $h(t)$ has the desired covariance function $\kappa(\tau)$ only if the spectral density of $h(t)$ has a proper rational form
	\begin{equation} \label{eq:rat}
	S_h(\omega) = \frac{p\text{th order polynomial in }\omega^2}{q\text{th 
			order polynomial in }\omega^2}, \quad	(p < q).
	\end{equation}
	By applying spectral factorization, the spectral density $S_h(\omega)$ can 	be expressed similar to Equation \ref{eq:spfactex}
	\begin{equation} \label{eq:tfact}
	S_h(\omega) = H(i\omega) \sigma_w H(-i\omega),
	\end{equation}
	where $H(i\omega)$ is a stable rational transfer function -- has poles in the left half plane. The procedure to convert a covariance function $\kappa(\tau)$ into SSM is outlined as follows \cite{sarkka2013spatiotemporal}:
	\begin{itemize}
		\item The spectral density $S_h(\omega)$ is computed by taking 	Fourier transform of $\kappa(\tau)$
		\item Spectral factorization of $S_h(\omega)$ is used to find a stable rational transfer function $H(i \omega)$ and $\sigma_w$ (as shown in Equation \ref{eq:tfact}). If $S_h(\omega)$ is not rational as shown in Equation \ref{eq:rat}, then an approximation is used to represent it in a rational form.
		\item Finally, the transfer function model is converted into an equivalent SSM driven by a scalar-valued white noise process with spectral density $\sigma_w$ 
	\end{itemize} 
	Many GPs having stationary covariance functions can be expressed into state-space form as in Equation \ref{eq:gpcfssm} with model matrices defined by $\F_{cf}$, $\matr{L}_{cf}$, $\matr{H}$, $\sigma_w$ and $\Pc^{cf}_0$. It should be noted that non-stationary covariance functions can also be converted into state-space forms \cite{benavoli2016state}, however the Fourier transform approach is no longer applicable.
	
	\subsection{GP-LFM in joint state-space form}
	Consider once again the latent force model given by Equations \ref{eq:lfmprocess} and \ref{eq:gpinput}, rewritten for convenience,
	\begin{align}
	\begin{split}
	\dot{\x}(t) &= \Ac \x(t) + \Bc \f(t), \\
	f^{(j)}(t) & \sim \GP (0, \kappa_{j}), \; j = 1,\ldots,\nf, \\
	\y(t_k) &= \matr{G}_c \x(t_{k}) + \matr{J}_c \f(t_{k}) + \vn_k,
	\end{split}
	\end{align}
	where each component, $f^{(j)}(t) \in \R$, of the latent force vector $\f(t)$ is modelled as a GP. For each component, given a \textit{stationary} covariance function $\kappa_{j}(\tau; \tcf_{j})$ with hyperparameters $\tcf_{j}$, a state-space representation can be constructed using the above procedure. The state-space representation for the $j$th component of $\f(t)$, is given by 
	\begin{align} \label{eq:augsslong}
	\begin{split}
	\dot{\z}^{(j)}(t) &= \F^{(j)} \z^{(j)}(t) + \matr{L}^{(j)} 
	w^{(j)}(t),\\
	f^{(j)}(t) &= \matr{H}^{(j)} \z^{(j)}(t),
	\end{split}
	\end{align}
	where $\z^{(j)}(t) \in \R^{m_j}$ is the state vector for $j$th latent force, $w^{(j)}(t)$ is a zero mean Gaussian process with spectral density $\sigma_w^{(j)}$, and $\F^{(j)} \in \R^{m_j \times m_j}$, $\matr{L}^{(j)} \in \R^{m_j \times 1}$,  $\matr{H}^{(j)} \in \R^{1 \times m_j}$ are the state-space matrices with $m_j$ being the dimension of the state vector $\z^{(j)}(t)$, $j=1,\ldots,\nf$. Formally, the linear time-invariant system represented in Equation \ref{eq:augsslong} is a stochastic differential equation where $w^{(j)}(t)$ represents a Brownian motion, and the mathematical description is given via stochastic integrals (see e.g. \cite{jazwinski1970stochastic} for an account of this). The state-space form of the LFM can be obtained by combining the component GP SSMs with the system SSM to yield the following augmented model:
	\begin{align} \label{eq:gplfmss}
	\begin{split}
	\begin{bmatrix}
	\dot{\x}(t) \\ \dot{\z}^{(1)}(t) \\ \dot{\z}^{(2)}(t) \\ \vdots 
	\\ \dot{\z}^{(\nf)}(t)
	\end{bmatrix} &= \begin{bmatrix}
	\Ac & \ve{b}_1 \matr{H}^{(1)} & \ve{b}_2 \matr{H}^{(2)} & \ldots & 
	\ve{b}_{\nf}
	\matr{H}^{(\nf)}\\
	\zeros & \F^{(1)} & \zeros & \ldots & \zeros\\
	\zeros & \zeros & \F^{(2)} & \ldots & \zeros\\
	\vdots & \vdots & \vdots & \ddots & \vdots \\
	\zeros & \zeros & \zeros & \ldots & \F^{(\nf)} 
	\end{bmatrix} \begin{bmatrix}
	{\x}(t) \\ {\z}^{(1)}(t) \\ {\z}^{(2)}(t) \\ \vdots 
	\\ {\z}^{(\nf)}(t)
	\end{bmatrix} +  
	\begin{bmatrix}
	\ve{0} \\ \tilde{\wn}^{(1)}(t) \\ \tilde{\wn}^{(2)}(t) \\ \vdots \\ 
	\tilde{\wn}^{(\nf)}(t)
	\end{bmatrix},\\
	\y(t_k) &= 
	\begin{bmatrix}
	\Gc & \ve{j}_1 \matr{H}^{(1)} & \ve{j}_2 \matr{H}^{(2)} & \ldots & 
	\ve{j}_{\nf} \matr{H}^{(\nf)}
	\end{bmatrix}
	\begin{bmatrix}
	{\x}(t_k) \\ {\z}^{(1)}(t_k) \\ {\z}^{(2)}(t_k) \\ \vdots 
	\\ {\z}^{(\nf)}(t_k)
	\end{bmatrix} + \vn_k,
	\end{split}
	\end{align}
	where $\ve{b}_1, \ve{b}_2,\ldots,\ve{b}_{\nf}$ are the columns of $\Bc$ matrix and $\ve{j}_1, \ve{j}_2,\ldots,\ve{j}_{\nf}$ are the columns of $\matr{J}_c$ matrix and vector $\tilde{\wn}^{(j)}(t) = \matr{L}^{(j)} w^{(j)}(t)$, is a vector-valued Gaussian process having spectral density $\matr{Q}_c^{(j)} = \matr{L}^{(j)} \sigma_w^{(j)} {(\matr{L}^{(j)})}^T$, $j = 1,\ldots,\nf$. 
	Equation \ref{eq:gplfmss} can be written using block matrices as follows:
	\begin{align} \label{eq:blockgplfmss}
	    \begin{split}
	        \begin{bmatrix}
            	\dot{\x}(t) \\ \dot{\z}(t) 
            \end{bmatrix} &= 
            \begin{bmatrix}
            	\Ac & \Bcstr \\
            	\zeros & \F^* \\
            \end{bmatrix} 
            \begin{bmatrix}
            	{\x}(t) \\ {\z}(t)
            \end{bmatrix} + 
            \begin{bmatrix}
	            \ve{0} \\ \tilde{\wn}_z(t)
	        \end{bmatrix},\\
            \y(t_k) &= 
	        \begin{bmatrix}
	            \Gc & \Jcstr
	        \end{bmatrix}
	        \begin{bmatrix}
	            {\x}(t_k) \\ {\z}(t_k) 
	        \end{bmatrix} + \vn_k
	    \end{split}
	\end{align}
	where
	\begin{align}
	    \begin{split}
	        \z(t) &\coloneqq  \begin{bmatrix} \left({\z}^{(1)}(t)\right)^T & \left({\z}^{(2)}(t)\right)^T & \hdots & \left({\z}^{(\nf)}(t)\right)^T \end{bmatrix}^T \\
	        \tilde{\wn}_z(t) &\coloneqq  \begin{bmatrix} \left(\tilde{\wn}_z^{(1)}(t)\right)^T & \left(\tilde{\wn}_z^{(2)}(t)\right)^T & \hdots & \left(\tilde{\wn}_z^{(\nf)}(t)\right)^T \end{bmatrix}^T \\
	        \Bcstr &\coloneqq \begin{bmatrix} \ve{b}_1 \matr{H}^{(1)} & \ve{b}_2 \matr{H}^{(2)} & \ldots & 
	        \ve{b}_{\nf} \matr{H}^{(\nf)} \end{bmatrix}\\
	        \Jcstr &\coloneqq \begin{bmatrix} \ve{j}_1 \matr{H}^{(1)} & \ve{j}_2 \matr{H}^{(2)} & \ldots & 
	        \ve{j}_{\nf} \matr{H}^{(\nf)} \end{bmatrix}\\
	        \F^* &\coloneqq   \begin{bmatrix}  
	                                \F^{(1)} & \zeros & \ldots & \zeros\\
                                    \zeros & \F^{(2)} & \ldots & \zeros\\
                                    \vdots & \vdots & \ddots & \vdots \\
                                    \zeros & \zeros & \ldots & \F^{(\nf)} 
	                                \end{bmatrix}
	    \end{split}
	\end{align}
	In shorthand notation, the augmented SSM as shown in Equation \ref{eq:blockgplfmss} can be represented by
	\begin{align} \label{eq:augssm}
	\begin{split}
	\dot{\x}^a_{c}(t) &= \F_{ac} \x^a_{c}(t) + \tilde{\wn}(t)\\
	\y(t_k) &= \matr{H}_{ac} {\x}^a_{c}(t_k) + \vn_k.
	\end{split}
	\end{align}
	Here $\x^a_{c}(t) \in \R^{n_{a}}$ is the augmented state vector, $\tilde{\wn}(t) \in \R^{n_a}$ is a concatenated vector-valued Gaussian process with spectral density $\matr{Q}_c$, as shown
	\begin{align} \label{eq:Qc}
	\tilde{\wn}(t) = \begin{bmatrix} \zeros \\ \tilde{\wn}^{(1)}(t) \\ 
	\tilde{\wn}^{(2)}(t) \\ 
	\vdots \\ \tilde{\wn}^{(\nf)}(t)	\end{bmatrix}, \quad
	\matr{Q}_c = \begin{bmatrix}
	\zeros & & & & \\
	& \matr{Q}_c^{(1)} & & &\\
	& & \matr{Q}_c^{(2)} & &\\
	& & & \ddots & \\
	& & & & \matr{Q}_c^{(\nf)}
	\end{bmatrix}, 
	\end{align} 
	and matrices $\F_{ac} \in \R^{n_a \times n_a}$, $\matr{L}_{ac} \in \R^{n_a \times \nf}$ and $\matr{H}_{ac} \in \R^{\no \times n_a}$ where $n_a = \ns + m_1+\ldots+m_{\nf}$.
	
	\subsection{Joint posterior inference of inputs and states}
	The discrete-time form of the augmented SSM (Equation \ref{eq:augssm}) can be obtained as 
	\begin{align} \label{eq:daugssm}
	\begin{split}
	\x^a_{k} &= \F_{ad} \x^a_{k-1} + \tilde{\wn}_{k-1},\\
	\y_k &= \matr{H}_{ad} {\x}^a_{k} + \vn_k,
	\end{split}
	\end{align}
	where the state-space matrices $\F_{ad} = e^{\F_{ac} \Delta t}$ and $\matr{H}_{ad} = \matr{H}_{ac}$. $\tilde{\wn}_{k-1}$ is a zero-mean Gaussian white noise vector representing the discrete-time form of $\tilde{\wn}(t)$ whose covariance is given by
	\begin{align} \label{eq:Qintegral}
	\matr{Q}_d = \int_{0}^{\Delta t}  \matr{\Psi}_{a}(\Delta t - \tau) 
	\matr{Q}_c {\matr{\Psi}_{a}(\Delta t - \tau)}^T d\tau,
	\end{align} 
	where $\matr{\Psi}_{a}(\tau) = e^{\F_{ac} \tau }$ is the matrix exponential of the state-transition matrix. The integral in Equation \ref{eq:Qintegral} is solved using matrix fraction decomposition (see \cite{sarkka2006thesis} for implementation details). Zero-mean Gaussian white noise $\wn_{k-1}$ with covariance $\Qn^x$ can be added to the process model in Equation \ref{eq:daugssm} to account for unmodelled dynamics. The resulting modified form of Equation \ref{eq:daugssm} can be written as 
	\begin{align} \label{eq:Fad_Had}
	\begin{split}
	\x^a_{k} &= \F_{ad} \x^a_{k-1} + {\tilde{\wn}}^a_{k-1},\\
	\y_k &= \matr{H}_{ad} {\x}^a_{k} + \vn_k,
	\end{split}
	\end{align}
	where $\tilde{\wn}^a_{k-1}$ is the modified white noise vector with modified covariance $\Qn^a$ as defined below:
	\begin{equation} \label{eq:Qa_calc}
	\tilde{\wn}^a_{k-1} = 
	\begin{bmatrix}
	\tilde{\wn}_{k-1} \\ \tilde{\wn}^{(1)}_{k-1} \\ \tilde{\wn}^{(2)}_{k-1} \\ 
	\vdots \\ \tilde{\wn}^{(\nf)}_{k-1}
	\end{bmatrix}, \quad 
	\Qn^a = \Qn_d + 
	\begin{bmatrix}
	\Qn^x &        &    		&   &\\
	    & \zeros &    		&   &\\
	    &  &\zeros    		&   &\\
	    &        & 	&\ddots   &\\
	    & 		 &			& & \zeros\\
	\end{bmatrix}.
	\end{equation}
	$\wn_{k-1}$ and $\vn_k$ are assumed to be uncorrelated to each other and their joint covariance is expressed through Equation \ref{eq:noisecov}.
	
	The posterior distribution of the states and the forces $p(\x^a_{k}|\y_{1:N},\tcf) = \mathcal{N}(\ve{m}^a_{k|N}, \Pc^a_{k|N})$ can be estimated with the classical Kalman filter (and smoother). Here, $\tcf$ is a vector comprising all covariance function hyperparameters i.e.  $\tcf = \{\tcf_{j}\}_{j=1}^{\nf}$. The estimation is started from a Gaussian noise prior $\mathcal{N}(\ve{m}^a_{0|0},\Pc^a_{0|0})$ where the covariance matrix	$\Pc^a_{0|0}$ has the following block diagonal form
	\begin{equation} \label{eq:P0a}
	\Pc^a_{0|0} = \text{blkdiag}\left(\Pc^x_{0|0}, \Pc^{(1)}_{0|0},\ldots, 
	\Pc^{(\nf)}_{0|0}  \right).
	\end{equation}
	$\Pc^x_{0|0}$ is the initial covariance matrix for the non-augmented state vector $\x_k$ chosen according to some prior knowledge. The covariance matrix $\Pc^{(j)}_{0|0} $ for the $j$th latent 
	force is set equal to the	steady state covariance matrix obtained using Equation \ref{eq:ctre}, i.e.\ $\Pc^{(j)}_{0|0} = \Pc^{(j)}_{\infty}$, $j =1,\ldots,\nf$.
	
	\subsection{Hyperparameter optimization for GP covariance functions} 
	\label{sec:hyp_opt}
	The structural model parameter matrices and that of the covariance functions corresponding to latent forces are transformed into parameters of the augmented GPLFM state-space model. As already stated in Section \ref{sec:objective}, the knowledge of the structural system parameters is assumed to be known \emph{a priori} (either from FE model or from system identification), and therefore the augmented state estimation results will depend only on the parameters of the chosen covariance functions for the latent force components. In general, a parametric family of covariance 
	functions is chosen and the hyperparameters are optimized based on the measurement data. Typical hyperparameters include lengthscale and signal variance for a standard family of covariance functions. The hyperparameters can be estimated in different ways, including maximization of marginal likelihood \cite{rasmussen2006gaussian}, maximum a posteriori \cite{murphy2012machine}, and Markov Chain Monte Carlo methods \cite{filippone2013comparative}. In this study, the optimized hyperparameters are obtained by maximizing the likelihood function based on the measurements. 
	Maximum likelihood estimates of the hyperparameters (i.e. signal variance and lengthscale) of the covariance function(s) can be obtained by minimizing the negative log-likelihood (or maximizing the log-likelihood) of the measurements as follows: 
	\begin{align}
	\begin{split}
	\hat{\tcf} 
	&= \underset{\tcf}{\arg\min} \left[- \log p(\y_{0:T}|\tcf)\right] \\
	&= \underset{\tcf}{\arg\min} \left[- \log \left( p(\y_0|\tcf) \prod_{k=1}^T p(\y_k|\y_{0:k-1},\tcf) \right)  \right] \\
	&= \underset{\tcf}{\arg\min} \left[- \log p(\y_0|\tcf) - \sum_{k=1}^T \log p(\y_k|\y_{0:k-1},\tcf)  \right]. \\
	\end{split}
	\end{align}
	Using Kalman filter recursions  (see \ref{sec:Kfs}), the probabilities can be obtained as $p(\y_0|\tcf) = \mathcal{N} \left( \y_0 | \matr{H}_{ad}(\tcf) \m^a_{0|0} \right)$ and $p(\y_k|\y_{0:k-1},\tcf) = \mathcal{N} \left( \y_k \big| \matr{H}_{ad}(\tcf) \hat{\m}^a_{k|k-1} \right)$ where $\m^a_{0|0}$ is the estimate of the initial augmented state vector $\x^a_0$ and  $\hat{\m}^a_{k|k-1}$ is the $k$th predicted augmented state vector. The minimization of the negative log-likelihood of the measurements can then be expressed in terms of the innovations $\ve{e}_k = \left( \y_k - \matr{H}_{ad}(\tcf) \hat{\m}^a_{k|k-1} \right)$ and the innovation covariance $\matr{S}_k$  (see \cite{sarkka2013bayesian} for more details) as:
	\begin{align}
	\begin{split}
	\hat{\tcf} 
	&=\underset{\tcf}{\arg\min} \left[ \sum_{k=1}^N 
	\left(\log \det \matr{S}_k  + 
	\ve{e}_k^T \matr{S}_k \ve{e}_k \right) \right]
	\end{split}
	\end{align}
	The expressions for $\ve{e}_k$ and $\matr{S}_k$ are provided in Equations \ref{eq:innov} and \ref{eq:Scov} respectively. The minimization can be done using optimization tools such as MATLAB's built-in functions \texttt{fminunc} or \texttt{fmincon}.
	It is noteworthy to mention that maximum likelihood optimization may get stuck in a local minimum; to avoid this one may need to start the optimization from different initial points.
	An algorithm depicting the steps involved in the proposed algorithm is shown in Algorithm \ref{alg:gplfm}.
	
	\begin{algorithm} [htbp!]
	\caption{GPLFM based joint input and state estimation}\label{alg:gplfm}
	\textbf{Requirements:} Provide continuous-time state-space matrices $\Ac$, $\Bc$, $\Gc$ and $\Jc$ of the structural model, and some previously measured data $\mathcal{D} = \{\y_1,\ldots,\y_N\}$ \\
	\For{$j=1,\ldots,n_f$}{
	Choose covariance functions (refer Section \ref{sec:choicegpcf}) and initialize the hyperparameters $\bm{\theta}.$}
	Convert the GP covariance functions into an equivalent continuous-time SSM and obtain $\F^{(j)}$, $\matr{H}^{(j)}$, $\matr{L}^{(j)}$, $\matr{P}^{(j)}_{0|0}$ and $\sigma_w^{(j)}$ (refer to Section \ref{sec:GPss} and \ref{sec:Matern}).\\
	Construct the continuous-time augmented SSM matrices, $\F_{ac}$, $\matr{H}_{ac}$, $\Qn_c$, as shown in Equation \ref{eq:augssm} and \ref{eq:Qc}. \\
	Select values for covariance matrices $\Qn^x$, $\Rn$, $\Pc^x_{0|0}$ and mean vector $\ve{m}^a_{0|0}$ (typically assumed zero vector), and calculate $\Pc^a_{0|0}$ and $\Qn^a$ from Equation \ref 
	{eq:P0a} and \ref{eq:Qa_calc} respectively.\\
	Compute the optimum hyperparameters, $\hat{\bm{\theta}}$, by maximizing the likelihood of the data $\mathcal{D}$ (see Section \ref{sec:hyp_opt})\\
	Use $\hat{\bm{\theta}}$ to compute $\hat{\F}_{ad}$, $\hat{\matr{H}}_{ad}$, $\hat{\Qn}^a$ and $\hat{\Pc}^a_{0|0}$ as depicted in Equations \ref{eq:Fad_Had} and \ref{eq:Qa_calc}.\\
	Use $\hat{\F}_{ad}$, $\hat{\matr{H}}_{ad}$, $\hat{\Qn}^a$, $\hat{\Pc}^a_{0|0}$ for estimating the inputs and the states with Kalman filter (and smoother) (\ref{sec:Kfs}).
	\end{algorithm}
	
	\subsection{Some properties of GPLFM} 
	\label{sec:GPLFMdiscussion}
	\noindent \textit{Property 1}: The augmented state-space model used in Kalman filter-based approaches \cite{lourens2012augmented, naets2015stable, maes2016joint} can be considered a special case of the proposed GPLFM. 
	\begin{proof}
    The continuous-time SSM of the augmented Kalman filter formulation with noisy discrete measurements can be written as
	\begin{align} \label{eq:akfss}
	\begin{split}
	\begin{bmatrix}
	\dot{\x}(t) \\
	\dot{\f}(t)
	\end{bmatrix} &= 
	\begin{bmatrix}
	\Ac & \Bc \\ \zeros & \zeros
	\end{bmatrix} \begin{bmatrix}
	{\x}(t) \\
	{\f}(t)
	\end{bmatrix} + 
	\begin{bmatrix}
	\zeros \\ \tilde{\wn}^f(t)
	\end{bmatrix}\\
	\y(t_k) &= 
	\begin{bmatrix}
	\Gc & \Jc 
	\end{bmatrix} 
	\begin{bmatrix}
	{\x}(t_k) \\
	{\f}(t_k)
	\end{bmatrix} + \vn_k,
	\end{split}
	\end{align} 
	where $\tilde{\wn}^f(t)$ is a vector Gaussian process driving the evolution of the force $\f(t)$. Comparing the above equation with the augmented SSM of GPLFM in Equation \ref{eq:blockgplfmss}, one can deduce that the augmented SSM of GPLFM reduces to that of AKF in Equation \ref{eq:akfss} if $\z(t) = \f(t)$ or equivalently, $\F^*= \zeros_{\nf \times \nf}$ (due to $\F^{(j)} = 0$), $\Bcstr = \Bc$ and $\Jcstr = \Jc$ (due to $\matr{H}^{(j)} = 1 \; \forall \;j = 1,\ldots,\nf$). Hence, the state-space model of AKF \cite{lourens2012augmented} can be considered as a special case of GPLFM. The state-space formulations in \cite{naets2015stable, maes2016joint} are similar to the augmented formulation in \cite{lourens2012augmented} and hence can be also considered special cases of GPLFM.	
	\end{proof}
	
	\noindent \textit{Property 2}:
	Assume that $(\Ac, \Gc)$ is observable, and that each latent force component $\z^{(j)}(t),\; j = 1,\ldots,\nf$ has an exponentially stable state-space representation. Then the augmented system $(\F_{ac}, \matr{H}_{ac})$ is detectable.
	
	\begin{proof}
    Since the state-space representation $\left(\F^{(j)}, \matr{H}^{(j)} \right)$ of the $j$th latent force component $\z^{(j)}(t)$ is exponentially stable, the combined state-space representation of all latent force components is also exponentially stable. 
    The detectability of the full system $\left(\F_{ac}, \matr{H}_{ac}\right)$ can be determined using the Popov-Belevitch-Hautus (PBH) criterion (\cite{hautus1969controllability}, which states that the full system is detectable if and only if the PBH matrix
    \begin{equation} \label{eq:pbh}
        PBH =   \begin{bmatrix}
                    s\eye-\F_{ac} \\
                    \matr{H}_{ac}
                \end{bmatrix}    
    \end{equation}
    is a full column-rank for all $s \in \mathbb{C}$ or the undetectable modes are stable. This criteria has been used to show un-observability of the augmented state-space model given in Equation \ref{eq:akfss} when only acceleration measurements are used (see Equation (21) in \cite{naets2015stable}). In the case of GPLFM, expanding Equation \ref{eq:pbh} in terms of block matrices 
    \begin{equation}
        PBH =   \begin{bmatrix}
                    s\eye - \Ac & -\Bcstr \\
                    \zeros & s\eye - \F^* \\
                    \Gc & \Jcstr
                \end{bmatrix}
    \end{equation}
    it can be seen that the state-space representation of GPLFM has full column-rank $\forall \; s \in \mathbb{C}$ due to $(\Ac, \Gc)$ being observable and $(\F^{(j)}, \matr{H}^{(j)})$ being stable for all $j=1,\ldots,\nf$. Therefore, the augmented system of GPLFM is detectable with all types of measurements.
	\end{proof}

	\noindent \textit{Property 3}:
	Assume that $\Ac$ and $\F^{(j)}$, $j=1,\ldots,\nf$ are stable, then irrespective of the type of measurements (e.g.\ velocity, accelerations) used, the augmented system $(\F_{ac}, \matr{H}_{ac})$ has no marginally stable transmission zeros  and admits stable inversion.
	
	\begin{proof}
	The proof presented here uses the Moylan's algorithm \cite{moylan1977stable} for continuous-time linear systems. 
		Consider the following matrix $\matr{U}(s)$
		\begin{equation}
		\matr{U}(s) = \begin{bmatrix}
		\Ac - s \eye & \Bcstr \\
		\zeros & \matr{F}^*- s \eye\\
		\Gc & \Jcstr
		\end{bmatrix}_{\brc{\ns + M + \no }  \times \brc{\ns + M }} 
		\end{equation}
		The dimensions of $\Ac$ are $\ns \times \ns$, $\Bcstr$ is $\ns \times M$, $\F^*$ is $M \times M$, $\Gc$ is $\no \times \ns$ and $\Jcstr$ is $\no \times M$. Note $M = m_1+\ldots+m_{\nf}$, where $m_j$ is the size of $\F^{(j)}$. 
		The dimensions of $\matr{U}$ are $(\ns+M+\no) \times (\ns+M)$, and therefore the maximum rank of $\matr{U}$ can be $(\ns+M)$. 
		Using the Schur complement of $(\Ac - s \eye)$, one can equivalently write 
		\begin{align}
		\begin{split}
		\text{rank} \brc{\begin{bmatrix}
			\Ac - s \eye & \Bcstr \\
			\zeros & \matr{F}^*- s \eye\\
			\Gc & \Jcstr
			\end{bmatrix}}
		&= \text{rank} \brc{ \begin{bmatrix}
			\brc{\Ac - s \eye} & \zeros \\
			\zeros &  \brc{\begin{bmatrix}
				\matr{F}^*- s \eye \\ \Jcstr
				\end{bmatrix} - \begin{bmatrix}
				\zeros \\ \Gc
				\end{bmatrix} \brc{\Ac - s \eye}^{-1} \Bcstr } 
			\end{bmatrix} }\\
		&= \text{rank} \brc{\Ac - s \eye} + \text{rank} \brc{\begin{bmatrix}
			\matr{F}^*- s \eye \\ \Jcstr
			\end{bmatrix} - \begin{bmatrix}
			\zeros \\ \Gc
			\end{bmatrix} \brc{\Ac - s \eye}^{-1} \Bcstr }\\
		&= \text{rank} \brc{\Ac - s \eye} + \text{rank} \brc{\begin{bmatrix}
			\matr{F}^*- s \eye \\ \Jcstr - \Gc \brc{\Ac - s \eye}^{-1} \Bcstr
			\end{bmatrix}_{\brc{\ns+M}\times M } }.
		\end{split}
		\end{align}
		Evaluating the rank of the above system at $s = 0$, it can be seen that
		\begin{align}
		\begin{split}
		\text{rank} \brc{\begin{bmatrix}
			\Ac - s \eye & \Bcstr \\
			\zeros & \matr{F}^*- s \eye\\
			\Gc & \Jcstr
			\end{bmatrix}_{s=0}} &= \text{rank} \brc{\Ac} + \text{rank} \brc{\begin{bmatrix}
			\matr{F}^* \\ \Jcstr - \Gc \Ac^{-1} \Bcstr
			\end{bmatrix}_{\brc{\ns+M}\times M } }
		\end{split}
		\end{align}
		Since $\F^{(j)}$, $j=1,\ldots,\nf$ are assumed stable, $\F^*$ is stable (implying none of the eigenvalues of $\F^*$ are zero), and $\text{rank}(\matr{F}^*) = M$. Therefore, 
		\begin{align}
		\begin{split}
		\text{rank} \brc{\begin{bmatrix}
			\matr{F}^* \\ \Jcstr - \Gc \Ac^{-1} \Bcstr
			\end{bmatrix}_{\brc{\ns+M}\times M }}= M
		\end{split}
		\end{align}
		irrespective of the rank of $\brc{\Jcstr - \Gc \Ac^{-1} \Bcstr}$. Also, $\Ac$ is assumed stable and has $\text{rank}(\Ac) = \ns$, and hence
		\begin{align}
		\begin{split}
		\text{rank} \brc{\Ac} + \text{rank} \brc{\begin{bmatrix}
			\matr{F}^* \\ \Jcstr - \Gc \Ac^{-1} \Bcstr
			\end{bmatrix}} = \ns + M 
		\end{split}
		\end{align}	
		This shows that the rank of the augmented GPLFM state-space model \emph{does not reduce at $s=0$}. Therefore the GPLFM has no marginally stable transmission zeros. If the GP dynamics associated with the forces are removed, that is if $\F^{(j)}=0$, then $\F^*$ degenerates to $\zeros_{\nf \times \nf}$, and $\Bcstr$ and $\Jcstr$ reduces to $\Bc$ and $\Jc$ respectively, and one reverts to the state-space model of AKF (see Property 1). The state-space model underlying the AKF has been shown to possess marginally stable transmission zeros when used with only acceleration/velocity measurements \cite{maes2015design}, which typically leads to drifts in force estimation. Since the GPLFM has no marginally stable transmission zeros, its inversion is stable and is robust to drift in force estimation.
	\end{proof}

	\section{On the choice of covariance functions} \label{sec:choicegpcf}
\begin{figure}[htbp!]
		\centering
		\includegraphics[scale=0.6]{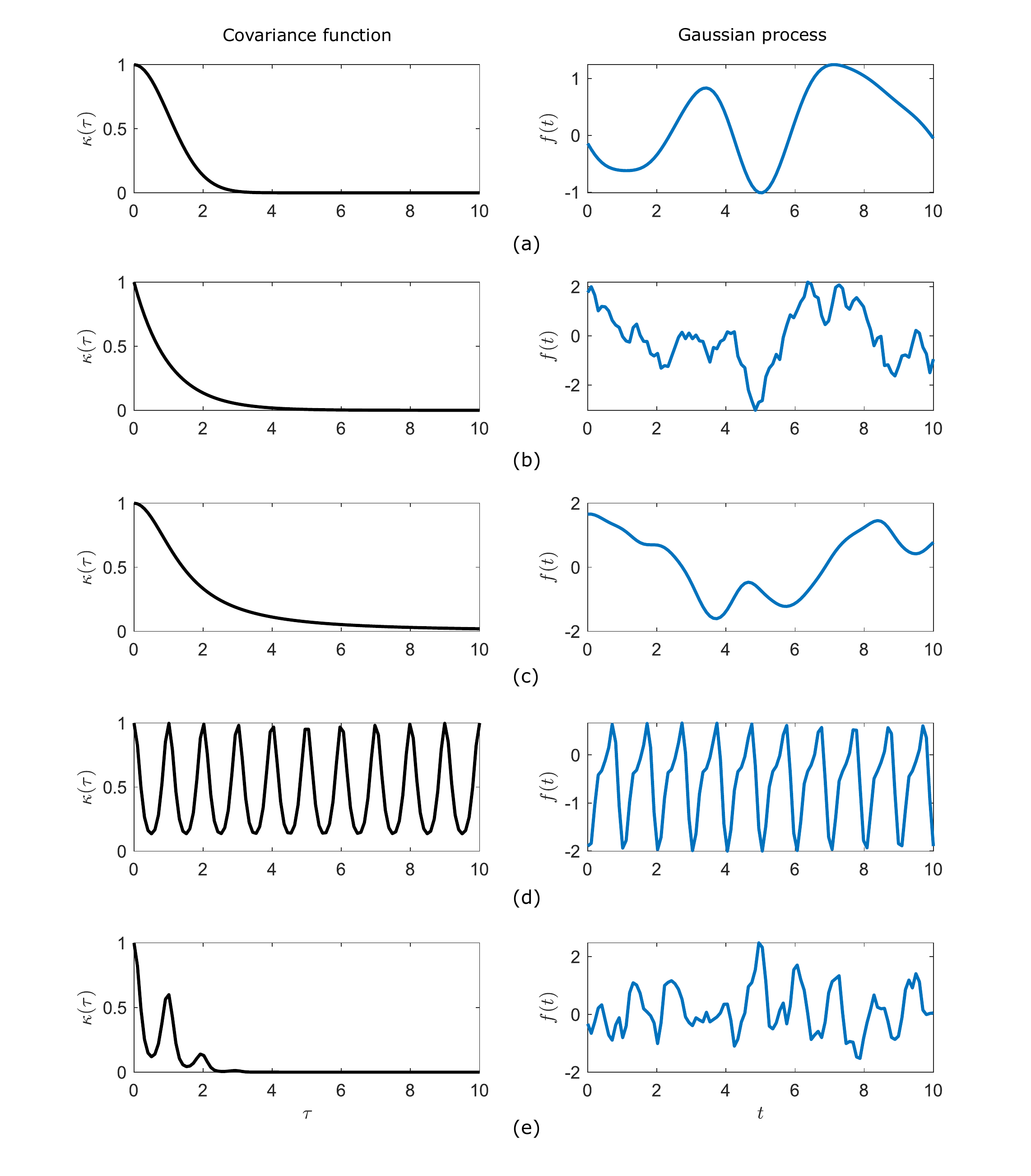}
		\caption{The left panel shows covariance functions and the right panel shows random time series drawn from Gaussian processes with corresponding covariance function in the left panel. Covariance functions: (a) squared exponential, (b) exponential, (c) rational quadratic, (d) periodic and (e) locally periodic (a product of squared exponential and periodic); $t$ represents time and $\tau$ denotes the time difference. All the hyperparameters of the covariance functions are assumed to equal to one.} 
		\label{fig:different_kernels}
	\end{figure}

	In practice, for modelling the forces one has to first choose the covariance functions and then optimize the hyperparameters such that the mismatch between the predicted responses and the observed responses is minimized. In general, this choice is determined by the expected properties of the underlying forces. 
	In this study, a valid choice of covariance function must satisfy these two criteria: (i) they must belong to the class of stationary covariance functions, and (ii) they must admit a stable state-space representation following the spectral factorization approach as described in Section 4.1.
	Several covariance functions exist that satisfy the two criteria, such as the Mat\'{e}rn class, the squared exponential, the exponential, the periodic and the rational quadratic covariance functions. An illustration of different covariance functions and their output processes are shown in Figure \ref{fig:different_kernels}. An added advantage of GPLFM is that different individual covariance functions can be combined to create a rich set of covariance functions, that is, any sum and/or product of the above covariance functions gives a valid covariance function. For example, the sum of squared exponential and periodic covariance functions can be used to model forces that have both random and periodic components. Thus, the set of available covariance functions not only includes the list of individual covariance functions but also all valid combinations constructed by taking sum and/or product between them. However, this comes at the cost of increased complexity due to an increase in the number of tunable hyperparameters.
	
\begin{figure}[htbp!]
		\centering
		\includegraphics[scale=0.7]{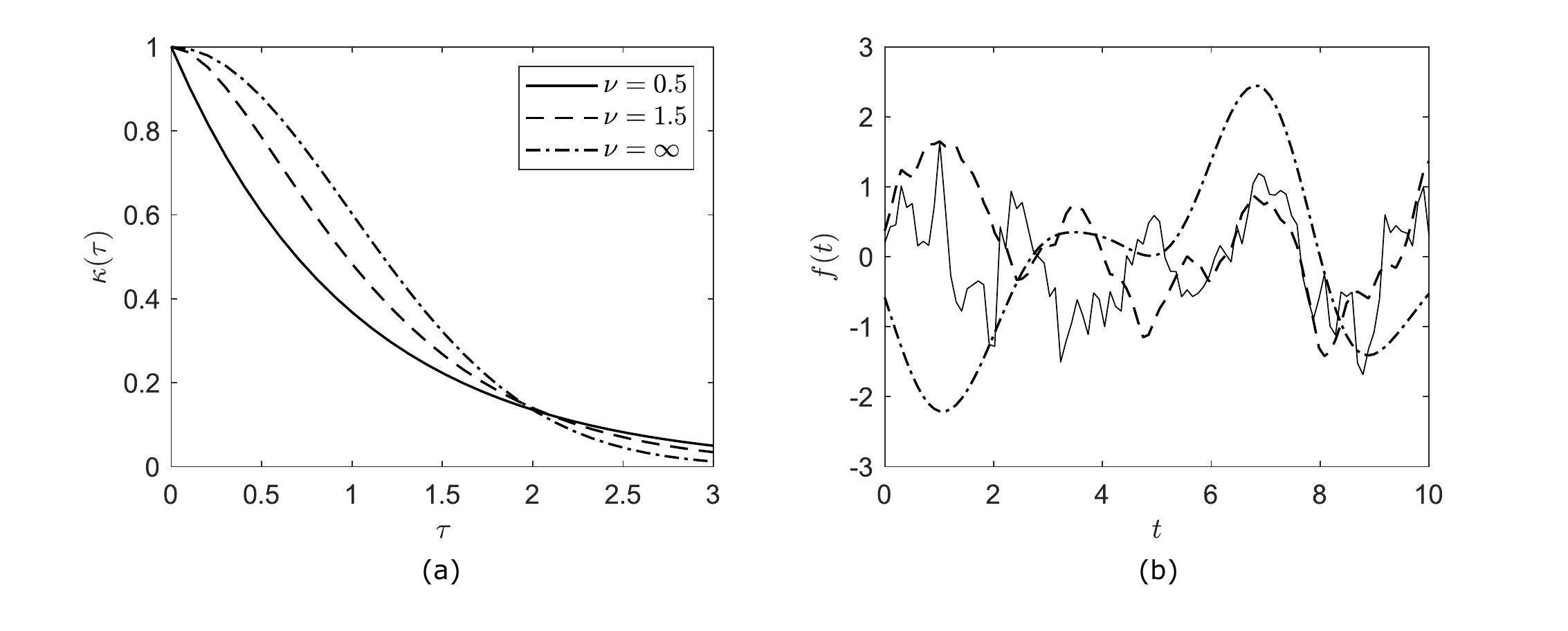}
		\caption{(a) Mat\'{e}rn covariance functions, and (b) random functions drawn from Gaussian processes with Mat\'{e}rn covariance functions, for three different values of $\nu = 0.5, 1.5, \infty$; all other hyperparameters are set to one. $t$ represents time and $\tau$ denotes the time difference.}. 
		\label{fig:different_matern}
	\end{figure}

	Typically, one would choose a covariance function (or a valid combination of covariance functions) that best fits some training data (i.e.,\ measurements of the true force). In the absence of training data, a covariance function can be chosen based on expert guess or prior knowledge. In the absence of both, one would resort to popular choices such as the Mat\'{e}rn family of stationary covariance functions. The Mat\'{e}rn family of stationary covariance functions offers great flexibility in modelling different types of random processes while having only a few tunable hyperparameters. The exponential and the squared exponential covariance functions are also special cases of the Mat\'{e}rn class of covariance functions. Illustrations of different Mat\'{e}rn covariance functions are provided in Figure \ref{fig:different_matern}.
	
	In the field of structural dynamics, the operational forces commonly sustained by structures can be modelled as random processes. For example, wind and seismic loads are typically non-stationary and narrow-banded random processes. Such non-stationary forces can be effectively modelled as GPs generated from stationary covariance functions. For example, rapidly varying (smoothly varying) random forces can be satisfactorily modelled by the GP with exponential (squared exponential) covariance functions. If in addition to randomness there are periodic trends, then a customized covariance function can be constructed by taking the sum/product of periodic and squared exponential covariance functions. 
	However, care needs to be taken when modelling non-random loads that are characterized by abrupt changes or `kinks' in their time history (e.g.\ impact loads, step loads). 
	As stationary covariance functions are usually continuous and differentiable everywhere, capturing the kinks or abrupt changes using GPs with stationary covariance functions can be difficult. In such cases, one may be obliged to adopt the exponential covariance function (which is continuous everywhere but not differentiable) or a combination of the exponential covariance function with the squared exponential function as a useful choice.
	
	\section{Numerical studies} \label{sec:numerical}
	For investigation of the performance of the proposed GPLFM, a 10-storey building is chosen. Each floor of the building is represented by a dof leading to a 10-dof shear building model. The	mass of each floor is assumed to be $200 \si{kg}$ and the inter-storey floor stiffness as $5 \times 10^5 \si{N/m}$. The damping matrix matrix $\matr{C}$ is assumed to be Rayleigh distributed as $\matr{C} = 0.1 \matr{M} + 0.0005 \matr{K}$. The modal frequencies and damping ratios are provided in Table \ref{table:modal_prop}. 
	
	\begin{table}[htbp!]
		\centering
		\caption{Undamped natural frequencies and damping ratios of the 10-dof shear building}
		\begin{tabular}{ccccccccccc} 		\toprule  
			Mode & 1 & 2 & 3 & 4 & 5 & 6 & 7 & 8 & 9 & 10 \\ \midrule
			Frequency (\si{Hz}) & 1.19 & 3.54 & 5.81 & 7.96 & 9.92 & 11.67  & 
			13.15 & 14.34 & 15.21 & 15.74\\
			Damping ratio (\%) & 0.86 & 0.78 & 1.05 & 1.35 & 1.64 & 1.90 & 2.13 
			& 2.31 & 2.44 & 2.52 \\ \bottomrule
		\end{tabular}
		\label{table:modal_prop}
	\end{table}
	
	Only acceleration measurements are assumed available at the floor levels of the structure. For the numerical simulations in this section, the acceleration responses of the structure have been sampled at 100 \si{Hz} and contaminated with 10\% white Gaussian noise to model measurement errors.  
	
	The performance of the proposed GPLFM method has been compared to a suite of Kalman filter-based joint input-state estimation methods, namely, the augmented Kalman filter (AKF) \cite{lourens2012augmented}, the dual Kalman filter (DKF) \cite{azam2015dual}, and the augmented Kalman filter with dummy measurements (AKFdm) \cite{maes2016joint}. All the Kalman filter based methods need the a priori knowledge of the mean and covariance of the initial state $\left(\m^x_{0|0},  \Pc^x_{0|0} \right)$, the mean and covariance of the input $\left( \m^f_{0|0}, \Pc^f_{0|0} \right)$, the state and input process noise covariance matrices $\left(\Qn^x, \Qn^f \right)$, and the measurement noise covariance matrix $\Rn$. In addition, AKFdm --- an extension of AKF, which was proposed to alleviate low frequency drift in force estimates encountered in AKF when using only accelerations --- needs the specification of a dummy displacement measurement covariance matrix $\Rn_{dm}$. Mean values of the initial state $\m^x_0$ and the input $\m^f_0$ are typically assigned zero values for most purposes. The initial state covariance $\Pc^x_0$ and the initial input covariance $\Pc^f_0$ represent the uncertainty in the initial state and input respectively. A lower (or higher) covariance implies stronger (or weaker) belief in the assumed initial values. Process noise covariance matrix $\Qn^x$ signifies the uncertainty in the assumed dynamical model of the structure arising primarily due to modelling error, and lower (or higher) values imply higher (or lower) confidence in the dynamic model. In comparison to $\Qn^x$, the action of input process noise covariance matrix $\Qn^f$ is a bit different, in that it represents the covariance of a fictitious noise process through which the input estimate evolves and plays the role of a regularizer that strongly influences the performance of input and state estimation in AKF, DKF and AKFdm algorithms. A higher $\Qn^f$ will allow the input to assume larger values while varying rapidly, whereas a lower $\Qn^f$ will restrict the input to take relatively smaller values while varying more gradually. In this study, calibration	of the values of $\Qn^f$ for use in AKF, DKF and AKFdm has been carried out mostly with the help of L-curve method along with some manual tuning. The L-curve is obtained by plotting the summed mean squared values of the innovation sequence i.e.\ $\sum \norm{\ve{e}_k}$ against the values of $\norm{\Qn^f}$, and the values of $\Qn^f$ corresponding to the ``corner'' of the L-curve are selected. The measurement noise covariance matrix $\Rn$ represents the degree of uncertainty in the measurements due to measurement instrument errors; an estimate of $\Rn$ can be obtained from the precision of the instrument prescribed by the manufacturer. The values of the dummy displacement measurement covariance matrix $\Rn_{dm}$, needed for AKFdm, has been recommended \cite{naets2015stable} to be chosen with an order of magnitude higher than the covariance of displacement responses for stable estimation.  
	
	In GPLFM, one needs to provide $\m^x_{0|0}$, $\Pc^x_{0|0}$, $\Qn^x$, $\Rn$, $\m^f_{0|0}$ and the choice of the family of covariance function for latent forces prior to implementation. 
	Unlike DKF, AKF and AKFdm where the tuning of $\Qn^f$ controls the accuracy of estimation results, the proposed GPLFM method does not need the specification of $\Qn^f$; instead in GPLFM the hyperparameters of chosen covariance functions dictate the performance of input and state estimation. The hyperparameters of the covariance function can be set based on expert knowledge or computed through maximum likelihood optimization based on measured data as described in Section \ref{sec:hyp_opt}. The latter has been adopted in the study to avoid the dependence on any prior guess or expert knowledge. Once the optimal hyperparameters are computed, the input and state estimation is performed using Kalman filter.

	In this study, exponential covariance functions \textemdash a class of the popular Whittle-Mat\'{e}rn family of stationary covariance functions (see \ref{sec:Matern}) \textemdash have been used for GP modelling of the different excitations, unless explicitly mentioned otherwise. Exponential covariance functions are non-differentiable and the GPs generated using them are continuous everywhere but not differentiable, and are thus useful in modelling random (less smooth) phenomena such as Brownian motion. Since most excitations sustained by operational structures are random in nature, GPs with exponential covariance functions are employed for modelling the input excitations in this study. 
	
	For all numerical simulations in this section, the following values are assumed: $\ve{m}_{0|0}^x = \zeros$, $\ve{m}_{0|0}^f = \zeros$,  $\Pc^x_{0|0} = 10^{-10} \times \eye$, $\Qn^x = 10^{-10} \times \eye$ and $\Rn = 0.1 \times \eye \;(\si{m/s^2})^2$ for GPLFM, DKF, AKF and AKFdm. In what follows, the performance of joint input and state estimation is assessed using different types of excitations i.e.\ impact, harmonic, seismic and random. 
	
	\subsection{Impact excitation}
	An impact excitation of magnitude $10^4\si{N}$ is applied at the topmost floor of the structure. The impact load starts at $t = 3\si{s}$ of the simulation, ramps up linearly from 0 to $10^4$ in 0.05s, peaks at $t = 3.05\si{s}$ and then ramps downs linearly from $10^4$ to 0 in another $0.05\si{s}$. The following values of covariances are used in input and state estimation 
	\begin{itemize}
		\item $\Qn^f = 10^{4} \times \eye \;\si{N}^2$ for DKF, AKF and AKFdm obtained using L-curve criterion as shown in Figure \ref{fig:lcurve_imp1},
		
		\begin{figure}[htbp!]
			\centering
			\includegraphics[scale=\mysize]{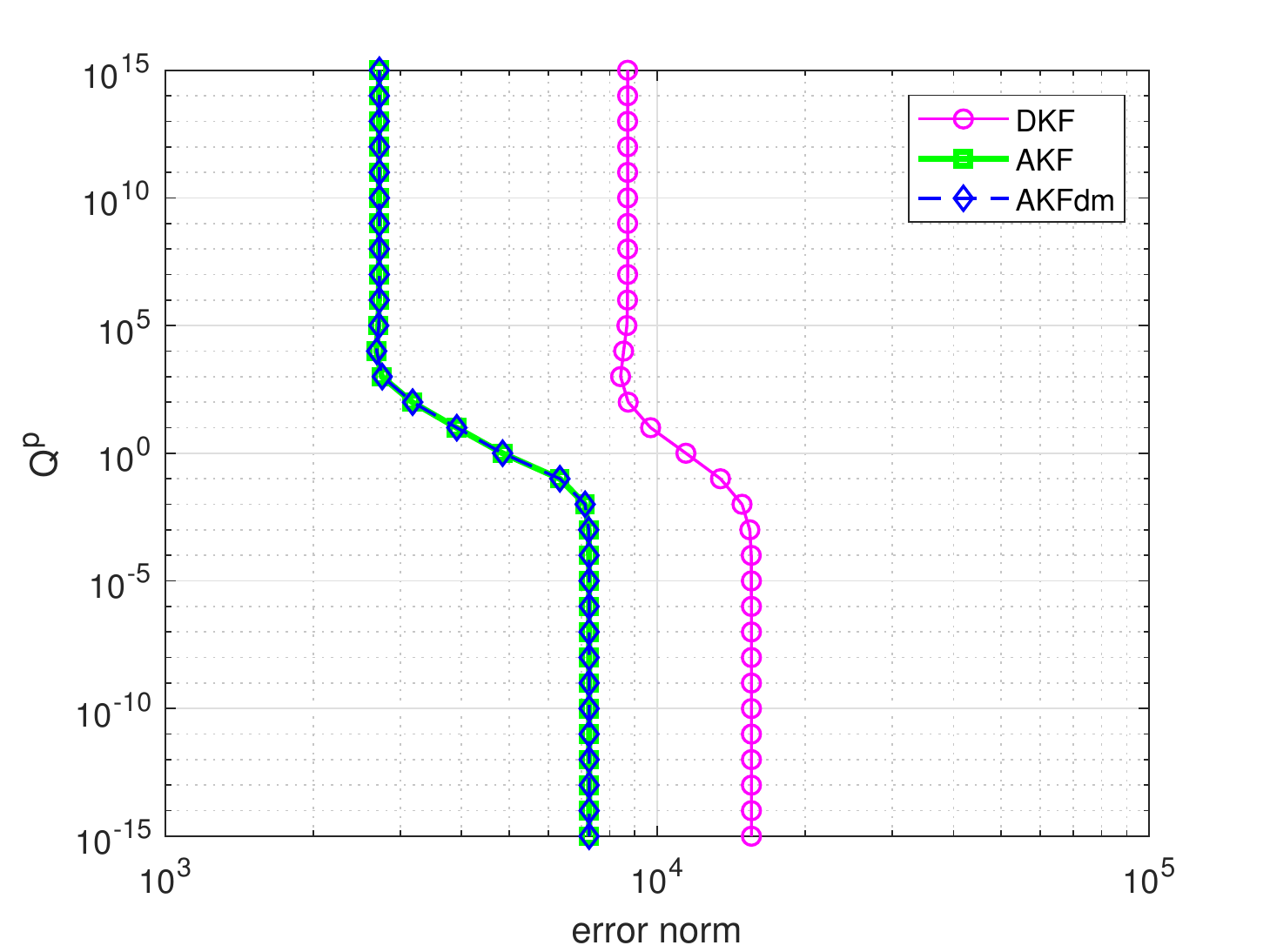}
			\caption{L-curve criterion for $\Qn^f$ calibration in DKF, AKF and AKFdm}
			\label{fig:lcurve_imp1}
		\end{figure}
		\item $\Rn_{dm} = 0.05 \times \eye \;\si{m}^2$ for AKFdm obtained by setting the covariance roughly 10 times the square of the maximum absolute displacement value.
	\end{itemize}
	
	Two cases of measurements are considered:
	\begin{enumerate}[label=(\alph*)]
		\item \textit{All acceleration measurements} \\
		In this case, the state and input estimation is performed using acceleration measurements from all floor levels. The estimated acceleration, velocity and displacement time histories at the 5th floor (chosen as a representative floor) is shown in Figure  \ref{fig:impact_allacc_repdof10_states}. It is found that all the algorithms are able to accurately estimate the acceleration and the velocity states, however the displacement states estimated by AKF and DKF are affected by low frequency drifts. The low frequency drift that manifests in the AKF and the DKF estimates are due to accumulation of integration errors. The AKFdm which uses dummy displacement measurements is able to arrest the drift and provide stable displacement estimates. Both AKFdm and GPLFM provide reasonably accurate estimates of the displacements, with GPLFM performing the best among all the algorithms. Similar estimation results are obtained at all other floor levels. 
		It can be seen that state estimates provided by AKFdm compares closely with that of AKF except the displacement estimates where drift errors due to AKF leads to large estimation error in estimates obtained by AKF. 
		
		\begin{figure}[htbp!]
			\centering
			\includegraphics[scale=\mysize]{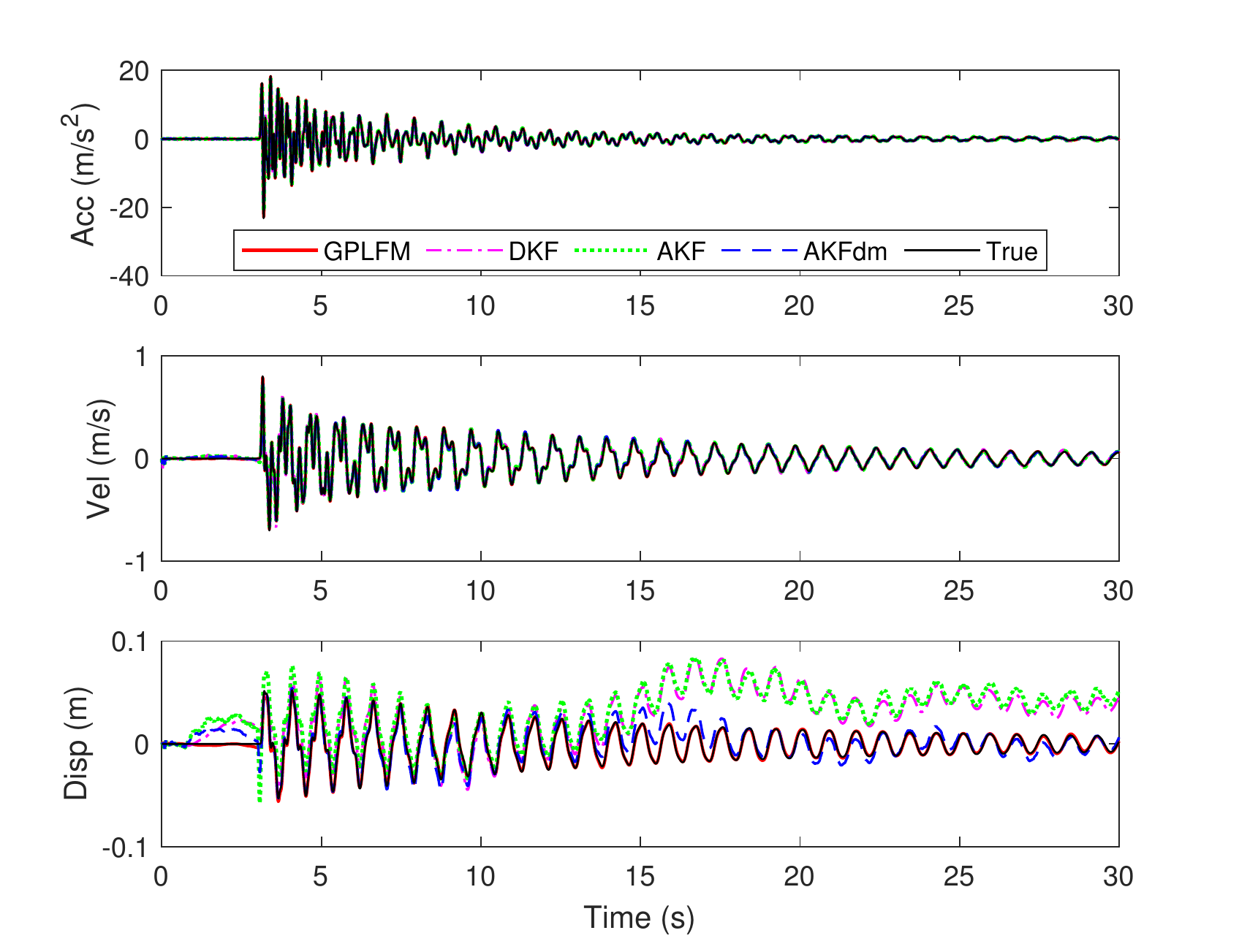}
			\caption{Estimated acceleration (top), velocity (middle) and displacement (bottom) time histories obtained from GPLFM, DKF, AKF and AKFdm at the 5th floor; impact excitation	applied at 10th floor and accelerations measured at all floors}
			\label{fig:impact_allacc_repdof10_states}
		\end{figure}	
		
		The estimated input force histories are shown in Figure \ref{fig:impact_allacc}. It is seen that input force estimates using DKF and AKF are affected by low frequency drift, alike the displacement estimates. AKFdm force estimate shows some post peak oscillations but does well in restricting the drift in estimates. GPLFM seem to provide good force estimates with no drift and negligible oscillations. It can be seen that none of the estimates from any of these algorithms could achieve the exact peak of the actual impulsive force, with the GPLFM estimate furnishing the closest estimate to the true force.  
		
		\begin{figure}[htbp!]
			\centering
			\includegraphics[scale=\mysize]{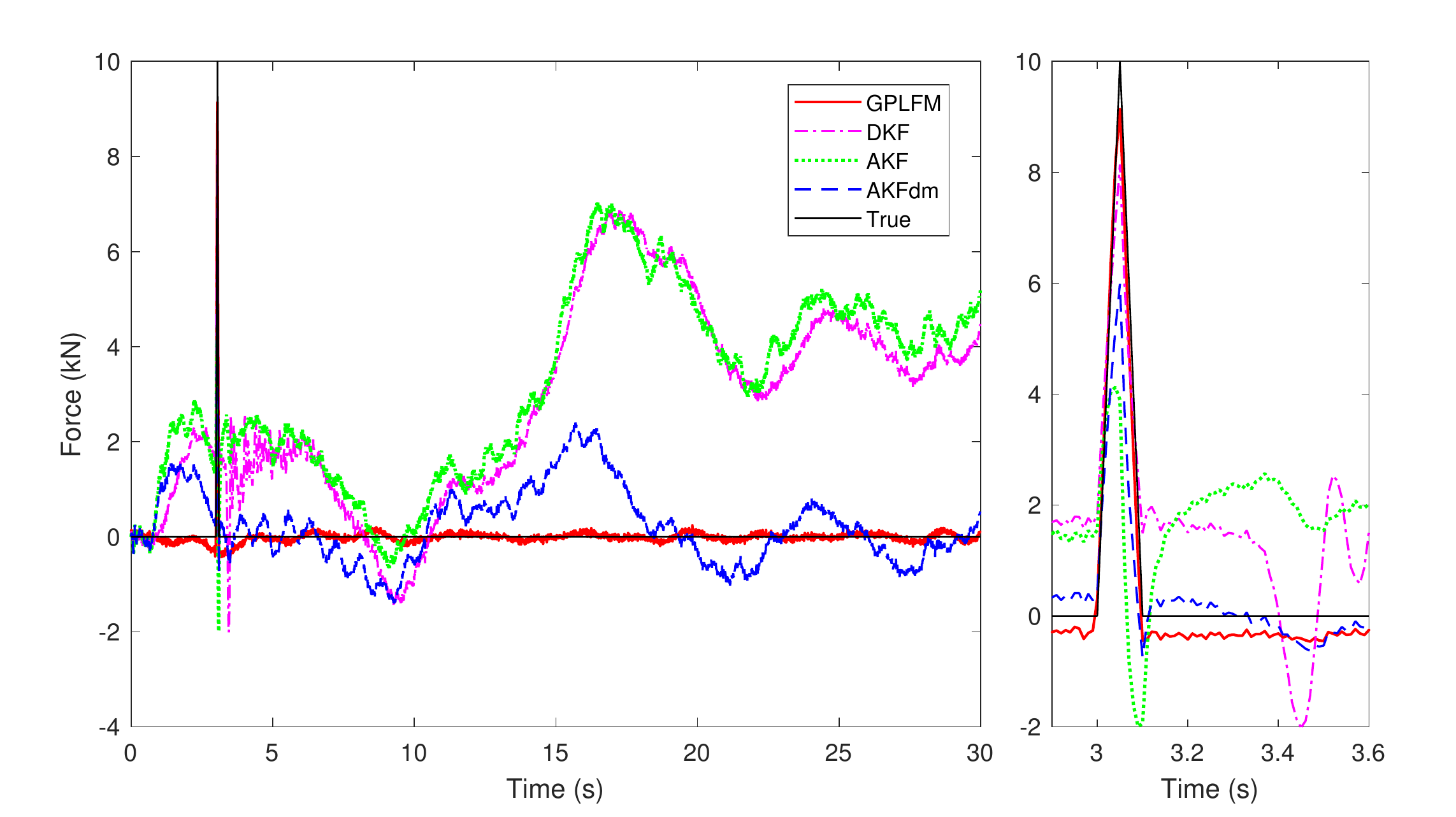}
			\caption{Estimated force time histories obtained from GPLFM, DKF, AKF and AKFdm using accelerations measured at all floors; impact excitation applied at the 10th floor}
			\label{fig:impact_allacc}
		\end{figure}
		
		
		\item \textit{Single acceleration measurement}\\
		Only a single acceleration at the 10th floor level is measured in this case. This corresponds to a collocated measurement case where the force and the measurement locations coincide. Collocated acceleration measurements are desirable in force estimation since the input force contributes directly to collocated acceleration measurements through the direct feedthrough term. The estimates of force time history are provided in Figure \ref{fig:impact_oneacc}. The force estimates obtained using AKF and DKF show low frequency drifts as was encountered previously. However, for AKF, the estimate of the force at the peak improved in this case. The peak force estimated by AKF, DKF and AKFdm match closely to each other, although underestimated by around 40\%. The peak force estimated by GPLFM appear slightly better than the estimated by DKF, AKF and AKFdm, however, the peak is underestimated by 30\% when using a single acceleration measurement. A post peak trough is observed close to  3.5\si{s} in the force estimates from all algorithms. In the authors' opinion, the appearance of the trough may have lead to a drop in the peak force estimated by these algorithms. 
		
		\begin{figure}[htbp!]
			\centering
			\includegraphics[scale=\mysize]{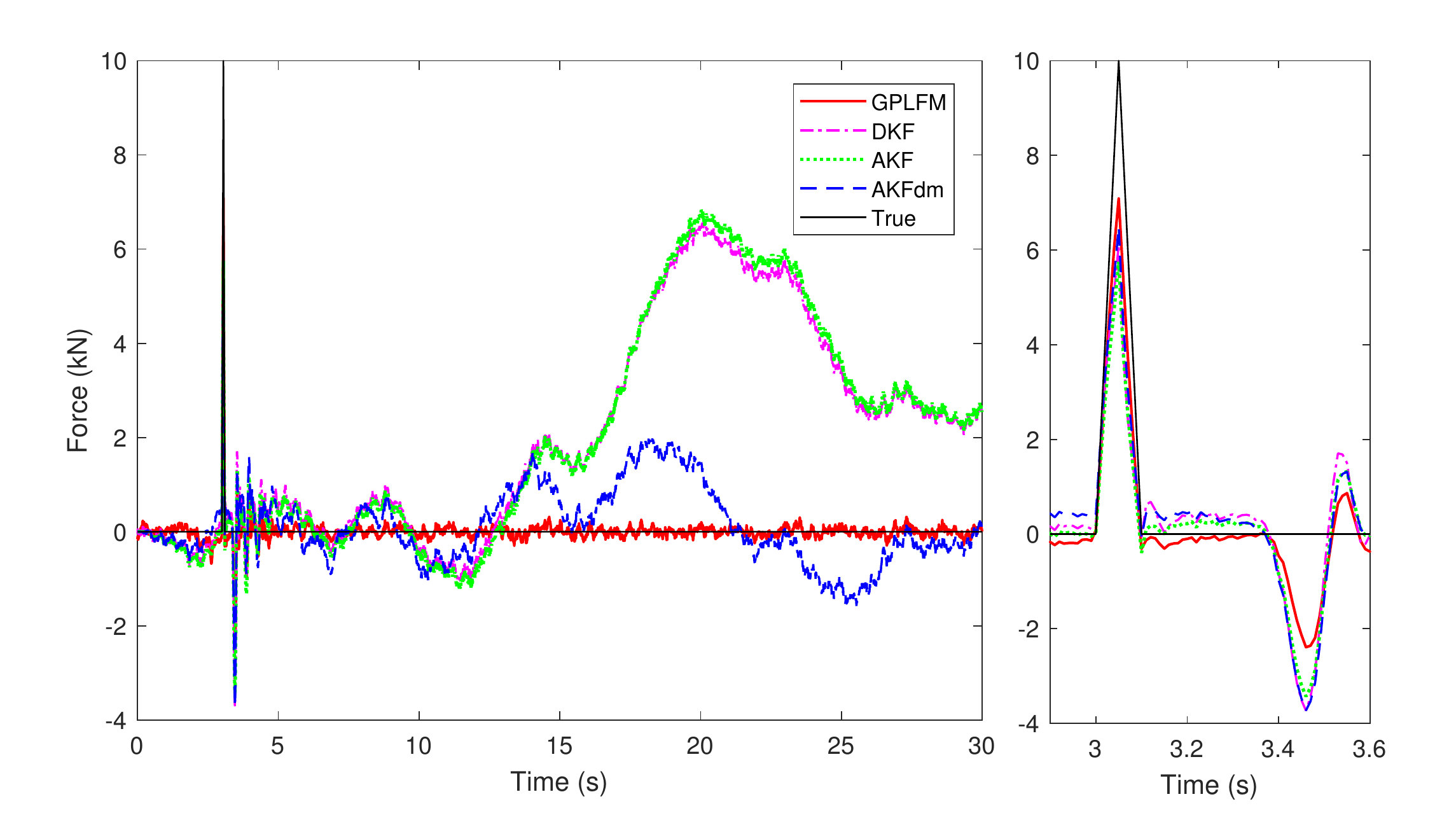}
			\caption{Estimated force time histories obtained from GPLFM, DKF, AKF and AKFdm using acceleration measured at the 10th floor; impact excitation applied at the 10th floor}
			\label{fig:impact_oneacc}
		\end{figure}  
		The acceleration, velocity and displacement state estimates at the 5th floor is shown in Figure \ref{fig:impact_oneacc_repdof10_states}. It is found that even with the use of a single collocated acceleration, the acceleration and velocity state estimates obtained by all the algorithms at other floors still show good accuracy. For the displacement state, however, the GPLFM provides a better estimate compared to the estimates from other algorithms.
		\begin{figure}[htbp!]
			\centering
			\includegraphics[scale=\mysize]{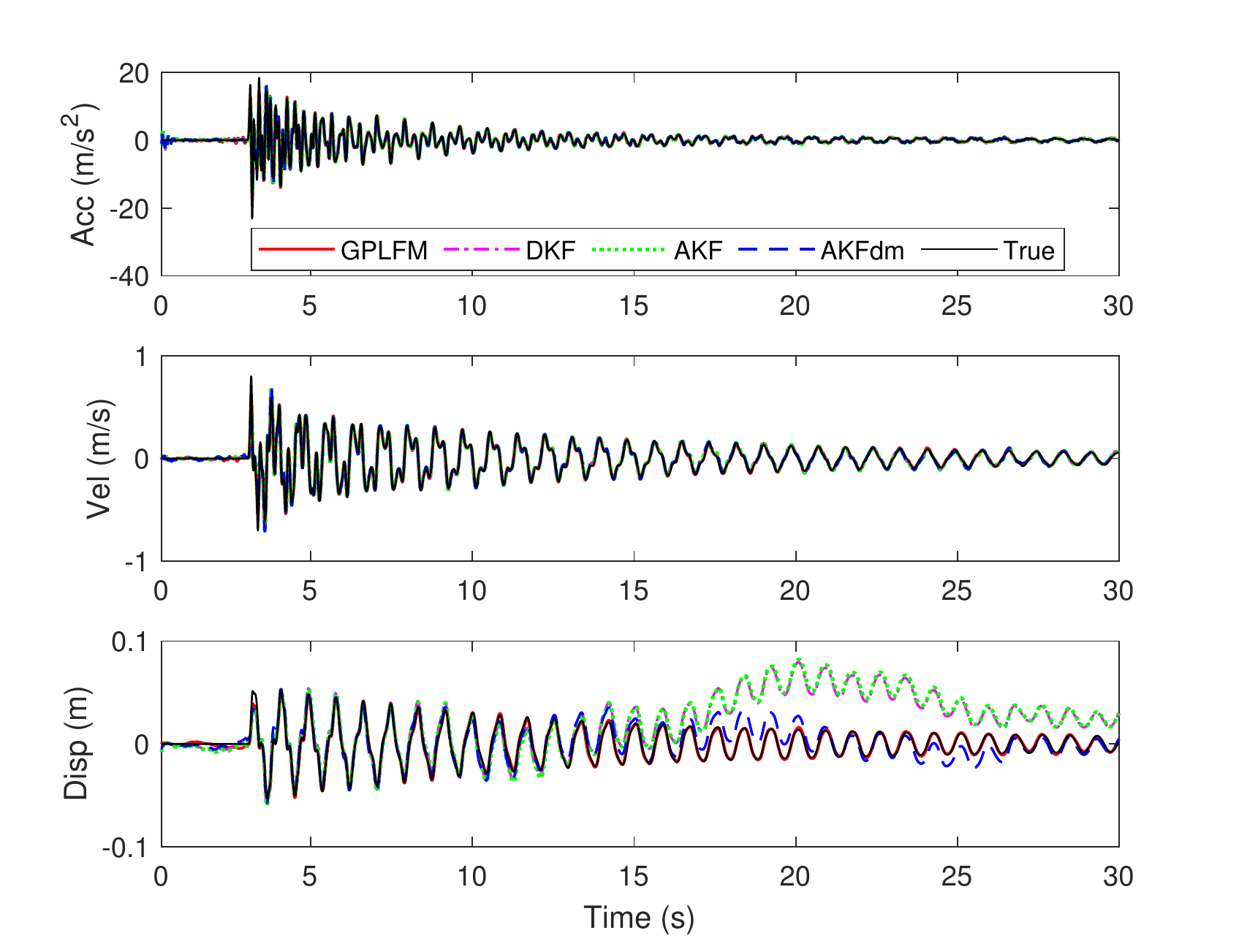}
			\caption{Estimated acceleration (top), velocity (middle) and displacement (bottom) time histories obtained from GPLFM, DKF, AKF and AKFdm at 5th floor; impact excitation applied at the 10th floor and acceleration measured at the 10th floor}
			\label{fig:impact_oneacc_repdof10_states}
		\end{figure}
	\end{enumerate}
	It should be mentioned here that the attempt to estimate an impact loading by a Gaussian process with exponential covariance function is a scenario of mis-specified covariance function model. The mis-specification arises because the impact excitation is being modelled here by a Gaussian process with a continuous exponential covariance function, although an impact is best represented by a discontinuous Dirac-delta function, atleast in theory. Such mis-specification should be avoided if possible, however, for practical reasons such as convenience of using standard forms of covariance function or vague prior information, one inevitably ends up in a situation which resembles some level of mis-specification \cite{rasmussen2006gaussian}. 
	
	\subsection{Harmonic excitation}
	In this section, the performances of GPLFM, DKF, AKF and AKFdm are assessed using harmonic excitation. A constant amplitude sinusoidal excitation of the form $f(t) = A_h \sin (2 \pi \lambda t)$ is applied to the topmost floor of the shear building, where $A_h = 100\si{N}$, $\lambda = 1\si{Hz}$. The following values are used prior to running the algorithms:
	\begin{itemize}
		\item $\Qn^f = 10^{4} \times \eye \;\si{N}^2$ for DKF, AKF and AKFdm set using L-curve criterion, 
		\item $\Rn_{dm} = 10^{-2} \times \eye \; \si{m}^2$ for AKFdm set at 100 times the absolute maximum of true displacement history,
		\item Mat\'{e}rn covariance function with $p=2$ (refer to Equation \ref{eq:Matern52}) is chosen for GPLFM 
	\end{itemize}
	
	For the sinusoidal case, a Mat\'{e}rn covariance function with $p=2$ is used in place of exponential covariance function as it was found difficult to obtain good force estimates with a exponential covariance function. Ideally, a periodic covariance function is most appropriate for estimating a periodic force history, however, a more convenient Mat\'{e}rn covariance function with $p=2$ having the property of being twice differentiable is used in this case. Once again this partly resembles a mis-specified covariance function.  
	
\begin{figure}[htbp!]
		\centering
		\includegraphics[scale=\mysize]{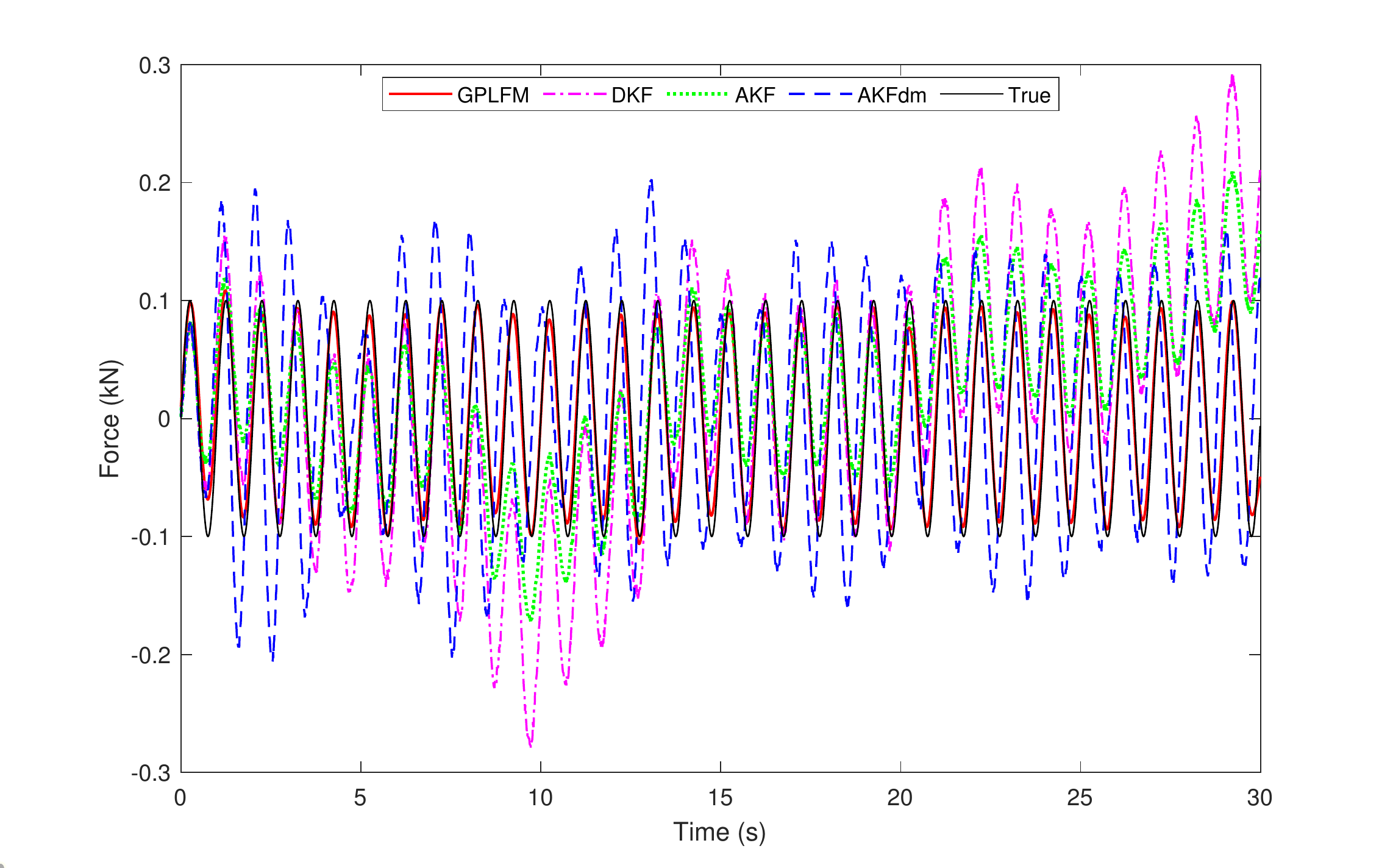}
		\caption{Estimated force time histories obtained from GPLFM, DKF, AKF and AKFdm using  acceleration measured at the 10th floor; sinusoidal excitation applied at the 10th floor}
		\label{fig:sine_oneacc}
	\end{figure}

\begin{figure}[htbp!]
		\centering
		\includegraphics[scale=\mysize]{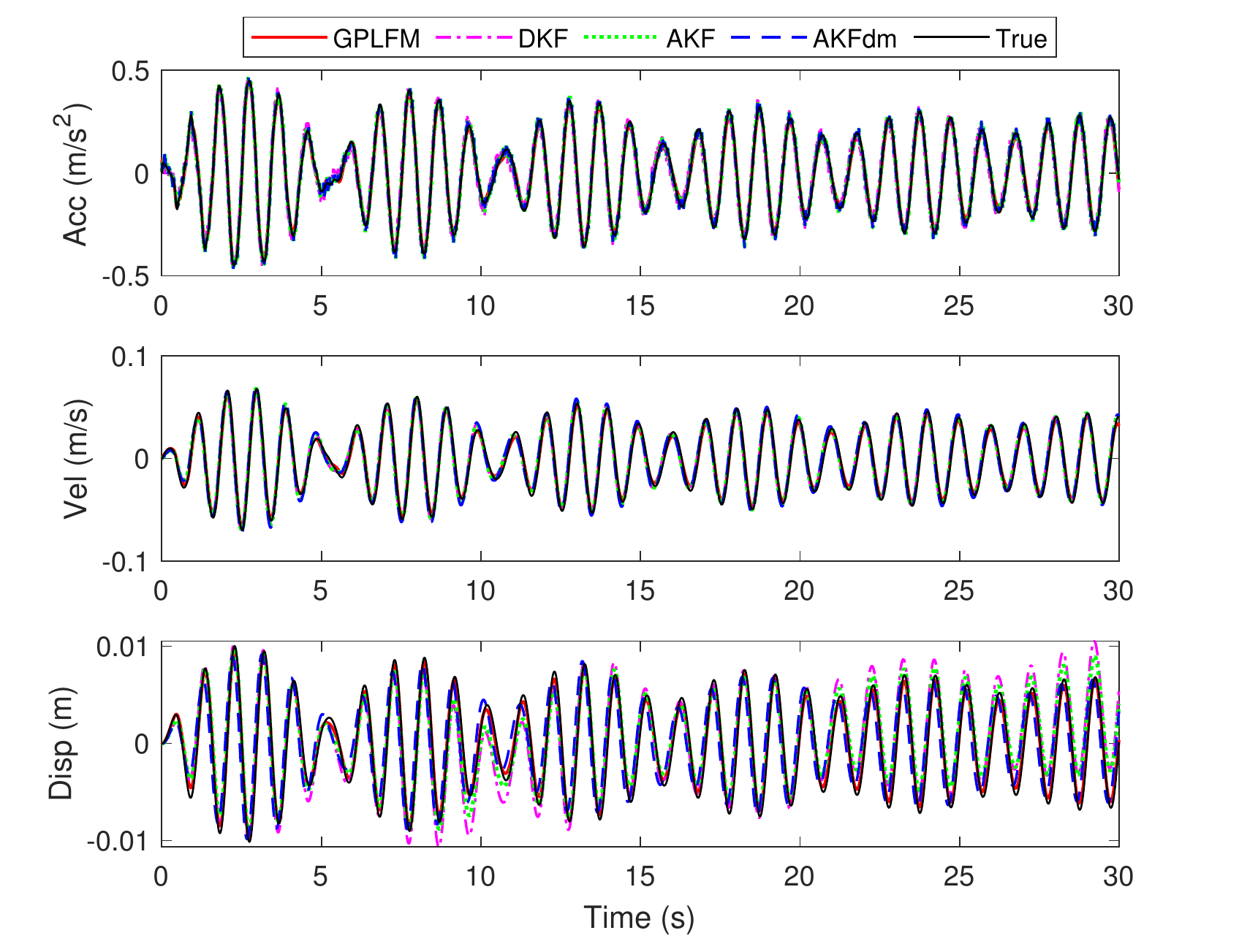}
		\caption{Estimated acceleration (top), velocity (middle) and displacement (bottom) time histories obtained from GPLFM, DKF, AKF and AKFdm at the 5th floor; sinusoidal excitation applied at the 10th floor and acceleration measured at the 10th floor}
		\label{fig:sine_oneacc_repdof10_states}
	\end{figure}

	Using a single collocated acceleration measurement at the topmost floor, the force history is estimated as shown in Figure \ref{fig:sine_oneacc}. It is found that GPLFM is able to provide very accurate estimate of the sinusoidal force history. Furthermore, it can be seen that DKF performed better than AKF in estimating the magnitude of the force, however, both their estimates are affected by drift. AKFdm force estimate tracks the true force in an unbiased manner, although some low frequency oscillations ride the force estimate. Reducing the value of $\Rn_{dm}$ removed the low frequency oscillations in the force estimate from AKFdm, however, that occured at the cost of poor estimation of the force magnitude. The state estimates obtained by the algorithms at the representative 5th floor compare reasonably well with respect to each other as seen in Figure \ref{fig:sine_oneacc_repdof10_states}. The drift in the estimates of DKF and AKF in this case is found to be comparably low. The estimation case with all accelerations produced similar results and is thus not included here. 

	\subsection{Seismic excitation}
	To assess the performance of different algorithms under seismic excitation, the El Centro earthquake record (Imperial Valley Station, May 18, 1940) is used as ground acceleration input to the 10-storey structure. Only the case of a single acceleration measurement at the 10th floor is considered. It was found that using multiple floor accelerations gave similar results and hence not discussed. It is important to note that accelerometers measure only absolute accelerations, and in this case, when no other external force is acting on the structure, measuring absolute accelerations reduces the direct feedthrough term in the observation equation to zero (refer to derivation in \ref{sec:feedthru_derive}). When the direct feedthrough term becomes zero, the DKF algorithm fails to estimate the input since the feedthrough matrix $\matr{J}$ in Equation \ref{eq:dssm} becomes zero and this leads to zero Kalman gain for the input, thereby degenerating the input update stage in DKF implementation. Thus, for this case, a comparison will be drawn among estimates from GPLFM, AKF and AKFdm.  The following values are adopted prior to running the algorithms:
	\begin{itemize}
		\item $\Qn^f = 0.1 \times \eye \;\si{N}^2$ for AKF and AKFdm 
		obtained using L-curve criterion,
		\item $\Rn_{dm} = \eye \; \si{m}^2$ for AKFdm set at roughly 10 times the 
		square of absolute maximum of true displacement history,
	\end{itemize}
	
\begin{figure}[htbp!]
		\centering
		\includegraphics[scale=\mysize]{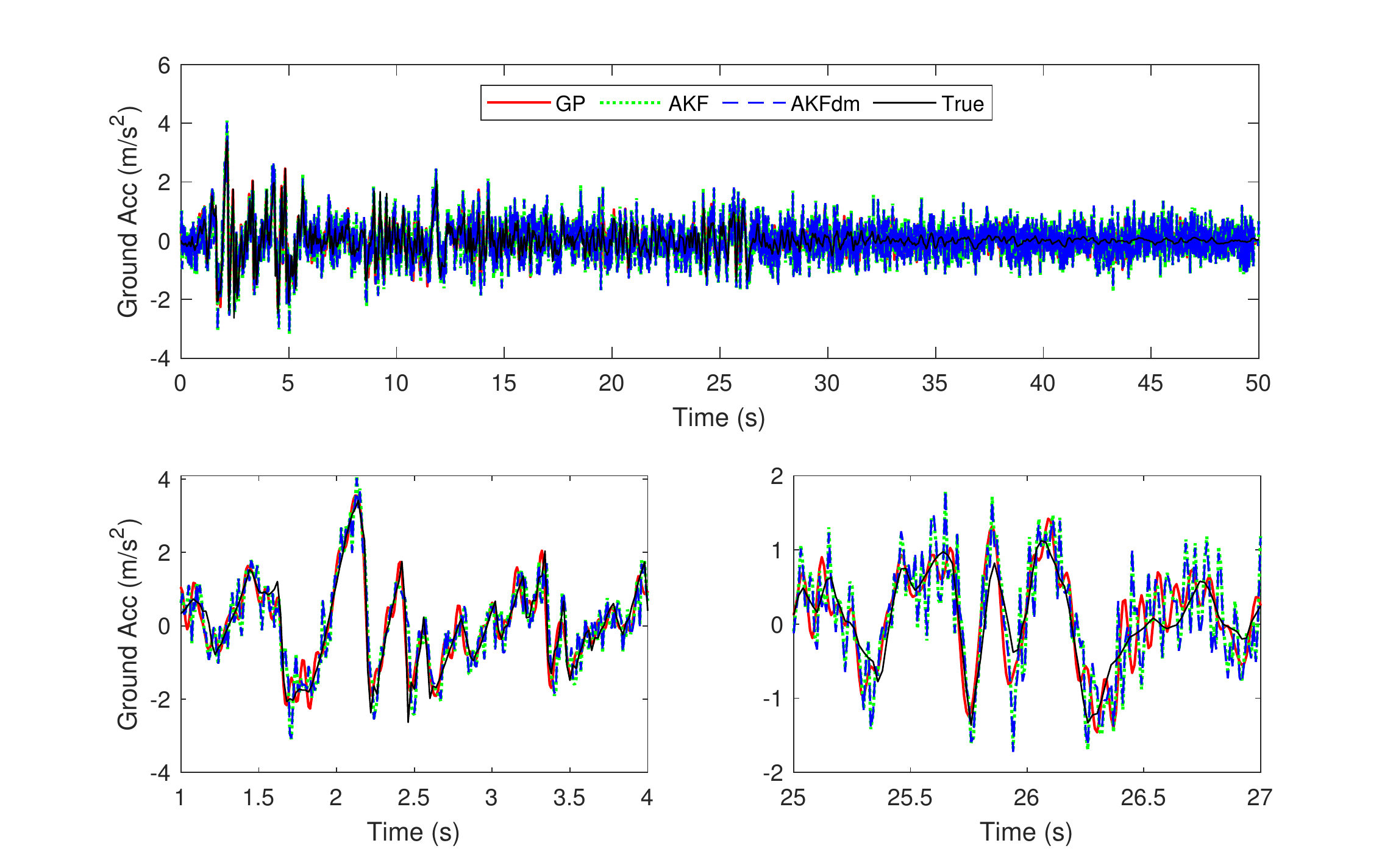}
		\caption{Estimated force time histories obtained from GPLFM, AKF and AKFdm using acceleration measured at the 10th floor; structure subjected to El Centro earthquake ground acceleration}
		\label{fig:seismic_oneacc}
	\end{figure}

	The estimated ground acceleration histories from GPLFM, AKF and AKFdm are shown in Figure \ref{fig:seismic_oneacc}. The GPLFM estimate is able to track the ground acceleration time history quite accurately. Both AKF and 
	AKFdm also demonstrate good tracking of the ground acceleration although with a bit of noise in the estimates.
	
\begin{figure}[htbp!]
		\centering
		\includegraphics[scale=\mysize]{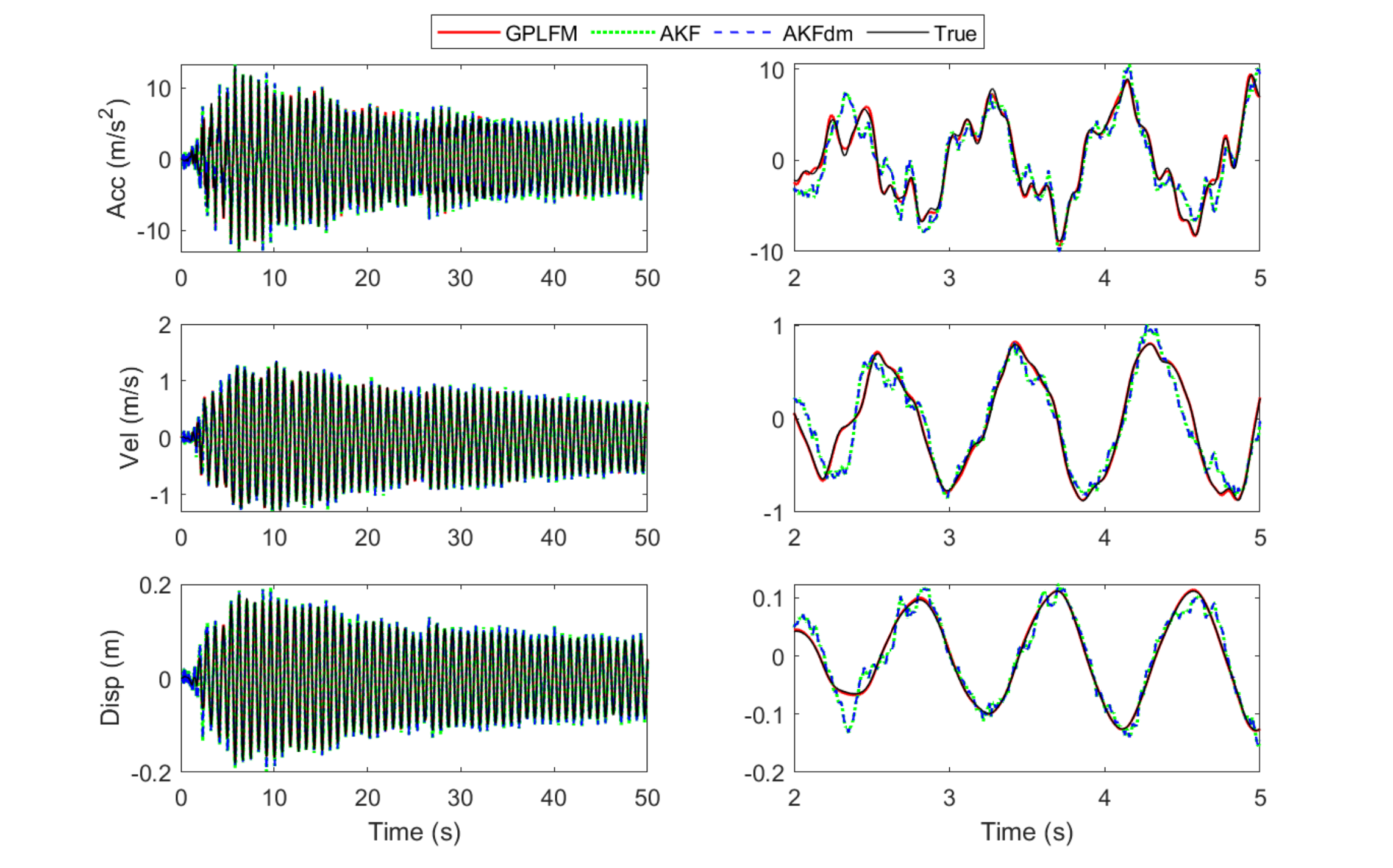}
		\caption{Estimated acceleration (top), velocity (middle) and displacement (bottom) time histories obtained from GPLFM, AKF and AKFdm at the 5th floor; structure subjected to El Centro earthquake ground  acceleration and acceleration measured at the 10th floor}
		\label{fig:seismic_oneacc_repdof5_states}
	\end{figure}

	The estimated acceleration, velocity and displacement states at the 5th floor are shown in Figure \ref{fig:seismic_oneacc_repdof5_states}. The state estimates from all the three algorithms at the 5th floor seem reasonably accurate with GPLFM furnishing the closest estimate to the true states. 
	
	\subsection{Random excitation}
	Most operational structures are subjected to random forces (such as wind, traffic, etc.) that can be usefully modelled by random white noise (or filtered white noise) excitation. In this subsection, the performance of 	GPLFM, DKF, AKF and AKFdm in input and state estimation is assessed under the application of random excitation to the 10-storey shear structure. Two scenarios of random excitation are considered: (I) a single random 
	excitation applied at the 10th floor of the structure, and (II) multiple random excitation applied to all floors of the structure. 
	
	\begin{enumerate} [label = (\Roman*)]
		\item \textit{Single random excitation}\\
		A zero mean Gaussian white noise excitation with standard deviation $1\si{kN}$ is applied to the 10th floor.	The following values of covariances are used:
		\begin{itemize}
			\item $\Qn^f = 10^{4} \times \eye \;\si{N}^2$ for DKF, AKF and AKFdm obtained using L-curve criterion and some manual tuning,
			\item $\Rn_{dm} = 0.05 \times \eye \; \si{m}^2$ for AKFdm set at 100 times the square of absolute maximum of true displacement history
		\end{itemize}
		
		First consider the case where accelerations at all floor levels are measured. Using all acceleration measurements, the estimated force histories from GPLFM, DKF, AKF and AKFdm are shown in Figure \ref{fig:random_allacc} and the estimated states for the 5th floor are shown in Figure \ref{fig:random_allacc_repdof5_states}. It is found that GPLFM estimate is able to provide a very good tracking of the true force history compared to the estimate from other algorithms. As regards the state estimates, both GPLFM and AKFdm provide good estimates of all states while AKF and DKF estimates for displacement states suffer from drift.
		
		Next, the case of a single collocated acceleration measured at the 10th floor is considered. The force and state estimation results from GPLFM, DKF, AKF and AKFdm obtained for this measurement case are shown in 
		Figure \ref{fig:random_oneacc} and 	\ref{fig:random_oneacc_repdof5_states} respectively. It is found that estimation results from using a single collocated acceleration are still reasonable but not as accurate when compared to the results from using all accelerations.
		
		\begin{figure}[htbp!]
			\centering
			\includegraphics[scale=\mysize]{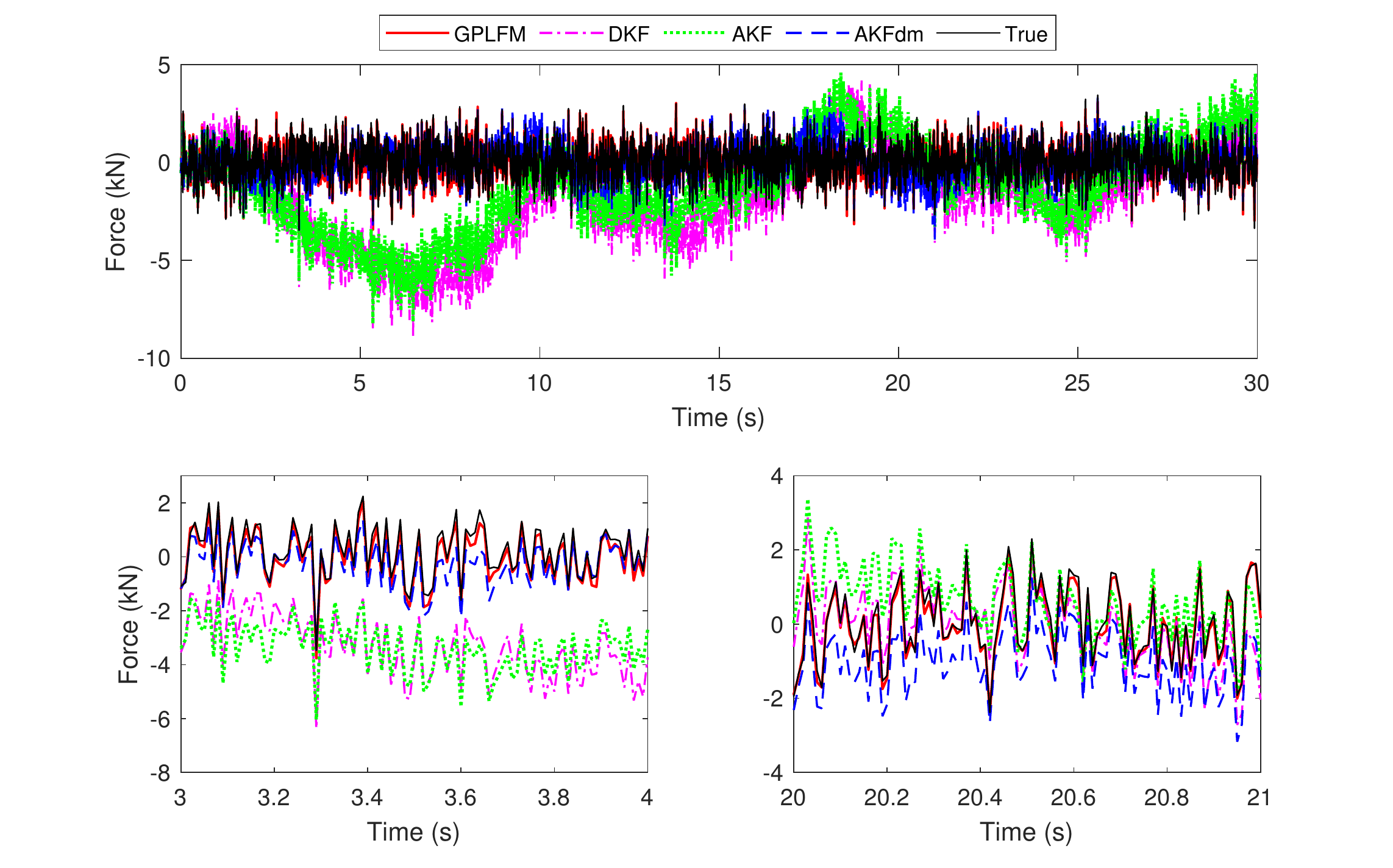}
			\caption{Estimated force time histories obtained from GPLFM, DKF, AKF and AKFdm using acceleration measured at all floors; structure subjected to random excitation at the 10th floor}
			\label{fig:random_allacc}
		\end{figure}
		
		\begin{figure}[htbp!]
			\centering
			\includegraphics[scale=\mysize]{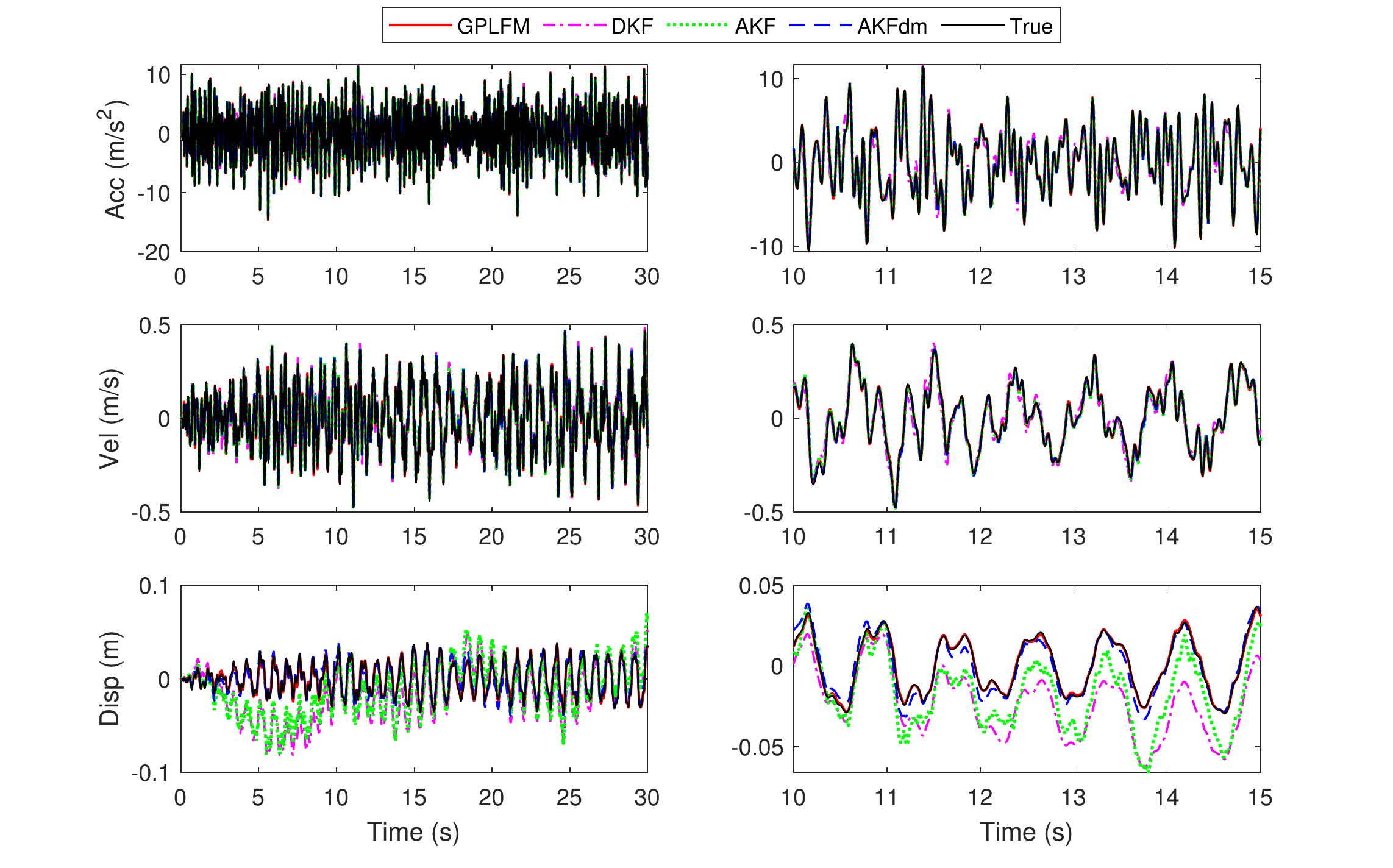}
			\caption{Estimated acceleration (top), velocity (middle) and displacement (bottom) time histories obtained from GPLFM, DKF, AKF and AKFdm at the 5th floor; random excitation applied at the 10th floor and accelerations measured at all floors}
			\label{fig:random_allacc_repdof5_states}
		\end{figure}
		
		\begin{figure}[htbp!]
			\centering
			\includegraphics[scale=\mysize]{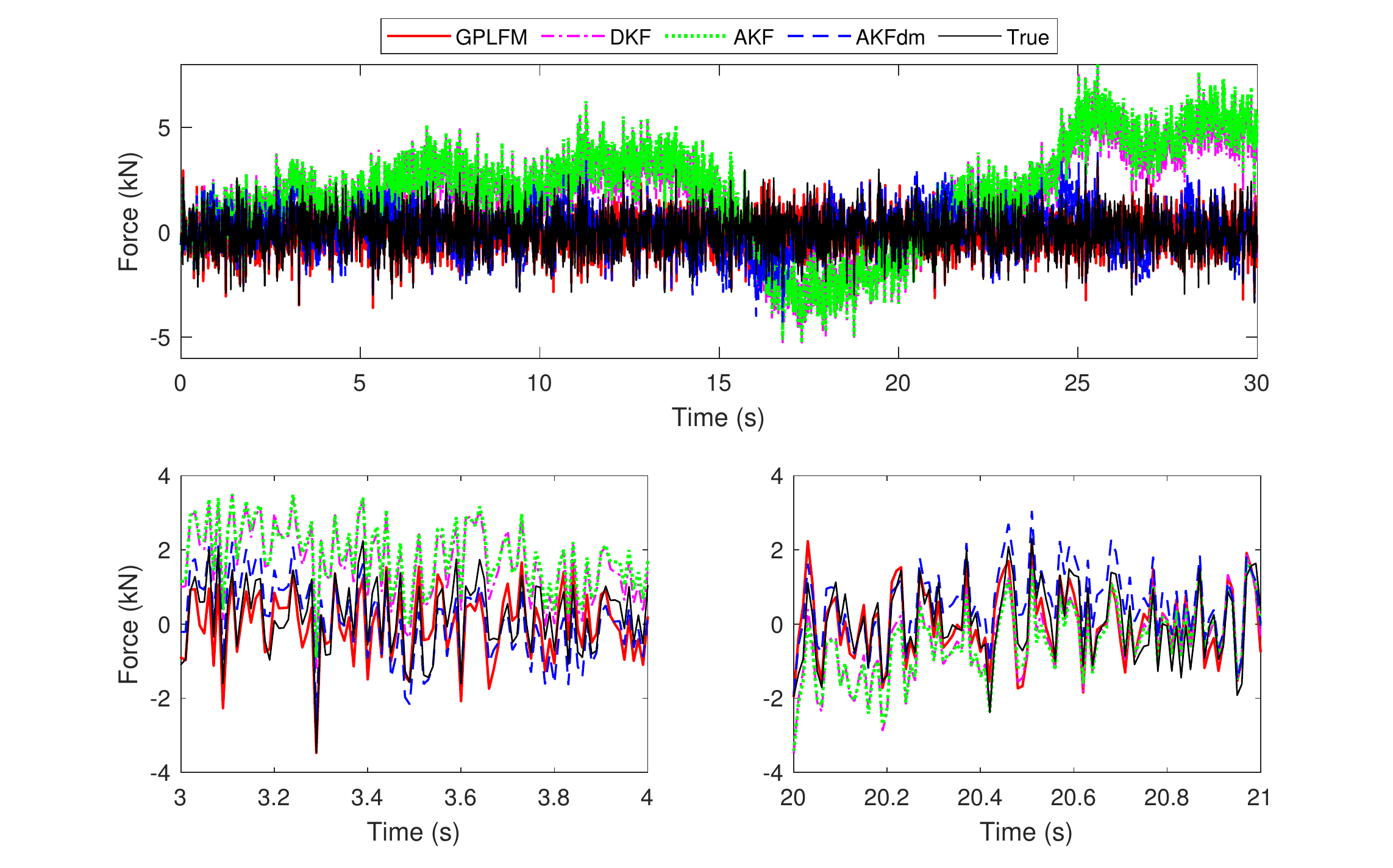}
			\caption{Estimated force time histories obtained from GPLFM, DKF, AKF, AKFdm using acceleration measured at the 10th floor; structure subjected to random excitation at the 10th floor}			\label{fig:random_oneacc}
		\end{figure}
		
		\begin{figure}[htbp!]
			\centering
			\includegraphics[scale=\mysize]{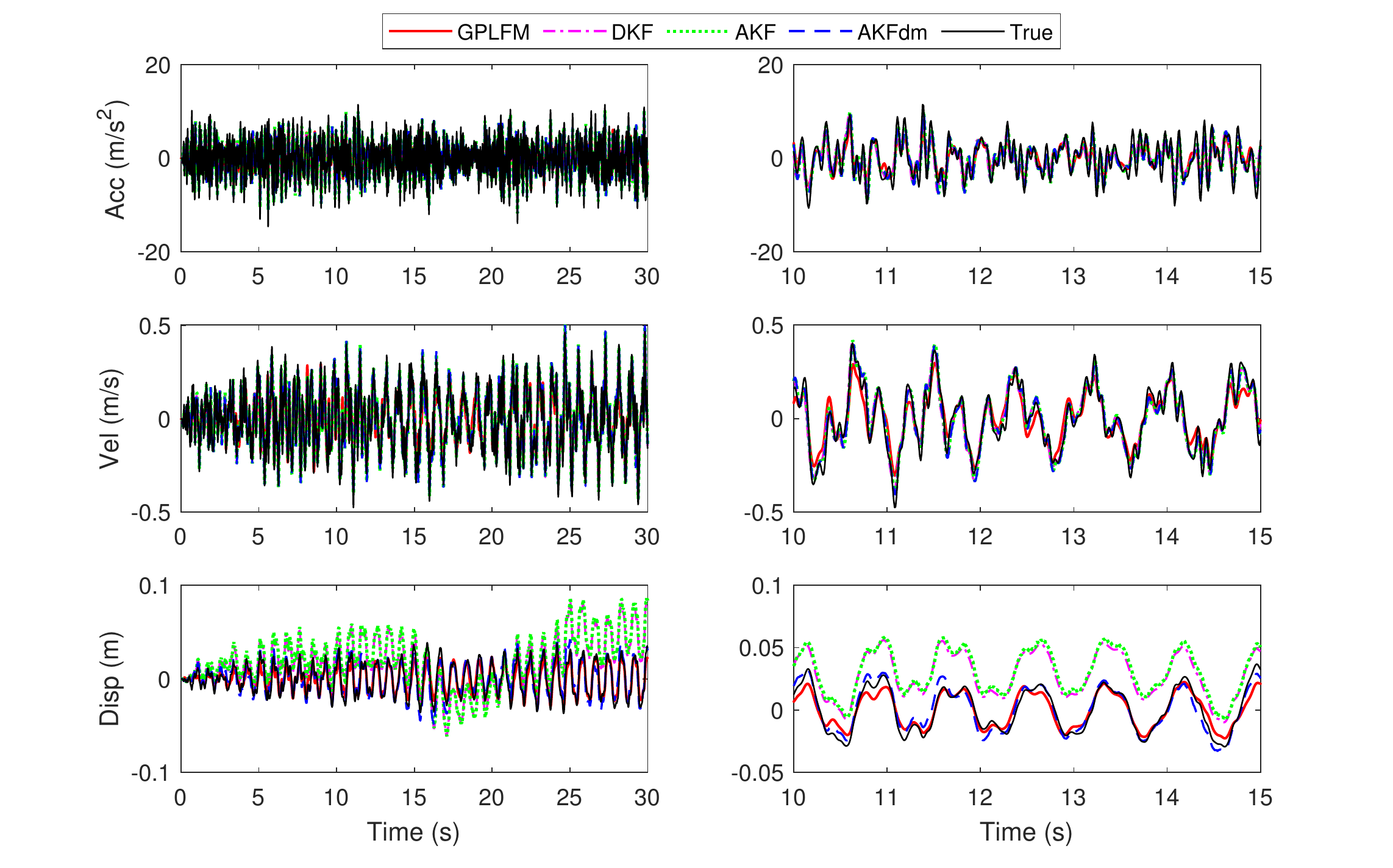}
			\caption{Estimated acceleration (top), velocity (middle) and displacement (bottom) time histories obtained from GPLFM, DKF, AKF and AKFdm at the 5th floor; random excitation applied at the 10th floor and acceleration measured at the 10th floor}
			\label{fig:random_oneacc_repdof5_states}
		\end{figure}
		
		Now, a case where accelerations are measured at floor levels 1, 3, 5, 7 and 9 is considered. This represents a scenario of non-collocated measurements, where the locations of measurements do not coincide with the location(s) of the applied force(s). The force and state estimation results for this case are shown in Figure \ref{fig:random_fewacc} and \ref{fig:random_fewacc_repdof5_states} respectively. It is found that input force estimation using non-collocated observations proves to be a challenging case. The DKF fails to estimate the forces due to the direct feedthrough matrix being zero. The force estimates from GPLFM, AKF and AKFdm seem to somewhat follow the true force history but still are far from accurate, however, among them the GPLFM estimate was found to have the least mean square error. On the other hand, the state estimation results from all the algorithms are found to be reasonably accurate except the state estimates from DKF. 
		\begin{figure}[htbp!]
			\centering
			\includegraphics[scale=\mysize]{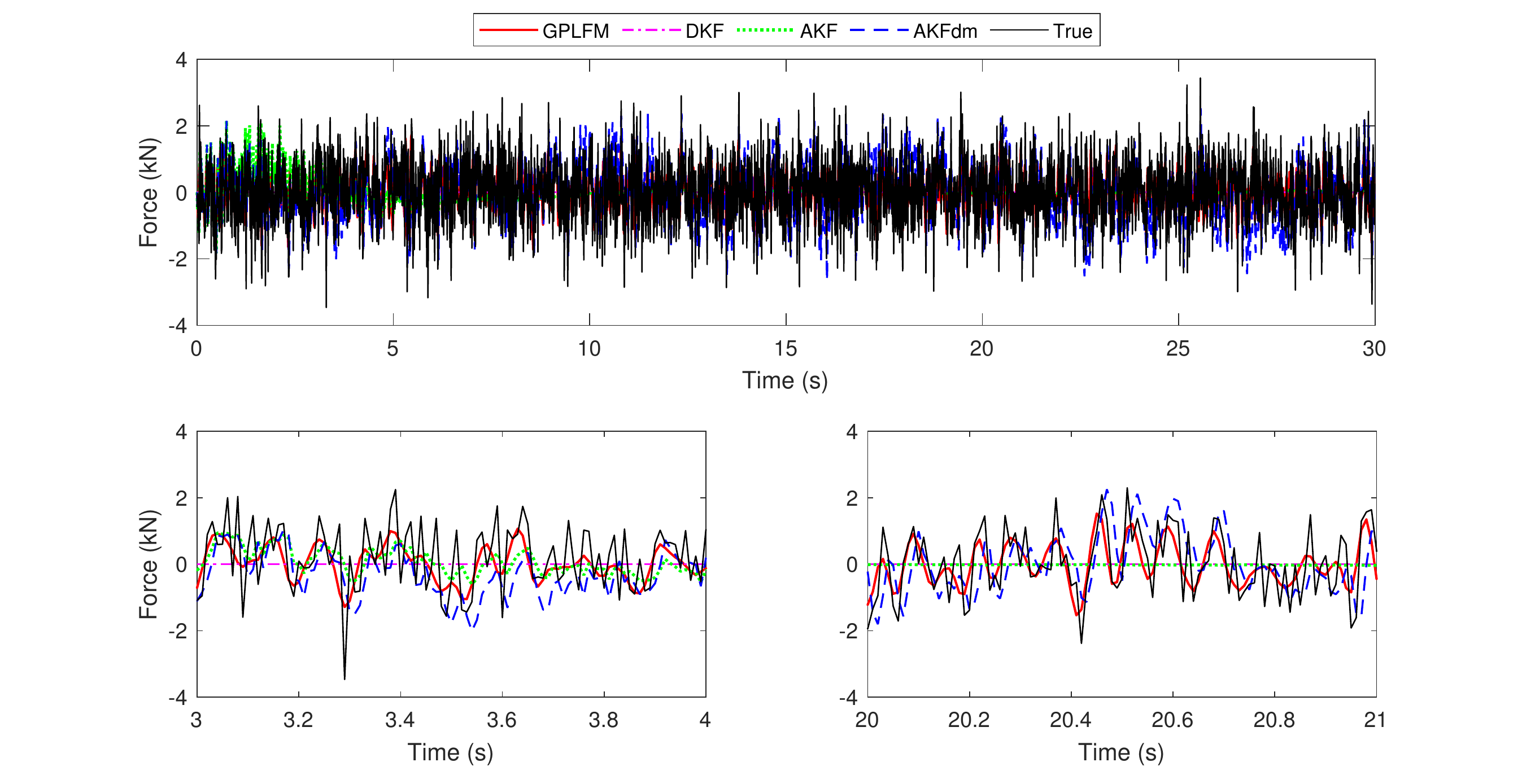}
			\caption{Estimated force time histories obtained from GPLFM, DKF, AKF, AKFdm using non-collocated accelerations measured at floors 1, 3, 5, 7 and 9; random excitation applied at the 10th floor}
			\label{fig:random_fewacc}
		\end{figure}
		
		\begin{figure}[htbp!]
			\centering
			\includegraphics[scale=\mysize]{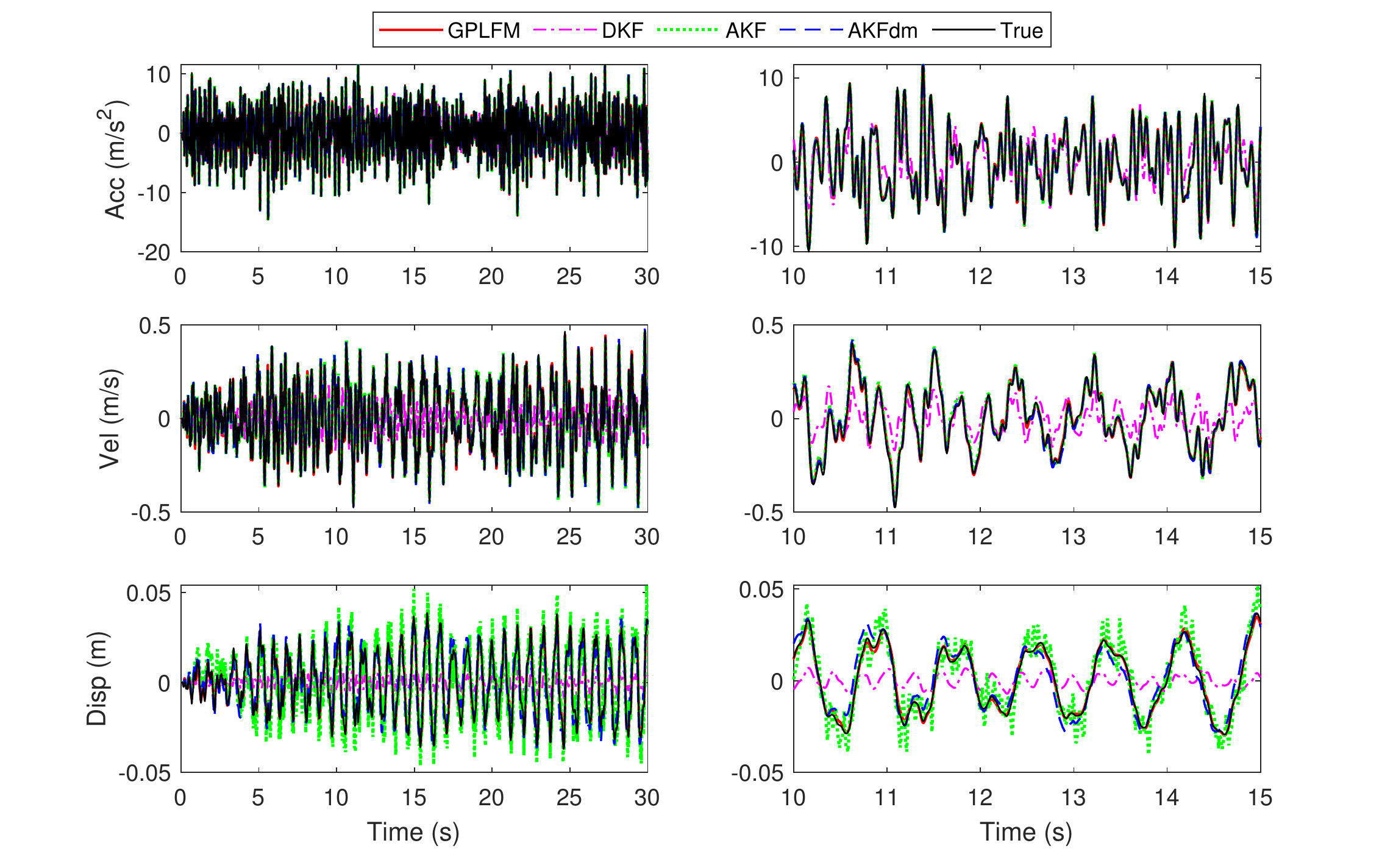}
			\caption{Estimated acceleration (top), velocity (middle) and displacement (bottom) time histories obtained from GPLFM, DKF, AKF and AKFdm at the 5th floor; random excitation applied at the 10th floor and non-collocated accelerations measured at floors 1, 3, 5, 7 and 9}
			\label{fig:random_fewacc_repdof5_states}
		\end{figure}

		\item \textit{Multiple random excitation}\\
		Here, a scenario of multiple random forces exciting the structure at all floor levels is considered. Random excitations having zero mean and standard deviation $0.1\si{kN}$ are applied at all floor levels, and acceleration measurements are assumed to be available at floors 1, 3, 5, 7 and 9. This represents a case where (a) the number of forces to be estimated are greater than the number of observations (10 forces  against 5 observations), and (b) the observations at floor levels 1, 3, 5, 	7 and 9 serve as both collocated observations and non-collocated observations; collocated for forces 1, 3, 5, 7, 9 and non-collocated for forces 2, 4, 6, 8, 10.
		
		For this the following values of covariances are chosen:
		\begin{itemize}
			\item $\Qn^f = 10^{3} \times \eye \;\si{N}^2$ for DKF and AKF chosen using L-curve criteron and some manual tuning,
			\item $\Qn^f = 10^{4} \times \eye \;\si{N}^2$ AKFdm chosen after some manual tuning, 
			\item $\Rn_{dm} = 10^{-2} \times \eye \; \si{m}^2$ for AKFdm set at approximately two times the square of the absolute maximum of true displacement.
		\end{itemize} 
		The estimated force time histories for two representative floors, 8th floor (non-collocated measurement) and 9th floor (collocated measurement), are illustrated in Figure \ref{fig:multicf_random8_9}. It can be seen that for the 8th floor, none of the algorithms are able to estimate the true force history correctly. Both the GPLFM and the AKF seem to provide no tracking of the true force at the 8th floor, besides AKF suffering from drift. The DKF is unable to estimate the forces at non-collocated measurement locations due to direct transmission term becoming zero leading to degeneracy in the input update stage of DKF. The AKFdm estimate of the force at the 8th floor seem to produce larger magnitudes but no accuracy in tracking the true force. Comparing this with the force estimates (in Figure \ref{fig:multicf_random8_9}) obtained at the 9th floor, it can be seen that GPLFM estimates of force history is quite accurate while AKF and DKF estimates for the collocated forces are able to somewhat trace the pattern of the force history albeit with significant drift. 		
		\begin{figure}[htbp!]
			\centering
			\includegraphics[scale=\mysize]{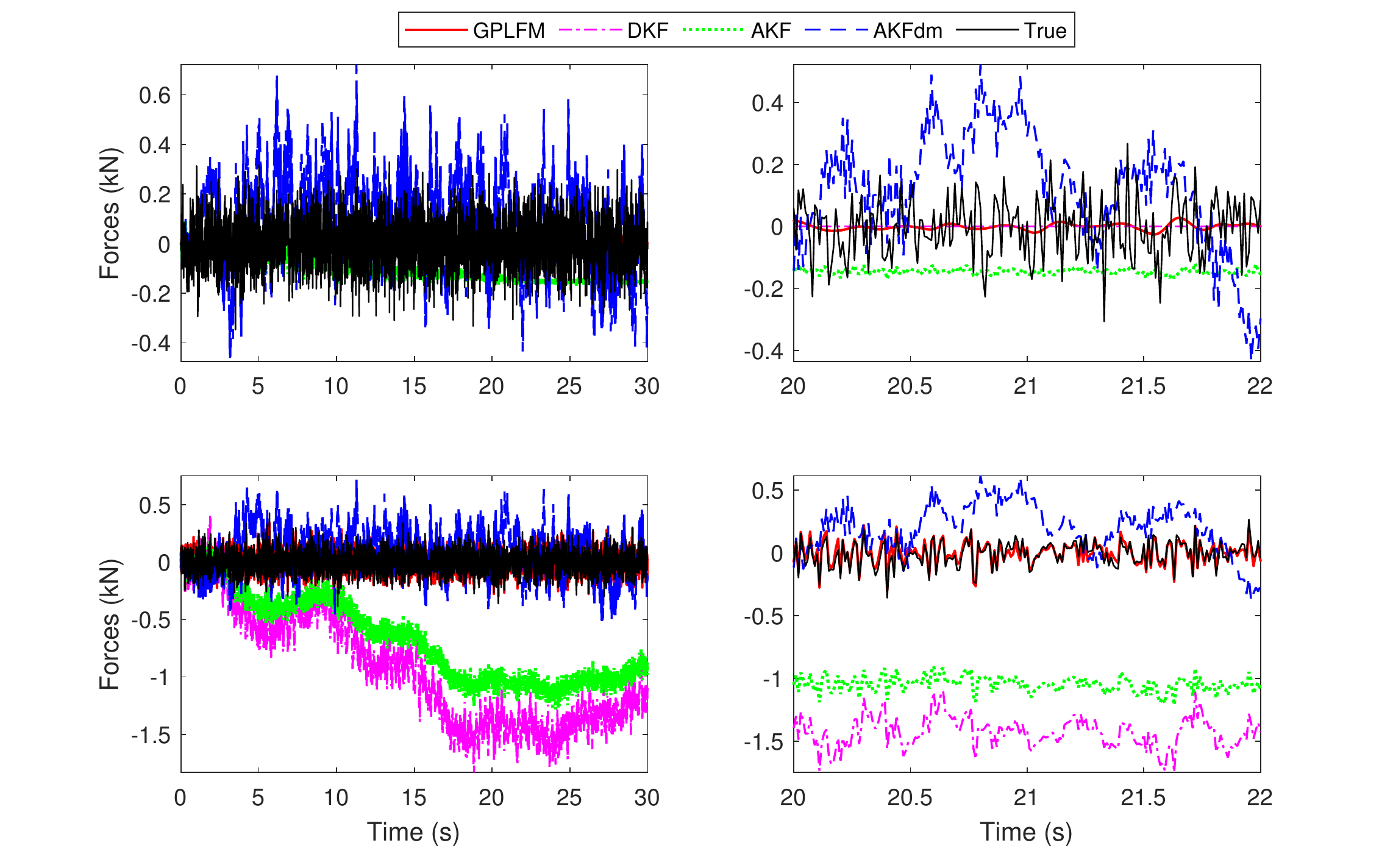}
			\caption{Estimated force time histories at 8th floor (top) and 9th floor (bottom) obtained from GPLFM, DKF, AKF and AKFdm using accelerations at floors 1, 3, 5, 7 and 9; random excitation applied at all floor levels}
			\label{fig:multicf_random8_9}
		\end{figure}
		
		The displacement and velocity state estimates at the 8th and 9th floors are shown in Figure \ref{fig:multicf_random_states8_9}. Both DKF and AKF are able to provide accurate velocity estimates alike GPLFM but suffer from drift in displacement estimates although they seem to follow the pattern of the true displacement state quite accurately. The GPLFM estimates for displacements and velocities are quite accurate both for the case of collocated and non-collocated measurements. 
		
		\begin{figure}[htbp!]
			\centering
			\includegraphics[scale=\mysize]{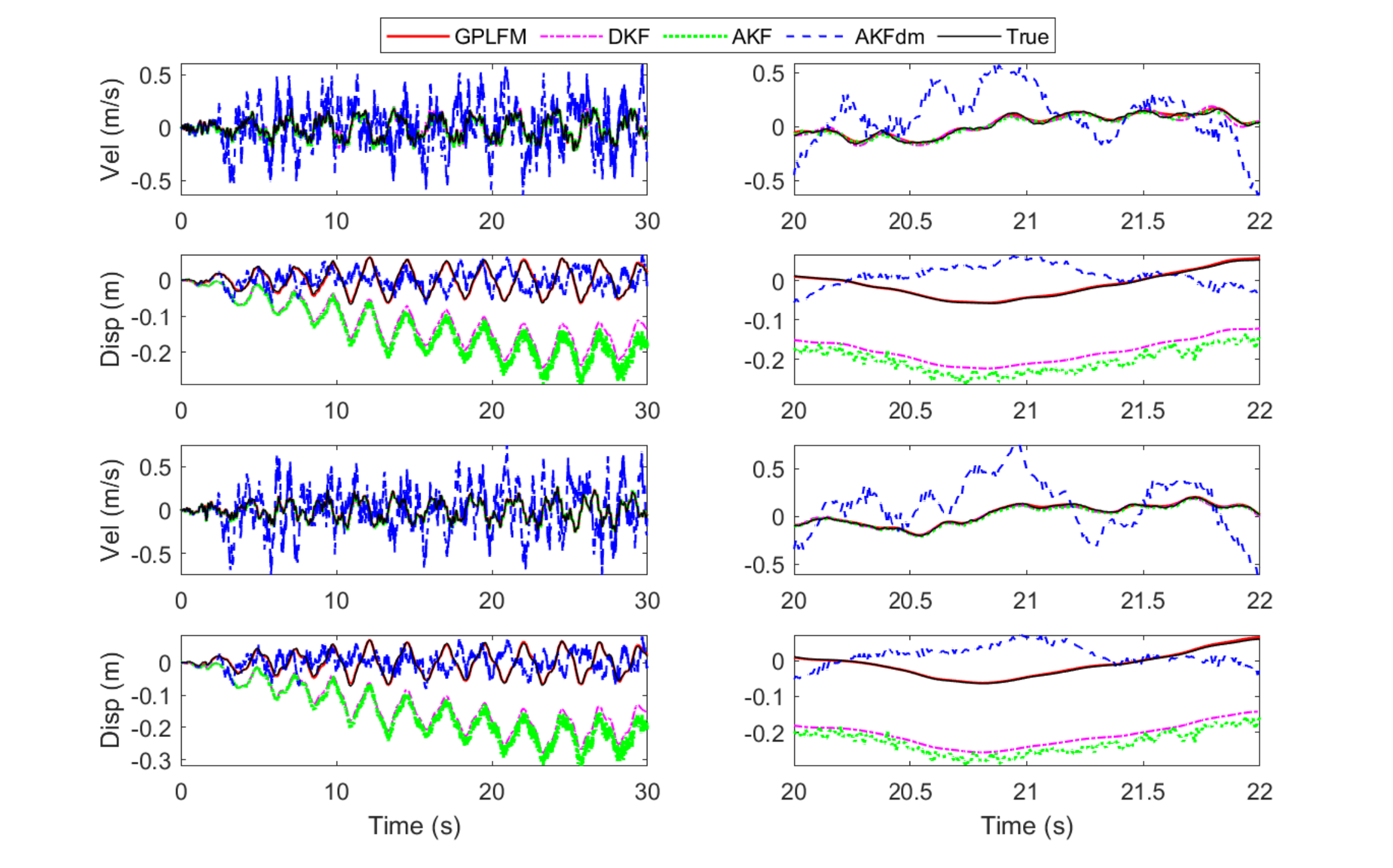}
			\caption{Estimated velocities and displacements at 8th floor (top two) and 9th floor (bottom two) using GPLFM, AKF and AKFdm for multiple random excitation applied at all floor levels; accelerations at floors 1, 3, 5, 7 and 9 are measured}
			\label{fig:multicf_random_states8_9}
		\end{figure}
	\end{enumerate}

	\section{Application: A 76-storey ASCE benchmark building} \label{sec:application}
\begin{figure}[htbp!]
		\centering
		\begin{subfigure}[b]{0.5\textwidth}
			\centering
			\includegraphics[scale = 0.25]{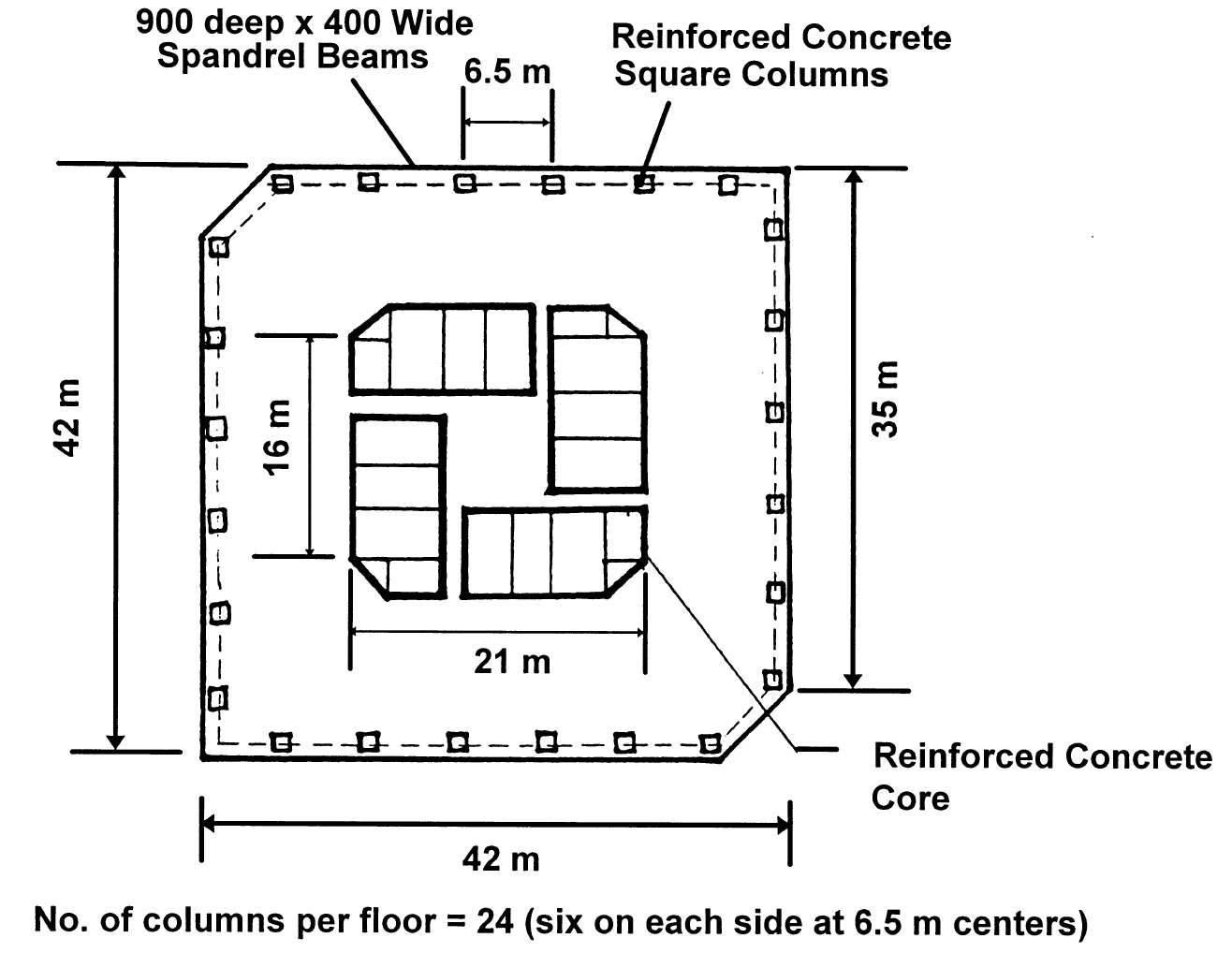}
			\caption{Plan}
		\end{subfigure}%
		~ 
		\begin{subfigure}[b]{0.5\textwidth}
			\centering
			\includegraphics[scale=0.3]{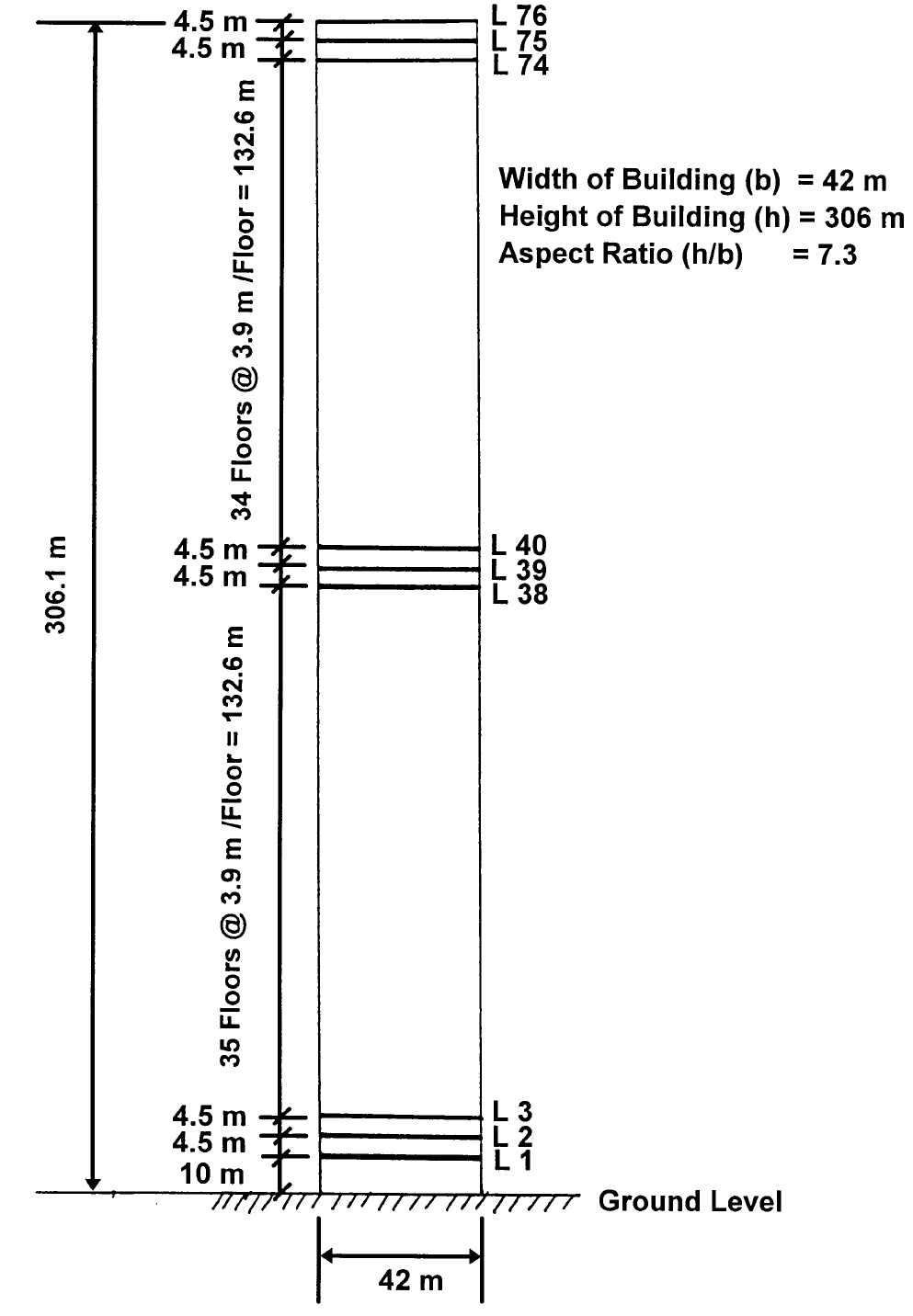}
			\caption{Elevation}
		\end{subfigure}
		\caption{Plan and elevation of the 76-storey ASCE benchmark building \cite{yang2004benchmark}}
		\label{fig:plan_elevation}
	\end{figure}

	An ASCE benchmark 76-storey office tower \cite{yang2004benchmark} is considered here for application. The 76-storey office tower was proposed for the city of Melbourne, Australia. The building plan and elevation view have been provided in Figure \ref{fig:plan_elevation}. 
	The 76-story tall building is modelled as a vertical cantilever beam. A finite element model is constructed by considering the portion of the building between two adjacent floors as a classical beam element of uniform thickness, leading to 76 translational and 76 rotational degrees of freedom. Then, all the 76 rotational degrees of freedom have been removed by the static condensation. This resulted in 76 degrees of freedom, representing the translation of each floor in the lateral direction. The first five natural frequencies are 0.16, 0.765, 1.992, 3.790 and 6.395 \si{Hz}. Damping ratios for the first five modes are assumed to be 1\%.  This model, having $76 \times 76$ mass, damping and stiffness matrices, is referred to as the 76-dof `true' model. For simplification of numerical computation, the 76-dof true model was reduced to a 23-dof system such that the first 46 ($= 2\times 23$) complex modes of the 76-DOF model are retained. The 23-dof system correspond to the translation at floors 3, 6, 10, 13, 16, 20, 23, 26, 30, 33, 36, 40, 43, 46, 50, 53, 56, 60, 63, 66, 70, 73 and 76, and is referred to as the 23-dof reduced-order model. Responses at nine floor levels are assumed to be available --- floor levels 1, 30, 50, 55, 60, 65, 70, 75 and 76. All relevant details and MATLAB files for the structure are provided in the website \cite{benchmarkdownload} for ASCE benchmark buildings.
	
\begin{figure}[htbp!]
		\centering
		\includegraphics[scale=\mysize]{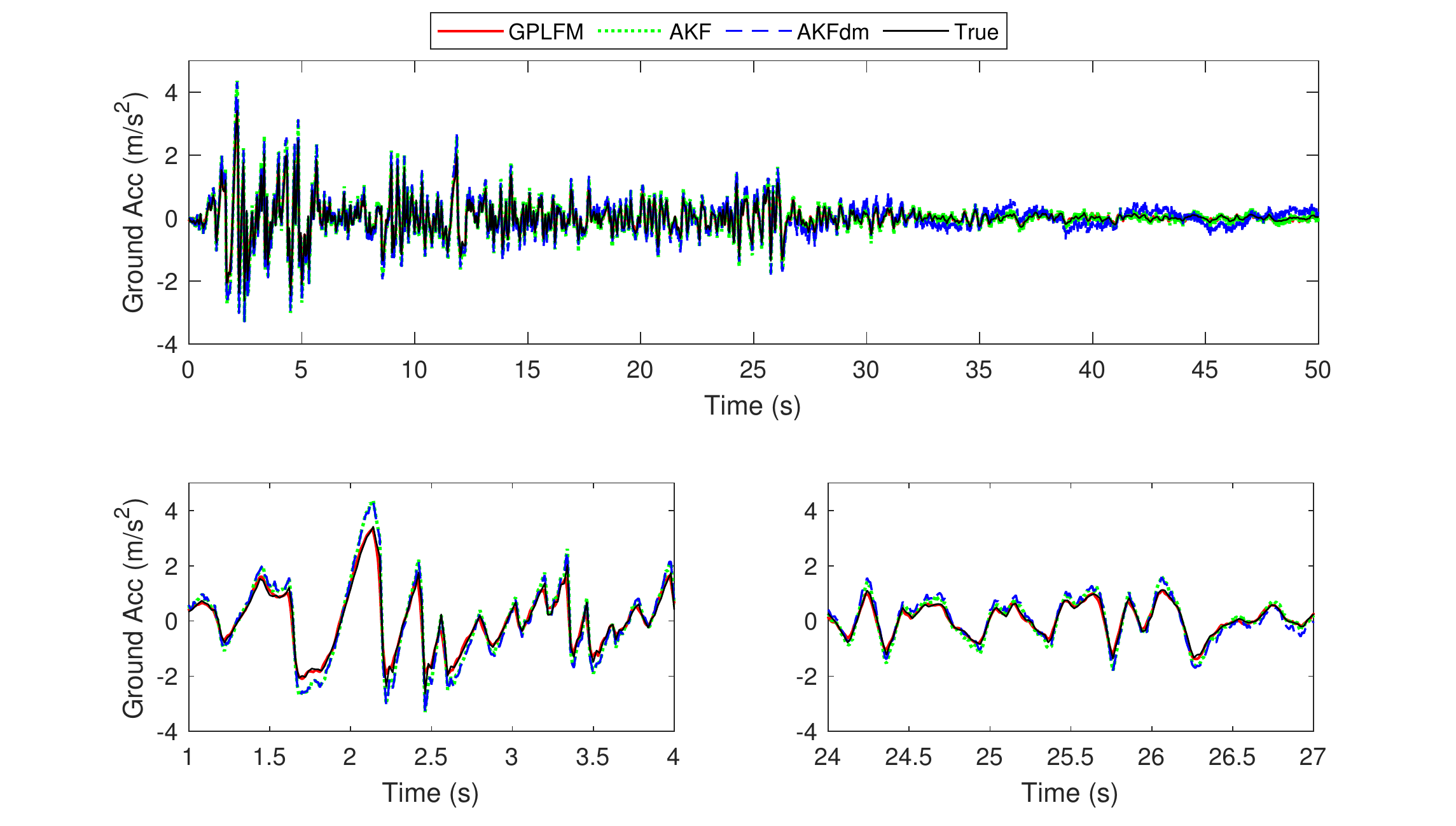}
		\caption{Estimated input ground acceleration time histories obtained from GPLFM, AKF and AKFdm using absolute accelerations at floors 1, 50 and 70; benchmark tower subjected to El Centro ground acceleration}
		\label{fig:SM_acc137_fullmodel_force_zoom}
	\end{figure}

\begin{figure}[htbp!]
		\centering
		\includegraphics[scale=\mysize]{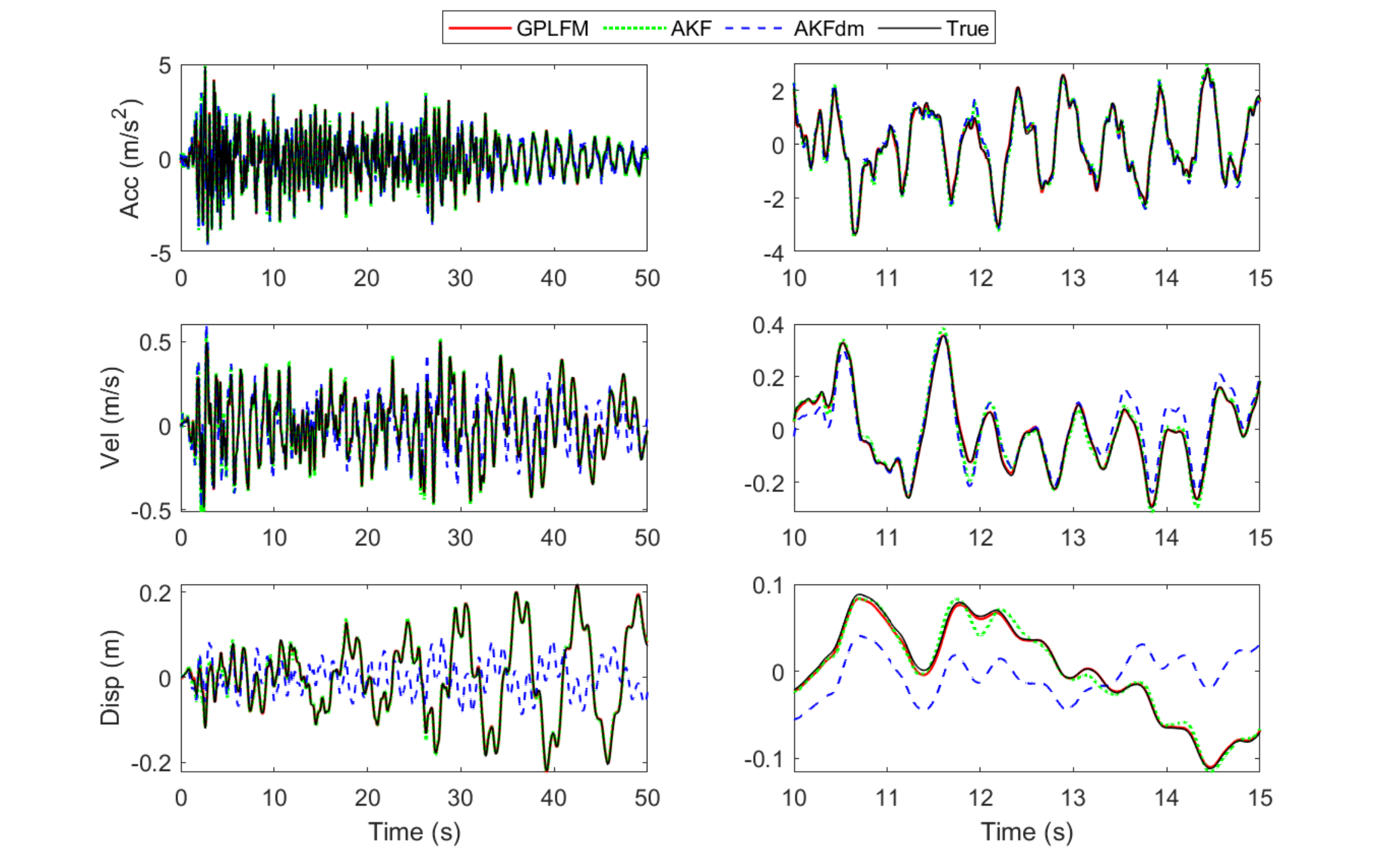}
		\caption{Estimated accelerations (top), velocities (middle) and displacements (bottom) at the 30th floor obtained from GPLFM, AKF and AKFdm; benchmark tower subjected to El Centro ground acceleration and absolute accelerations measured at floors 1, 50 and 70}
		\label{fig:SM_acc137_fullmodel_flr30states}
	\end{figure}

	Two separate scenarios of excitation, seismic and wind, are considered in this application. For the case of seismic excitation, the 76-dof true model is subjected to El Centro earthquake ground motion, and absolute acceleration responses, sampled at 100\si{Hz}, are generated at floor levels 1, 50 and 70, for a duration of 50\si{s}. The generated acceleration responses are then contaminated with 10\% measurement noise. These noisy measurements along with the 23-dof reduced-order model are used by GPLFM, AKF and AKFdm for input and state estimation. The following values are used for this scenario:
	\begin{itemize}
		\item $\Pc^x_{0|0} = 10^{-10} \times \eye$ and $\Qn^x = 10^{-10} \times \eye$ for GPLFM, AKF and AKFdm, 
		\item $\Rn = 0.5 \times \Rn_{true}$ for GPLFM, AKF and 	AKFdm, where $\Rn_{true}$ is the true measurement noise covariance 	with the $i$th diagonal element obtained as $\Rn_{true}(i,i) = 	0.01*\text{Var}(\y_i)$ 
		\item $\Qn^f = 10^{3} \times \eye \;\si{N}^2$ for AKF and AKFdm obtained using L-curve criterion and some manual tuning,
		\item $\Rn_{dm} = 10 \times \eye \; \si{m}^2$ for AKFdm set at approximately 10 times the square of absolute maximum of true displacement history,
		\item An exponential covariance function (refer to Equation \ref{eq:exp}) is used for GPLFM
	\end{itemize}
	The estimated ground input time histories are shown in Figure \ref{fig:SM_acc137_fullmodel_force_zoom}. The estimated acceleration, velocity and displacement states at floor level 30, chosen as the representative floor, are shown in Figure \ref{fig:SM_acc137_fullmodel_flr30states}. It is found that all the three algorithms i.e.\ GPLFM, AKF and AKFdm are able to track the ground motion input with good accuracy, with estimates from AKF and AKFdm showing slight overshoots. The state estimates obtained by GPLFM and AKF also match the true states quite accurately while the displacement estimates from AKFdm are under-estimated.
	
	Next, the 76-storey model is subjected to wind excitation at the topmost floor i.e.\ 76th floor. For wind excitation, 200\si{s} of wind loading history was obtained from the 76th column of \texttt{wind\_cross.mat} data file and then multiplied with a factor of 100; the multiplying factor was chosen such that the displacement obtained at the 76th floor was less than 0.5\si{m}. Two cases of measurements are considered: (a) three accelerations measured at floor levels 1, 50 and 70, and (b) one displacement measured at 1st floor in addition to three accelerations measured at floor levels 1, 50 and 70. Note that these measurements now resemble a non-collocated measurement scenario. The GPLFM, AKF and AKFdm are used for input and state estimation for these cases; results from DKF are omitted as it DKF has been shown to suffer from degeneracy when non-collocated acceleration measurements are used. The following modified values are used in estimation:
	\begin{itemize}
		\item $\Pc^x_0 = 10^{-10} \times \eye$ and $\Qn^x = 10^{-10} \times \eye$ for GPLFM, AKF and AKFdm, 
		\item $\Rn = 0.5 \times \Rn_{true}$ for GPLFM, AKF and AKFdm, where $\Rn_{true}$ is the true measurement noise covariance with the $i$th diagonal element obtained as $\Rn_{true}(i,i) = 0.01*\text{Var}(\y_i)$ 
		\item $\Qn^f = 10^{10} \times \eye \;\si{N}^2$ for AKF and $\Qn^f = 2 \times 10^{7} \times \eye \;\si{N}^2$ for AKFdm obtained with manual tuning,
		\item $\Rn_{dm} = 3 \times \eye \; \si{m}^2$ for AKFdm set at approximately 10 times the square of absolute maximum of true displacement history.
		\item An exponential covariance function is used for GPLFM
	\end{itemize}
	\begin{enumerate}[label = (\alph*)]
		\item \textit{Three acceleration measurements}\\	
		The estimated displacement, velocity and acceleration states at the 30th floor level are shown in Figure \ref{fig:W_acc137_fullmodel_flr30states}. It is seen that the state estimates from GPLFM and AKF are reasonably close to the true value, however, the velocity and displacement estimates from AKFdm do not match the true states that closely. 
		\begin{figure}[htbp!]
			\centering
			\includegraphics[scale=\mysize]{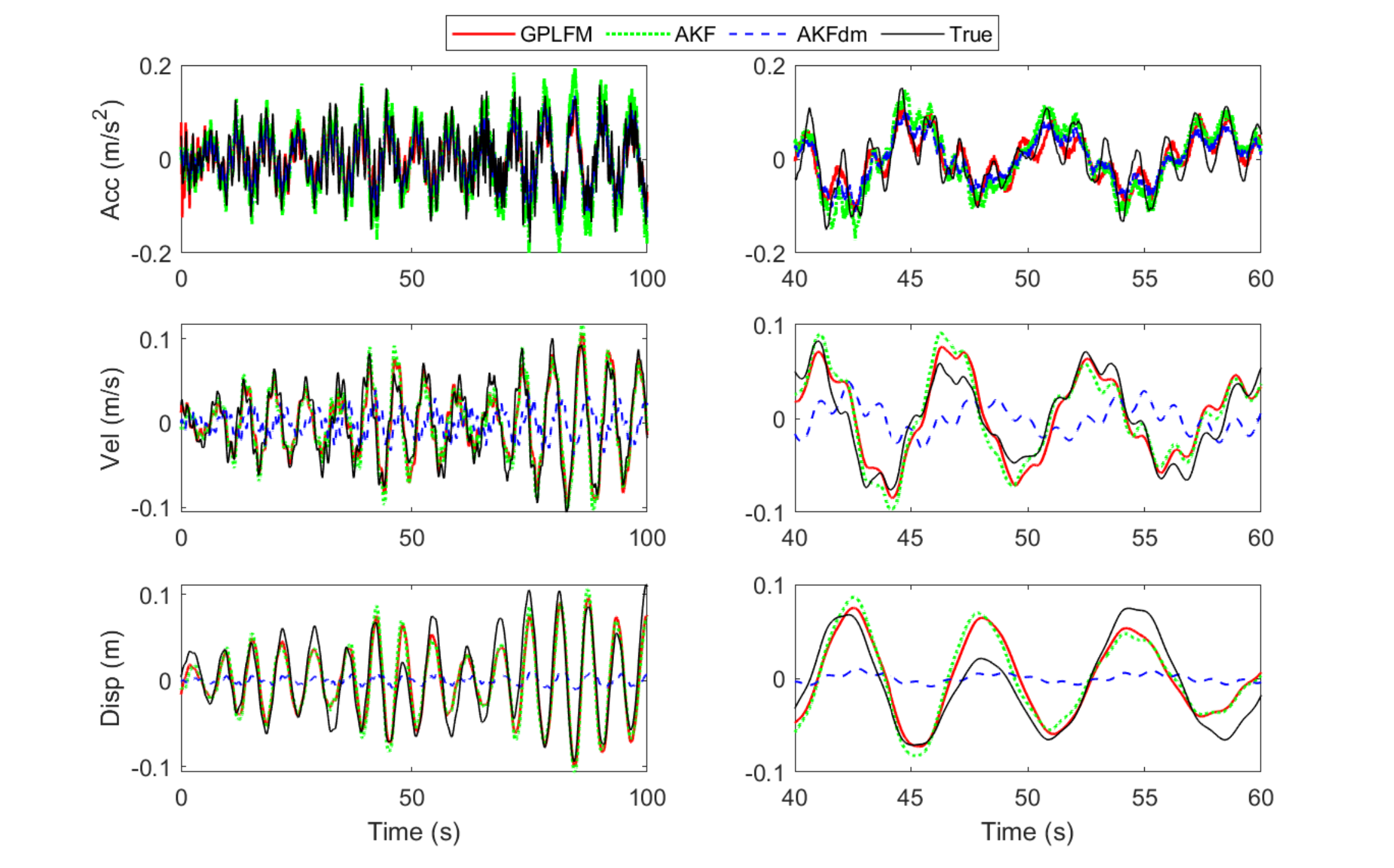}
			\caption{Estimated accelerations (top), velocities (middle) and displacements (bottom) at the 30th floor obtained from GPLFM, AKF and AKFdm;  benchmark tower subjected to wind excitation at the 76th floor and accelerations measured at floors 1, 50 and 70} 
			\label{fig:W_acc137_fullmodel_flr30states}
		\end{figure}
		\begin{figure}[htbp!]
			\centering
			\includegraphics[scale=\mysize]{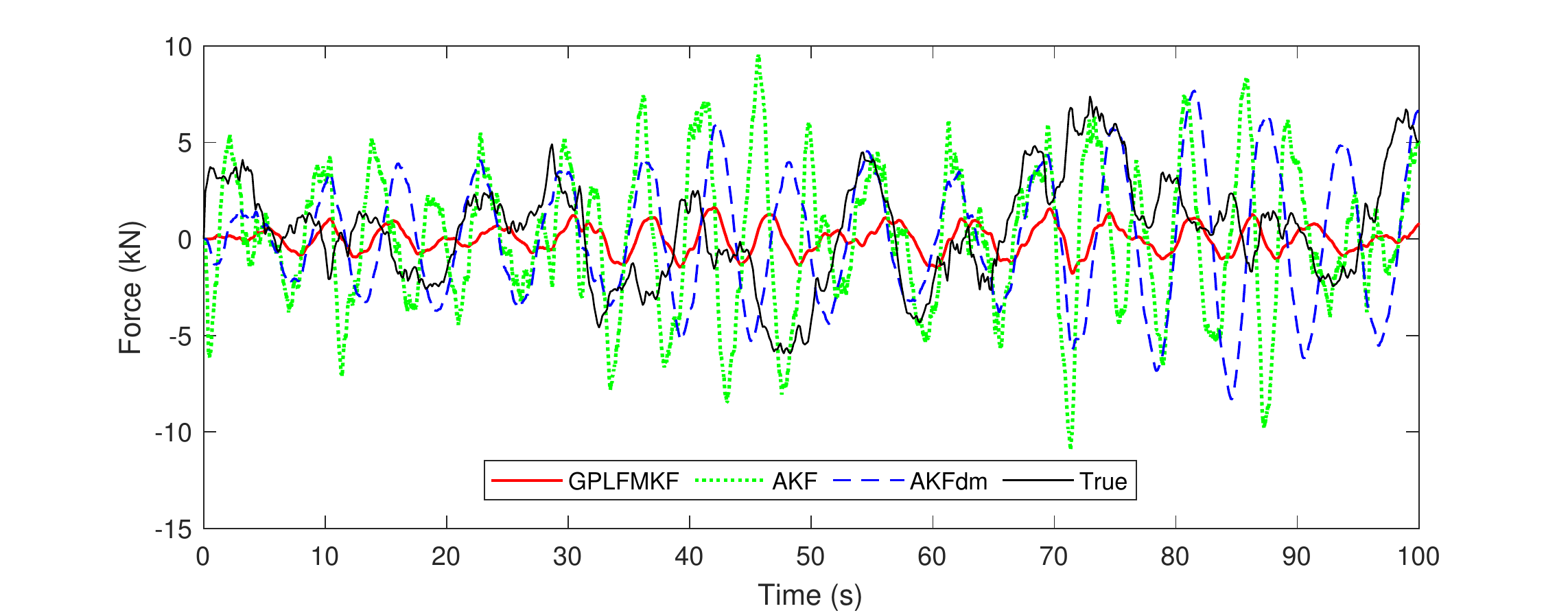}
			\caption{Estimated force time histories obtained from GPLFM, AKF and AKFdm using for accelerations at floors 1, 50 and 70; benchmark tower subjected to wind excitation at the 76th floor}	\label{fig:W_acc137_fullmodel_force_zoom}
		\end{figure}
		From the estimated force time histories shown in Figure \ref{fig:W_acc137_fullmodel_force_zoom}, it is apparent that none of the algorithms are able to provide proper tracking of the true input force, nevertheless, the state estimates obtained from GPLFM and AKF are still reasonable. Using nine more acceleration measurements did not help much in improving the force estimates. To improve the force estimation, the use of an additional displacement measurement is considered next.
		
		\item \textit{One displacement and three acceleration measurements}\\
		A displacement measurement at the 1st floor is added to three accelerations measured at floors 1, 50 and 70 to improve upon the input force estimate from the previous case. A readjusted value of  $\Qn^f = 10^{2} \times \eye \;\si{N}^2$ is now used for AKF and AKFdm. The estimated forces time histories are shown in Figure \ref{fig:W_acc137_disp1_fullmodel_force_filter}. The force estimates obtained from GPLFM, AKF and AKFdm are very close to each other and seem to track the force history with a constant delay of around 2\si{s}. This delayed behavior can sometimes be expected in force estimation approach as reported in \cite{naets2014multibody, naets2015structdyn}. The delayed behavior can be removed by applying a Kalman smoother as shown in Figure \ref{fig:W_acc137_disp1_fullmodel_force_smoother}.
        \begin{figure}[htbp!]
        \centering
        \begin{subfigure}{1.05\textwidth}
            \centering
            \includegraphics[width=\linewidth]{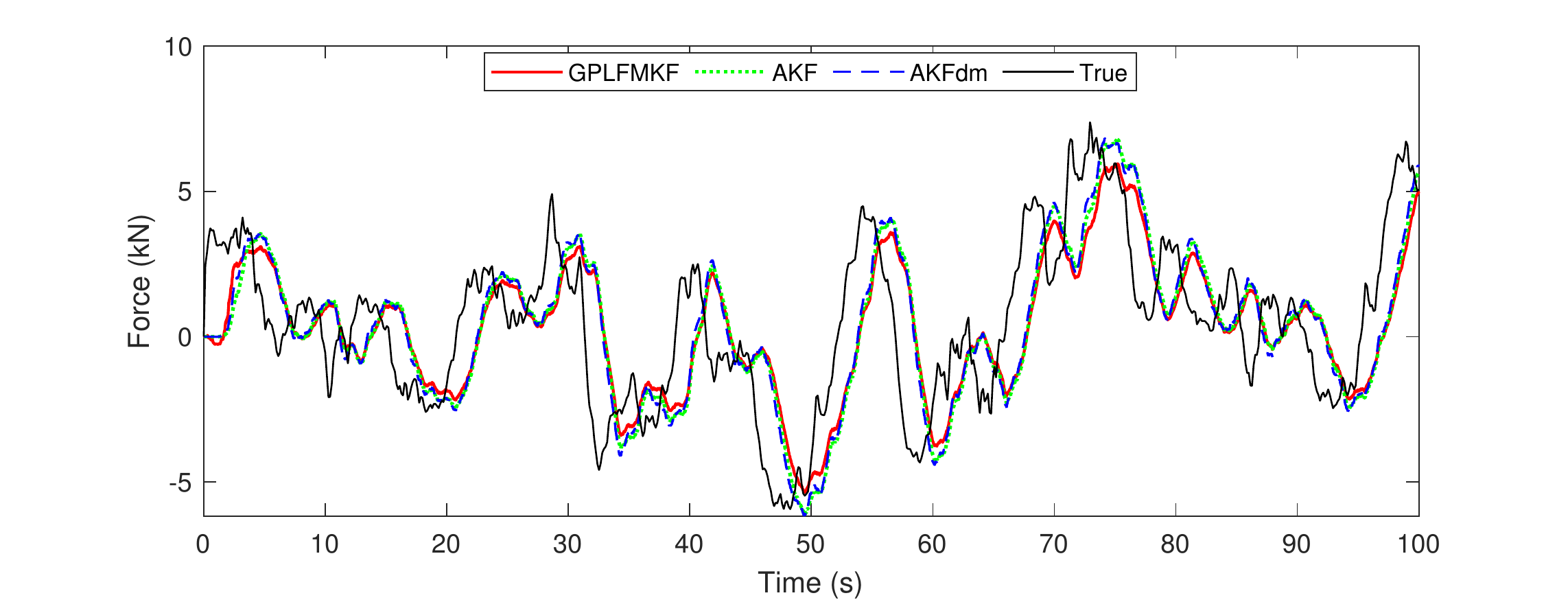}
            \caption{GPLFM with Kalman filter}
            \label{fig:W_acc137_disp1_fullmodel_force_filter}
        \end{subfigure}
        \begin{subfigure}{1.05\textwidth}
            \centering
            \includegraphics[width=\linewidth]{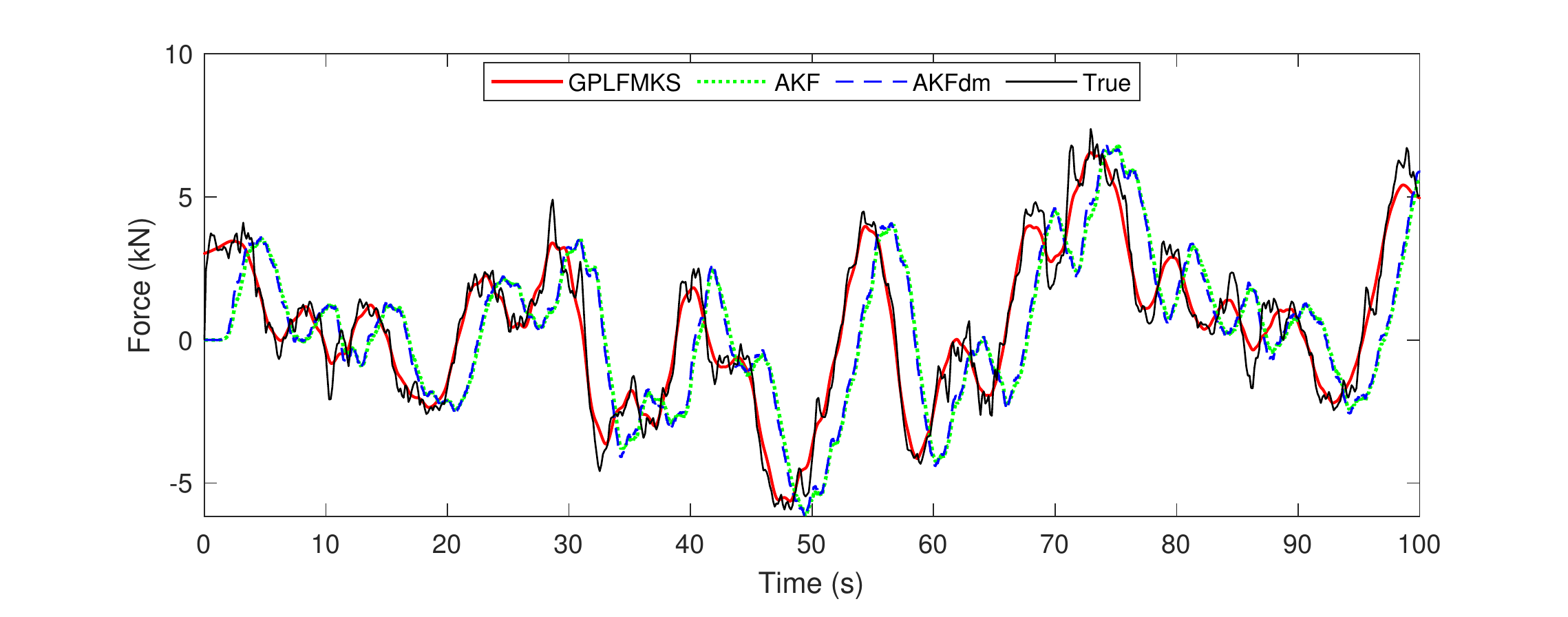}
            \caption{GPLFM with Kalman smoother}
            \label{fig:W_acc137_disp1_fullmodel_force_smoother}
        \end{subfigure}
        \caption{Estimated force time histories obtained from GPLFM, AKF and AKFdm using displacement at 1st floor and accelerations at floors 1, 50 and 70; benchmark tower subjected to wind excitation at the 76th floor}
        \end{figure}
		The acceleration, velocity and displacement state estimates for the 30th floor show good agreement with the true values as depicted in Figure \ref{fig:W_acc137_disp1_fullmodel_flr30states}.
		\begin{figure}[htbp!]
			\centering
			\includegraphics[scale=\mysize]{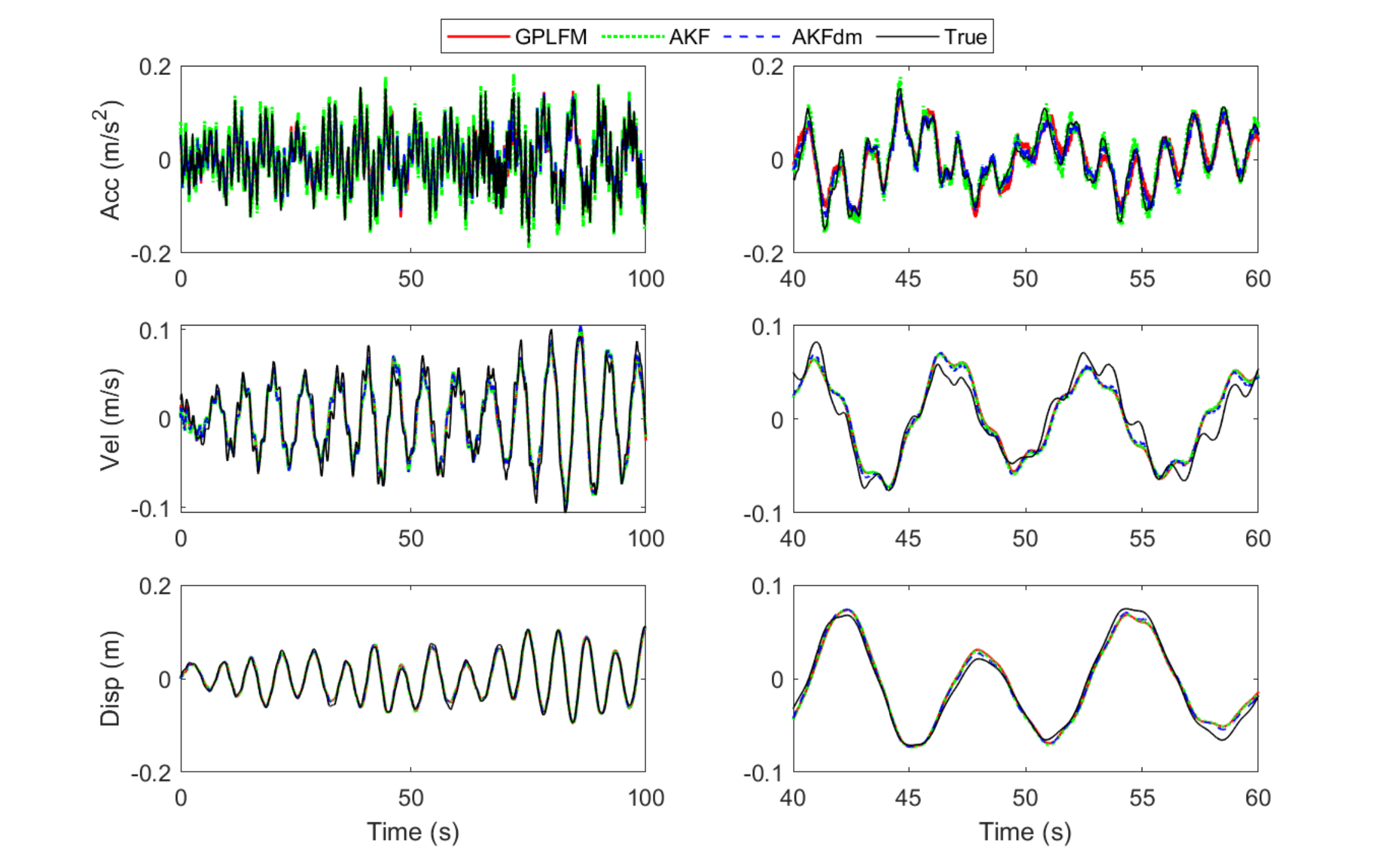}
			\caption{Estimated accelerations (top), velocities (middle) and displacements (bottom) at the 30th floor obtained from GPLFM, AKF and AKFdm; benchmark tower subjected to wind excitation at the 76th floor, and a displacement at 1st floor and accelerations at floors 1, 50, 70 are measured}
			\label{fig:W_acc137_disp1_fullmodel_flr30states}
		\end{figure}

	\end{enumerate} 	
	
	\section{Conclusions}
	In this paper, a novel implementation of GPLFM for joint input-state estimation of linear time-invariant structural systems with unknown inputs is proposed. The unknown inputs to a linear structural system are modelled as GPs with specified covariance functions. The GPLFM is then formulated into an augmented state-space model with additional states representing the latent force components. The GPLFM state-space model so obtained is shown to be a generalization of the augmented state-space models used in popular Kalman filter-based joint input-state estimation algorithms. Moreover, the GPLFM formulation is analytically proven to be observable and robust to drift in force estimation even when used with only acceleration measurements. The behavior of the Gaussian process latent force components are determined by the hyperparameters of the chosen covariance functions. In this study, these hyperparameters are tuned using maximum likelihood optimization based on the observed data avoiding the need to set them based on expert guess. 
	
	The effectiveness and performance of GPLFM is demonstrated through several numerical studies using different excitations and measurement scenarios on a 10-dof shear building as well as a numerical model of the 76-dof ASCE benchmark office tower. The performance of GPLFM is compared to that of DKF, AKF and AKFdm. It is found from the numerical studies that GPLFM is able to deliver very accurate estimates of inputs and states in most cases. Moreover, GPLFM does not suffer from drift in force and/or displacement estimates, unlike AKF and DKF. The choice of covariance function, however, plays an important role in the accuracy of the input estimation and relies on the user's ability to encode prior information of the behavior of the latent forces such as discontinuity, periodicity, etc. Finally, it should be mentioned that similar to Kalman filter-based joint input-state estimation approaches such as AKF, DKF and AKFdm, GPLFM is amenable to online implementation once the hyperparameters of the covariance function are properly calibrated.
	
	\section{Acknowledgments}
	Funding for this study through the Natural Sciences Engineering Research Council of Canada (Canada Research Chairs program) is gratefully acknowledged by the authors.

\appendix
\section{Mat\'{e}rn class of covariance functions} \label{sec:Matern}
A commonly used class of covariance functions is the Mat\'{e}rn family 
\cite{rasmussen2006gaussian, whittle1954stationary, matern1960spatial}
\begin{equation} \label{eq:maternmain}
\kappa(\tau; \nu, \alpha, l) = \alpha^2 \frac{2^{1-\nu}}{\Gamma(\nu)} \left( 
\frac{\sqrt{2 \nu} \tau}{l}\right)^{\nu} K_\nu \left(\frac{\sqrt{2 \nu} 
\tau}{l}\right)
\end{equation}
where $\alpha^2$ and $l$ are positive hyperparameters that denote the signal variance 
and lengthscale respectively, and $\nu$ is a parameter controlling the 
smoothness of the Gaussian process. In this paper, the set of hyperparameters of a 
covariance function is denoted by $\tcf = \{\alpha^2, l\}$. These 
hyperparameters control the variability of 
the underlying Gaussian process. $\Gamma(\nu)$ is the Gamma function, $K_{\nu}$ is a modified Bessel function of second kind 
\cite{abramowitz1965handbook}. For one-dimensional processes, the spectral 
density of the Mat\'{e}rn covariance function (Equation \ref{eq:maternmain}) is
\begin{equation}
S(\omega) = \alpha^2 \frac{2 \pi^{0.5}\Gamma(\nu+1/2)}{\Gamma(\nu)} 
\lambda^{2\nu} \left(\lambda^2 + \omega^2\right)^{-(\nu + 1/2)}
\end{equation}
where $\lambda = \sqrt{2 \nu}/l$, and in this work $\nu = p +1/2$, where $p$ is 
a non-negative integer. Thus,
\begin{equation}
S(\omega) \propto \left(\lambda^2 + \omega^2 \right)^{-(p+1)}
\end{equation}
This has a proper rational fraction form and can be spectrally factorized as
\begin{equation}
S(\omega) \propto \left(\lambda + i \omega\right)^{-(p-1)} \left(\lambda - i 
\omega\right)^{-(p+1)}
\end{equation}
from which a stable transfer function can be obtained as
\begin{equation}
H(i\omega) = \left(\lambda - i \omega\right)^{-(p+1)}
\end{equation}
The corresponding spectral density of the Gaussian white noise process $w(t)$ is
\begin{equation}
\sigma_w = \frac{2 \alpha^2 \pi^{0.5} \lambda^{(2p+1)} 
\Gamma(p+1)}{\Gamma(p+1/2)}
\end{equation}
\begin{enumerate} [label = (\alph*)]
	\item Setting $p=0$, the exponential covariance function is obtained with 
	$\lambda = 1/l$, $\sigma_w = 2\alpha^2 / l$ and the corresponding LTI model 
	matrices in Equation \ref{eq:gpssm} read as: 
	\begin{equation} \label{eq:exp}
	\F_{cf} = -\lambda, \; \matr{L}_{cf} = 1, \; \matr{H}_{cf} = 1
	\end{equation}
	
	\item Setting $p=1$, the Mat\'{e}rn class with $\nu = 3/2$ is obtained with 
	$\lambda = \sqrt{3}/l$, $\sigma_w = 12\sqrt{3} \alpha^2 / l^3$ and the 
	corresponding LTI model matrices in Equation \ref{eq:gpssm} read as: 
	\begin{equation}
	\F_{cf} = \begin{bmatrix}	0 & 1 \\ -\lambda^2 & -2\lambda 
	\end{bmatrix}, \; \matr{L}_{cf} = \begin{bmatrix}
	0 \\ 1	\end{bmatrix}, \; \matr{H}_{cf} = \begin{bmatrix}
	1 & 0
	\end{bmatrix}
	\end{equation}
	
	\item Setting $p=2$, the Mat\'{e}rn class with $\nu = 5/2$ is obtained with 
	$\lambda = \sqrt{5}/l$, $\sigma_w = 400\sqrt{5} \alpha^2 / 3l^5$ and the 
	corresponding LTI model matrices in Equation \ref{eq:gpssm} read as: 
	\begin{equation} \label{eq:Matern52}
	\F_{cf} = \begin{bmatrix}	0 & 1 & 0 \\ 0 & 0 & 1 \\-\lambda^3 & 
	-3\lambda^2 & 
	-3\lambda \end{bmatrix}, \; \matr{L}_{cf} = \begin{bmatrix}
	0 \\ 0\\ 1	\end{bmatrix}, \; \matr{H}_{cf} = \begin{bmatrix}
	1 & 0 & 0
	\end{bmatrix}
	\end{equation}
\end{enumerate}

\section{Kalman filter and smoother} \label{sec:Kfs}
Consider a discrete-time time-invariant linear system with additive Gaussian 
noise
\begin{align}
\begin{split}
\x_k &= \F \x_{k-1} + \wn_{k-1}\\
\y_k &= \matr{H} \x_k + \vn_k
\end{split}
\end{align}
where $\x_0 \sim \mathcal{N}(\hat{\m}_{0|0}, \Pc_{0|0})$ is the initial state 
distribution, $\wn_{k-1} \sim \mathcal{N}(\zeros, \Qn)$ and $\vn_{k} \sim 
\mathcal{N}(\zeros, \Rn)$ are the process noise and the measurement noise 
distribution respectively, with zero cross-correlation between $\wn_{k-1}$ and $\vn_k$.  
For such a system, the Kalman filter \cite{kalman1960} computes the mean and the 
covariance of posterior filtering distribution sequentially using the following 
steps:		
\begin{align}
\hat{\m}_{k|k-1} &= \F  \hat{\m}_{k-1|k-1}  \\
\Pc_{k|k-1} &= \F \Pc_{k-1|k-1} \F + \Qn \\
\ve{e}_k &= \y_k - \matr{H} \hat{\m}_{k|k-1}	\label{eq:innov} \\
\matr{S}_k & = \matr{H} \Pc_{k|k-1} \matr{H}^T + \Rn \label{eq:Scov} \\
\matr{K}_k &= \Pc_{k|k-1} \matr{H}^T \matr{S}_k^{-1} \\
\hat{\m}_{k|k} &= \hat{\m}_{k|k-1} + \matr{K}_k \ve{e}_k \\
\Pc_{k|k} &= \Pc_{k|k-1} - \matr{K}_k \matr{S}_k \matr{K}_k^T
\end{align}
for $k = 1,\ldots,N$. Here $\hat{\m}_{k|k-1}$ and $\hat{\m}_{k|k}$ represent 
the $k$th predicted and filtered state estimate, respectively, and, $\Pc_{k|k-1}$ 
and $\Pc_{k|k}$ denote the $k$th predicted and filtered state error covariance 
matrices respectively. The recursion in Kalman filter is started from initial 
state $\hat{\m}_{0|0}$ and covariance $\Pc_{0|0}$. Following filtering step, 
the (fixed interval) smoothing step is given by the Rausch-Tung-Striebel (RTS) smoother \cite{rauch1965maximum}:
\begin{align}
\matr{N}_k &= \Pc_{k|k} \F^T \left(\Pc_{k+1|k}\right)^{-1}\\
\hat{\m}_{k|N} &= \hat{\m}_{k|k} + \matr{N}_k \left(\hat{\m}_{k+1|N} - 
\hat{\m}_{k+1|k} \right) \\
\Pc_{k|N} &= \Pc_{k|k} +  \matr{N}_k \left(\Pc_{k+1|N} - \Pc_{k+1|k} \right) 
\matr{N}_k^T 
\end{align}
for $k = N-1,\ldots,1$.


\section{Derivation of $\matr{K}_{xf}(t,t')$} \label{sec:Kfx_derivation}
A complete derivation of the expression in Equation \ref{eq:Kxf} is provided here.
\begin{align*}
\begin{split}
\matr{K}_{xf}(t,t') &= \E{\x(t) \f(t')^T} \\
&= \E{\left( \bm{\Psi}(t) \x_0 + \int_0^t \bm{\Psi}(t-s) \Bc \f(s) ds \right) \f(t')^T} \quad \text{(using Equation \ref{eq:GPx})}\\
&= \E{ \bm{\Psi}(t) \x_0 \f(t')^T  + \int_0^t \bm{\Psi}(t-s) \Bc \f(s) \f(t')^T ds  }\\
&= \bm{\Psi}(t) \E{ \x_0 \f(t')^T } + \int_{0}^{t} \bm{\Psi}(t-s) \Bc \E{\f(s) \f(t')^T} ds \quad \text{($\because \;\E{\cdot}$ is linear)}\\
&= \bm{\Psi}(t) \E{ \x_0 \f(t')^T } + \int_{0}^{t} \bm{\Psi}(t-s) \Bc \matr{K}_{ff}(s,t') ds \quad \text{(using Equation \ref{eq:Kff})}\\
&= \zeros + \int_{0}^{t} \bm{\Psi}(t-s) \Bc \matr{K}_{ff}(s,t') ds\\
&= \int_{0}^{t} \bm{\Psi}(t-s) \Bc \matr{K}_{ff}(s,t') ds
\end{split}
\end{align*}
where in the second last line, the initial state $\x_0$ is assumed to be independent of the force applied $\f(t)$ leading to $\E{ \x_0 \f(t')^T } = \E{ \x_0 } \E{ \f(t')^T }$ and then using $\E{\f(t)} = \zeros$ (from the definition of GP over $\f(t)$ from Eqn 22), we get $\E{ \x_0 } \E{ \f(t')^T } = \zeros$.

\section{Direct-feedthrough matrix in case of seismic excitation} \label{sec:feedthru_derive}
When a structural system is 
subjected to earthquake ground motion, there is no external physical force 
acting on the structure. The equation of motion of the structure, when 
considering absolute motion $\ve{u}_{a}(t)$ of the structure, is given 
by  
\begin{equation} \label{eq:eom_abssm}
\matr{M} \ddot{\ve{u}}_{a}(t) + \matr{C} \left[ \dot{\ve{u}}_{a}(t) - 
\matr{S}_g \dot{\ve{u}}_g(t) \right] + 
\matr{K}  \left[{\ve{u}}_{a}(t) - \matr{S}_g {\ve{u}}_g(t) \right]  = 
\ve{0}
\end{equation}
where $\ve{u}_g(t) \in \R^{n_g}$ is the time-history of ground
motion caused due to earthquake. Often for convenience, one rewrites Equation 
\ref{eq:eom_abssm} in terms of relative motion, $\ve{u}(t) = \ve{u}_{a}(t) - 
\matr{S}_g \ve{u}_g(t) $, 
and one obtains
\begin{equation} \label{eq:eom_relsm}
\matr{M} \ddot{\ve{u}}(t) + \matr{C} \dot{\ve{u}}(t) + 
\matr{K}  \ve{u}(t)  = - \matr{M} \matr{S}_g \ddot{\ve{u}}_g(t)
\end{equation}The absolute acceleration response of the 
structure can be obtained as
\begin{align}
\begin{split}
\ddot{\ve{u}}_{a}(t) &= \ddot{\ve{u}}(t) + \matr{S}_g \ddot{\ve{u}}_g(t)\\
    &= \left( -\matr{M}^{-1} \matr{K} \ve{u}(t) - \matr{M}^{-1} 
    \matr{C} \dot{\ve{u}}(t) - \matr{S}_g \ddot{\ve{u}}_g(t) \right) 
    + \matr{S}_g \ddot{\ve{u}}_g(t) \\
    &= -\matr{M}^{-1} \matr{K} \ve{u}(t) - \matr{M}^{-1} \matr{C} 
    \dot{\ve{u}}(t)\\
    &= \underbrace{\begin{bmatrix}
    -\matr{M}^{-1} \matr{K} & -\matr{M}^{-1} \matr{C}
    \end{bmatrix}}_{\Gc} 
	\begin{bmatrix}
	\ve{u}(t) \\ \dot{\ve{u}}(t)
	\end{bmatrix} 
\end{split}
\end{align}
Since accelerometers always measure absolute accelerations, therefore under the influence of seismic excitations the direct-feedthrough term $\Jc$ vanishes.

\end{document}